\documentclass[10pt,journal,compsoc,x11names]{IEEEtran}
\usepackage{amsmath,amsfonts}
\usepackage{algorithmic}
\usepackage{algorithm}
\usepackage{array}
\usepackage[caption=false,font=normalsize,labelfont=sf,textfont=sf]{subfig}
\usepackage{textcomp}
\usepackage{stfloats}
\usepackage{url}
\usepackage{verbatim}
\usepackage{graphicx}
\usepackage{cite}
\usepackage{multirow}
\usepackage{hyperref}

\usepackage{forest}
\usetikzlibrary{shadows}
\usepackage{amsmath}

\usepackage{amssymb}
\usepackage{cleveref}

\definecolor{lightcoral}{rgb}{0.94, 0.5, 0.5}
\definecolor{lightgreen}{rgb}{0.56, 0.93, 0.56}
\definecolor{harvestgold}{rgb}{0.85, 0.57, 0.0}
\definecolor{brightlavender}{rgb}{0.75, 0.58, 0.89}
\definecolor{capri}{rgb}{0.0, 0.75, 1.0}
\definecolor{carminepink}{rgb}{0.92, 0.3, 0.26}
\definecolor{celadon}{rgb}{0.67, 0.88, 0.69}
\definecolor{darkpastelgreen}{rgb}{0.01, 0.75, 0.24}

\crefformat{section}{\S#2#1#3} 
\crefformat{subsection}{\S#2#1#3}
\crefformat{subsubsection}{\S#2#1#3}

\usepackage{amsthm}
\usepackage{amsmath}%
\usepackage{MnSymbol}%

\makeatletter
\newtheorem*{rep@theorem}{\rep@title}
\newcommand{\newreptheorem}[2]{%
\newenvironment{rep#1}[1]{%
 \def\rep@title{#2 \ref{##1}}%
 \begin{rep@theorem}}%
 {\end{rep@theorem}}}
\makeatother
\newtheorem{theorem}{Theorem}[section]
\newreptheorem{theorem}{Theorem}

\newreptheorem{lemma}{Lemma}

\newreptheorem{proposition}{proposition}
\newtheorem{definition}[theorem]{Definition}

\usepackage{bm}

\usepackage{booktabs}
\usepackage{xcolor,pifont}
\usepackage{bbding}
\usepackage{colortbl}

\usepackage[inline,shortlabels]{enumitem}

\usepackage{tikz,xcolor,hyperref}% Make Orcid icon
\definecolor{lime}{HTML}{A6CE39}
\DeclareRobustCommand{\orcidicon}{%
    \begin{tikzpicture}
    \draw[lime, fill=lime] (0,0) 
    circle [radius=0.16] 
    node[white] {{\fontfamily{qag}\selectfont \tiny ID}};    \draw[white, fill=white] (-0.0625,0.095) 
    circle [radius=0.007];    \end{tikzpicture}
    \hspace{-2mm}}
\foreach \x in {A, ..., Z}{%
    \expandafter\xdef\csname orcid\x\endcsname{\noexpand\href{https://orcid.org/\csname orcidauthor\x\endcsname}{\noexpand\orcidicon}}
}

\begin{document}

\title{Lifelong Learning of Large Language Model \\based Agents: A Roadmap}

\author{
Junhao~Zheng$^{*}$, 
Chengming~Shi$^{*}$, 
Xidi~Cai$^{*}$, 
Qiuke~Li$^{*}$, 
Duzhen~Zhang,
Chenxing~Li, 

Dong~Yu\orcidA{},~\IEEEmembership{Fellow,~IEEE}, Qianli~Ma$^\dagger$,~\IEEEmembership{Member,~IEEE}% <-this % stops a space
\thanks{Version: v1 (major update on January 13, 2025)}
\thanks{$^\dagger$Corresponding author: Qianli Ma.}
\thanks{$^{*}$The first four authors contributed equally to this research.}
\thanks{Junhao Zheng, Chengming Shi, Xidi Cai, Qiuke Li, Qianli Ma are with the School of Computer Science and Engineering, South China University of Technology, Guangzhou 510006, China (E-mail: junhaozheng47@outlook.com; cscmshi@mail.scut.edu.cn; xidicai067@gmail.com; lqk867543@gmail.com; qianlima@scut.edu.cn).}
\thanks{Duzhen Zhang is with the Mohamed bin Zayed University of Artificial Intelligence, Abu Dhabi, UAE (E-mail: bladedancer957@gmail.com).}
\thanks{Chenxing Li is with the Tencent, AI Lab, Beijing, China (E-mail: chenxingli@tencent.com).}
\thanks{Dong Yu is with the Tencent, AI Lab, Bellevue, WA 98004 USA (E-mail: dyu@global.tencent.com).}}% <-this % stops a space

\markboth{Journal of \LaTeX\ Class Files, January 2025}%
{Shell \MakeLowercase{\textit{et al.}}: A Sample Article Using IEEEtran.cls for IEEE Journals}


\IEEEtitleabstractindextext{

\begin{abstract}
Lifelong learning, also known as continual or incremental learning, is a crucial component for advancing Artificial General Intelligence (AGI) by enabling systems to continuously adapt in dynamic environments. While large language models (LLMs) have demonstrated impressive capabilities in natural language processing, existing LLM agents are typically designed for static systems and lack the ability to adapt over time in response to new challenges. This survey is the first to systematically summarize the potential techniques for incorporating lifelong learning into LLM-based agents. We categorize the core components of these agents into three modules: the perception module for multimodal input integration, the memory module for storing and retrieving evolving knowledge, and the action module for grounded interactions with the dynamic environment. We highlight how these pillars collectively enable continuous adaptation, mitigate catastrophic forgetting, and improve long-term performance. This survey provides a roadmap for researchers and practitioners working to develop lifelong learning capabilities in LLM agents, offering insights into emerging trends, evaluation metrics, and application scenarios. Relevant literature and resources are available at \href{https://github.com/qianlima-lab/awesome-lifelong-llm-agent}{https://github.com/qianlima-lab/awesome-lifelong-llm-agent}.
\end{abstract}

}

\maketitle

%%%%%%%%%%%%%%%%%%%%%%%%%%%%%%%%%%BEGIN%%%%%%%%%%%%%%%%%%%%%%%%%%%%%%%%%%

\section{Introduction}
\label{sec:introduction}

\begin{flushleft}
\leftskip=1cm\emph{``Intelligence is the ability to adapt to change.''} \\
\vspace{.3em}
\leftskip=5cm---\emph{Stephen Hawking}
\end{flushleft}

\begin{figure}[!t]
    \centering
    \includegraphics[width=0.95\linewidth]{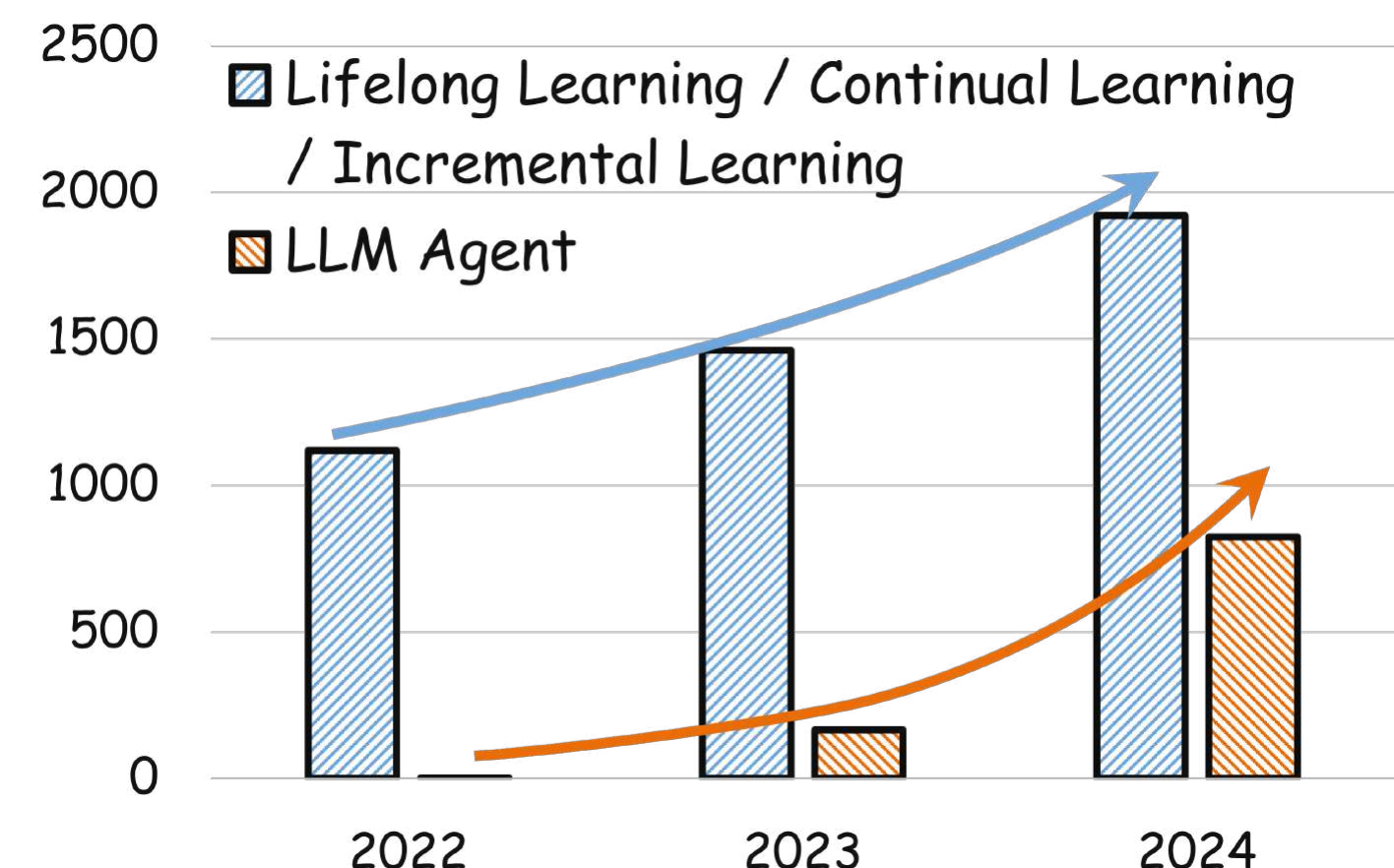}
    \label{fig:research_trend}
    \caption{Number of publications on \emph{lifelong learning} and \emph{LLM Agents} (from Google Scholar). The publications have grown rapidly in recent three years.}
\end{figure}

\IEEEPARstart{L}{ifelong} learning \cite{zheng2024towards,chen2018lifelong}, also known as continual or incremental learning \cite{de2021continual,wang2024comprehensive}, has become a key focus in the development of intelligent systems. As shown in Figure \ref{fig:research_trend}, lifelong learning has attracted increasing research attention in recent years. It plays a crucial role in allowing these systems to continuously adapt and improve over time. As noted by Legg et al. \cite{legg2008machine}, human intelligence is fundamentally about \emph{fast adaptation to a wide range of environments}, highlighting the need for AI systems to exhibit this same level of adaptability. Lifelong learning refers to a system's ability to acquire, integrate, and retain knowledge while avoiding the forgetting of previously learned information. This ability is particularly important for systems that operate in dynamic, complex environments, where new tasks and challenges frequently arise. In contrast to traditional machine learning models, which are typically trained on fixed datasets and optimized for specific tasks, lifelong learning systems are designed to evolve. They accumulate new knowledge and continuously refine their capabilities as they encounter new situations.

Despite its potential, there remains a significant gap between advancements in AI and the practical application of lifelong learning. While humans can naturally integrate new knowledge while retaining the old, current AI systems face two main challenges in lifelong learning: \emph{catastrophic forgetting} \cite{mccloskey1989catastrophic} and \emph{loss of plasticity} \cite{dohare2024loss,abbas2023loss}. These challenges form the \emph{stability-plasticity dilemma} \cite{robins1995catastrophic}. On one hand, catastrophic forgetting occurs when systems forget previously learned information as they learn new tasks, which is particularly problematic when the environment changes. On the other hand, the loss of plasticity refers to the system’s inability to adapt to new tasks or environments. These two issues represent opposing ends of the learning spectrum: static systems avoid forgetting but lack the ability to adapt, while systems focused on adaptation are at risk of forgetting past knowledge. Overcoming this dilemma is key to advancing artificial intelligence and is a foundational challenge on the path to Artificial General Intelligence (AGI) \cite{legg2008machine}.

\begin{figure*}[!t]
    \centering
    \subfloat[Lifelong Learning of LLMs]{
        \includegraphics[width=0.49\linewidth]{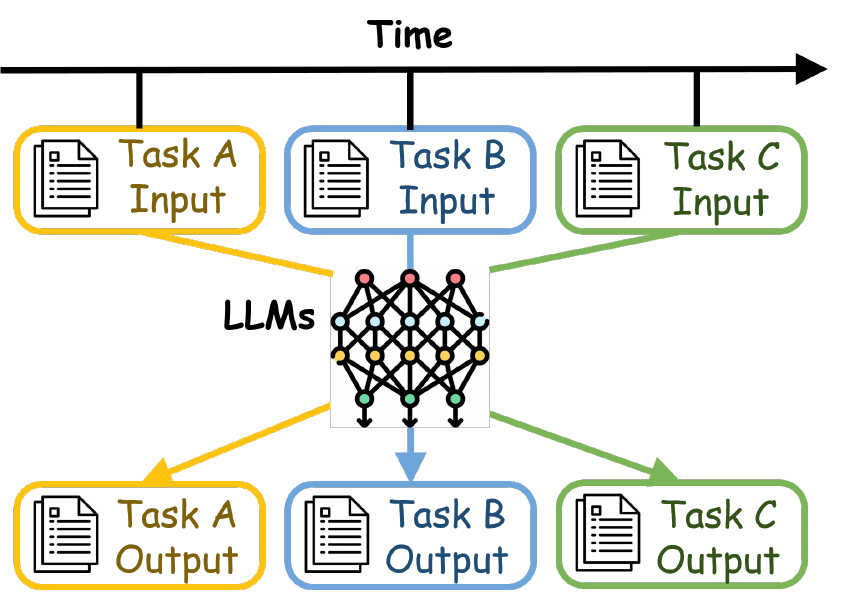}
    }
    \subfloat[Lifelong Learning of LLM Agents]{
        \includegraphics[width=0.49\linewidth]{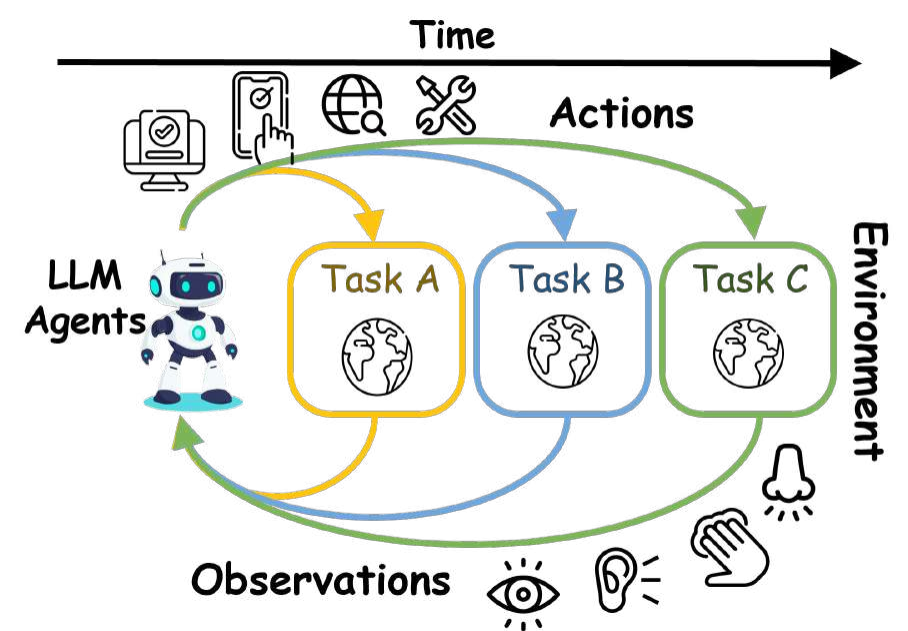}
    }
    \caption{Comparison of lifelong learning between LLMs and LLM Agents. (a) Traditional lifelong learning paradigm of LLMs, where LLMs are viewed as static black-box systems \emph{without feedback from the environment}; (b) the novel lifelong learning paradigm of LLM agents focused on in this survey, where agents \emph{interact with ever-changing environments}. Please refer to Figure \ref{fig:illustration_lifelong_learning_llm_agents} for an illustration.}
    \label{fig:illustration_comparison_llm_agent}
\end{figure*}

\begin{figure}[!t]
    \centering
    \includegraphics[width=0.99\linewidth]{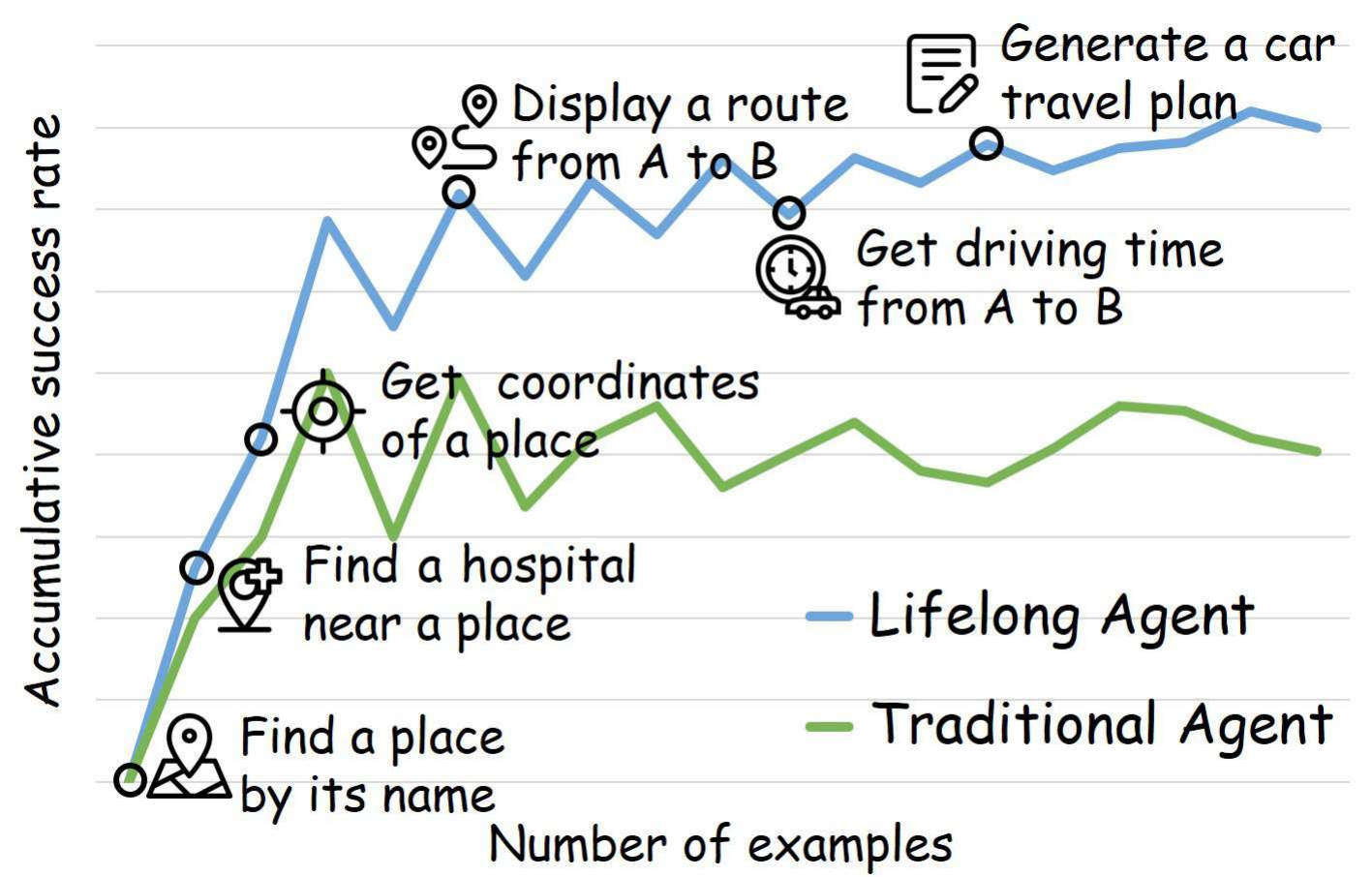}
    \caption{A lifelong LLM agent can adapt to its environment and achieve behavioral evolution through interaction (adapted from AWM \cite{wang2024agent}).}
    \label{fig:action_evolution}
\end{figure}

\subsection{Motivation for Building Lifelong Learning LLM Agents}
\label{sec:introduction:motivation_designing_lifelong_learning_llm_agents}

The recent advancements in large language models (LLMs) \cite{brown2020language, achiam2023gpt} have significantly transformed the field of natural language processing. Models like GPT-4 \cite{achiam2023gpt} are designed to process and generate human-like text by learning from vast amounts of textual data. They excel in tasks such as text generation, machine translation, and question answering due to their ability to understand complex language patterns. However, traditional LLMs \cite{brown2020language, achiam2023gpt} are static after training, meaning they cannot adapt to new tasks or environments once deployed. Their knowledge remains fixed, and they struggle to integrate new information without retraining, limiting their applicability in dynamic real-world scenarios.

In contrast, LLM agents represent a more advanced form of artificial intelligence. Unlike standard LLMs, which process input text and generate output based on prior training, LLM agents \cite{wang2024survey, xi2023rise} are autonomous entities capable of interacting with their environment. These agents can perceive multimodal data (e.g., text, images, sensory data), store this information in memory, and take actions to influence or respond to their surroundings \cite{liu2024autoglm, zhao2024expel, pian2024modality}. Designed to continuously adapt to new contexts, LLM agents learn from their interactions and experiences, improving their decision-making capabilities over time. Illustrations are provided in Figure \ref{fig:illustration_comparison_llm_agent} and Figure \ref{fig:action_evolution}.

The motivation for incorporating lifelong learning into LLM agents arises from the need to develop intelligent systems that can not only adapt to new tasks but also retain and apply prior knowledge across a wide range of dynamic environments, aligning with Legg et al.'s \cite{legg2008machine} view of intelligence as \emph{fast adaptation to a wide range of environments}. Currently, existing LLM agents are typically developed as static systems, limiting their ability to evolve in response to new challenges. Moreover, most lifelong learning research on LLMs \cite{wang2024comprehensive, zheng2024towards} focuses on handling ever-changing data distributions without interacting with an environment. For example, continual fine-tuning of LLMs to adapt to instructions from specific domains \cite{zheng2024towards}. However, these approaches still treat LLMs as static black-box systems and do not address the practical need for LLMs to learn interactively within real-world environments. Figure \ref{fig:illustration_comparison_llm_agent} compares the traditional lifelong learning paradigm with the novel paradigm of LLM agents that interact with dynamic environments, as discussed in this survey.

\begin{figure*}[!t]
    \centering
    \includegraphics[width=0.99\linewidth]{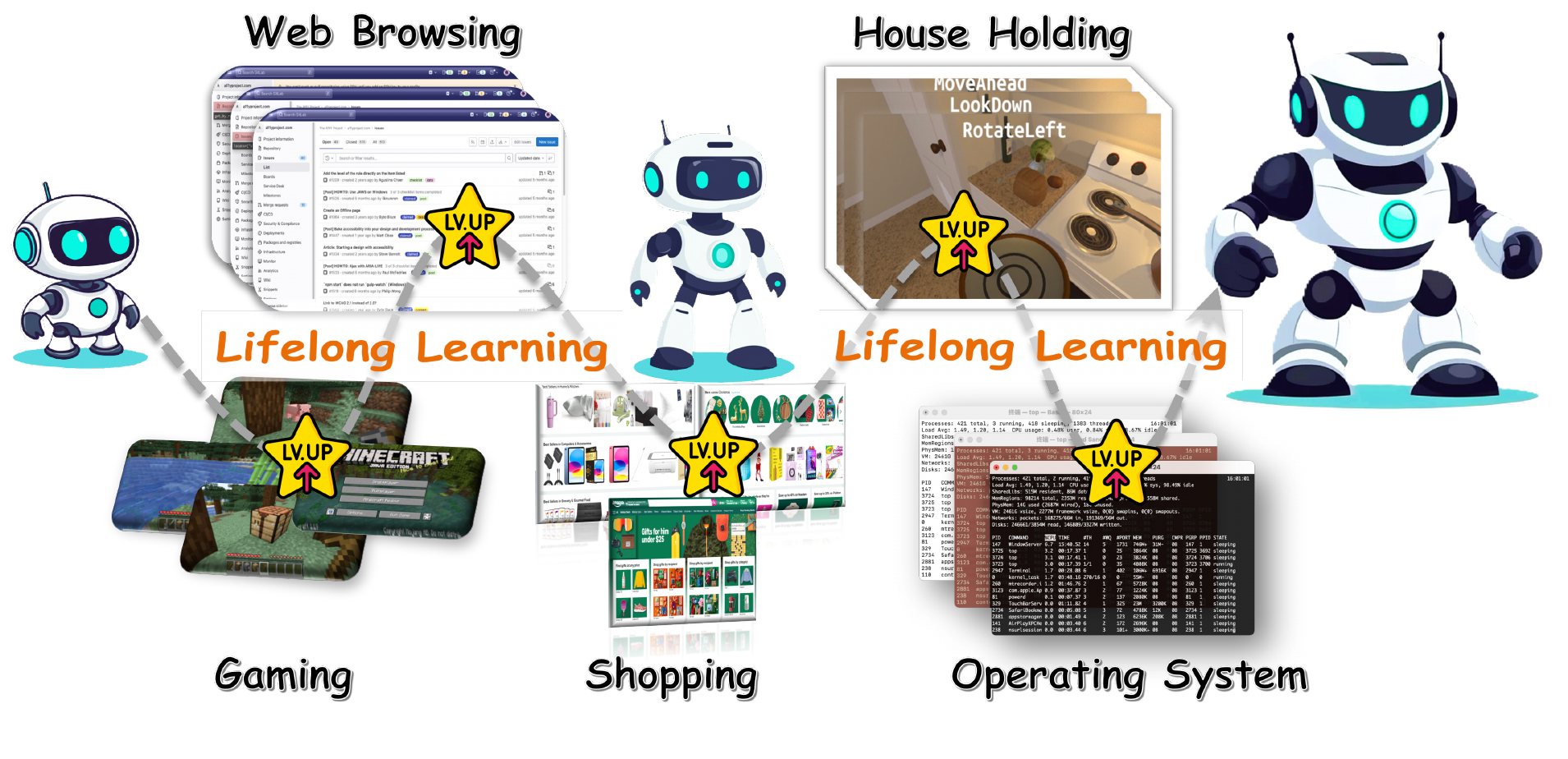}
    % \vspace{-2.5em}
    \caption{Illustration of lifelong learning in large language model-based agents. In real-world applications, LLM agents are expected to adapt to various environments such as gaming, web browsing, shopping, household tasks, and operating systems without the need to design environment-specific agents for every new environment.}
    \label{fig:illustration_lifelong_learning_llm_agents}
\end{figure*}

In real-world applications, LLM agents are expected to adapt to diverse environments such as gaming, web browsing, shopping, household tasks, and operating systems without the need to design separate agents for each new context. By incorporating lifelong learning capabilities, these agents can overcome such limitations. They continuously learn and store knowledge from multiple modalities (e.g., visual, textual, sensory data), enabling real-time adaptation and decision-making as environments change \cite{jang2024videowebarena, koh2024visualwebarena, liu2024visualagentbench, zhou2024webarena}. Integrating lifelong learning into LLM agents unlocks their full potential for dynamic real-world applications \cite{liu2023agentbench, drouin2024workarena}. Consequently, these agents can evolve continuously, acquire new knowledge, and preserve critical information, enhancing their adaptability and versatility. This ongoing learning process is essential for environments where new challenges regularly emerge, such as in autonomous robotics, interactive assistants, and adaptive decision-support systems \cite{xi2023rise}. An illustration of a lifelong learning LLM agent is provided in Figure \ref{fig:illustration_lifelong_learning_llm_agents}.

\subsection{Scope of the Survey}
\label{sec:introduction:scope_of_the_survey}

This survey provides a comprehensive overview of the key concepts, techniques, and challenges involved in developing lifelong learning systems for LLM-based agents. As the first survey to systematically summarize potential techniques for lifelong learning in LLM agents, it addresses the following research questions (RQs):

\begin{enumerate}[leftmargin=1cm, label=\textbf{RQ\arabic{*}}:]
    \item What are the core concepts, development processes, and fundamental architectures of LLM agents designed for lifelong learning? (Section~\ref{sec:building_lifelong_learning_llm_agents})
    \item How can LLM agents continuously perceive and process single-modal and multi-modal data to adapt to new environments and tasks? (Sections~\ref{sec:single_modal_perception}, \ref{sec:multi_modal_perception})
    \item What strategies can mitigate catastrophic forgetting and enable the retention of previously learned knowledge? (Sections~\ref{sec:working_memory}, \ref{sec:episodic_memory}, \ref{sec:semantic_memory}, \ref{sec:parametric_memory})
    \item How can LLM agents perform various actions, such as grounding, retrieval, and reasoning, in dynamic environments? (Sections~\ref{sec:grounding_actions}, \ref{sec:retrieval_actions}, \ref{sec:reasoning_actions})
    \item What evaluation metrics and benchmarks are used to assess the performance of lifelong learning in LLM agents? (Section~\ref{sec:evaluation_of_lifelong_llm_agents})
    \item What are the real-world applications and use cases of lifelong learning LLM agents, and how do they benefit from continuous adaptation? (Section~\ref{sec:application_of_lifelong_llm_agents})
    \item What are the key challenges, limitations, and open problems in the development of lifelong learning for LLM-based agents? (Section~\ref{sec:practical_insights_and_future_directions})
\end{enumerate}

By addressing these research questions, this survey serves as a step-by-step guide for understanding the design, challenges, and applications of lifelong learning in LLM agents. It reviews state-of-the-art techniques and highlights emerging trends and future directions for research in this rapidly evolving field.

\subsection{Contribution of the Survey}
\label{sec:introduction:contribution_of_the_survey}

To the best of our knowledge, this is the \emph{first} survey that systematically reviews the latest advancements at the intersection of lifelong learning and LLM agents. The main contributions of this survey are as follows:

\begin{itemize}[label=$\blacksquare$]
    \item \textbf{Foundational Overview:} Provides a thorough overview of the foundational concepts and architectures essential for implementing lifelong learning in LLM agents.
    \item \textbf{In-Depth Component Analysis:} Examines key components, including perception, memory, and action modules, that enable adaptive behavior in LLM agents.
    \item \textbf{Comprehensive Discussion:} Discusses real-world applications, evaluation metrics, benchmarks, as well as key challenges and future research directions in the domain of lifelong learning for LLM agents.
\end{itemize}

\subsection{Survey Structure} 
\label{sec:introduction:structure}

This survey is organized as follows. \textbf{Section \ref{sec:related_work}} reviews related surveys and literature on LLM agents and lifelong learning. \textbf{Section \ref{sec:building_lifelong_learning_llm_agents}} introduces the foundational concepts, development processes, and overall architecture of LLM agents designed for lifelong learning. \textbf{Sections \ref{sec:single_modal_perception} and \ref{sec:multi_modal_perception}} discuss the design of lifelong learning LLM agents from a perceptual perspective, focusing on single-modal and multi-modal approaches, respectively. \textbf{Sections \ref{sec:working_memory}, \ref{sec:episodic_memory}, \ref{sec:semantic_memory}, and \ref{sec:parametric_memory}} examine the design of LLM agents from a memory perspective, covering working memory, episodic memory, semantic memory, and parametric memory. \textbf{Sections \ref{sec:grounding_actions}, \ref{sec:retrieval_actions}, and \ref{sec:reasoning_actions}} explore the design of LLM agents from the action perspective, including grounding actions, retrieval actions, and reasoning actions. \textbf{Section \ref{sec:evaluation_of_lifelong_llm_agents}} presents evaluation metrics and benchmarks for assessing lifelong learning in LLM agents. \textbf{Section \ref{sec:application_of_lifelong_llm_agents}} delves into real-world applications and use cases of lifelong learning LLM agents. \textbf{Section \ref{sec:practical_insights_and_future_directions}} offers practical insights and outlines future research directions. Finally, \textbf{Section \ref{sec:conclusion}} concludes the survey.

\section{Related Work}
\label{sec:related_work}

\renewcommand{\multirowsetup}{\centering}
\begin{table*}[htbp]
  \centering
  \caption{Summary of the connections and differences between our work and existing surveys. "Scope": \emph{LLMs} means large language models, \emph{LL} means lifelong learning, \emph{NLP} means natural language processing. A {\color[RGB]{3,191,61}{\Checkmark}} symbol indicates that a survey explicitly addresses a given domain.}
    \resizebox{\linewidth}{!}{
        \begin{tabular}{lp{36.28em}cccc}
    \toprule
    \multicolumn{1}{c}{\multirow{2}[4]{*}{Survey}} & \multicolumn{1}{c}{\multirow{2}[4]{*}{Contribution}} & \multicolumn{4}{c}{Scope} \\
\cmidrule{3-6}          & \multicolumn{1}{c}{} & LLMs & LL & NLP & Agents \\
    \midrule
    Zheng et al. \cite{zheng2024towards} & \multicolumn{1}{l}{A survey on lifelong learning methods for LLMs.} & \color[RGB]{3,191,61}{\Checkmark} & \color[RGB]{3,191,61}{\Checkmark} & \color[RGB]{3,191,61}{\Checkmark} &  \\
    \rowcolor[rgb]{ .949,  .949,  .949} Wu et al. \cite{wu2024continual} & \multicolumn{1}{l}{A survey on continual learning stage for LLMs.} & \color[RGB]{3,191,61}{\Checkmark} & \color[RGB]{3,191,61}{\Checkmark} &       &  \\
    Shi et al. \cite{shi2024continual} & \multicolumn{1}{l}{A survey on LLMs within the context of continual learning.} & \color[RGB]{3,191,61}{\Checkmark} & \color[RGB]{3,191,61}{\Checkmark} &       &  \\
    \rowcolor[rgb]{ .949,  .949,  .949} Ke and Liu\cite{ke2022continual} & \multicolumn{1}{l}{A survey on continual learning of natural language processing tasks.} &       & \color[RGB]{3,191,61}{\Checkmark} & \color[RGB]{3,191,61}{\Checkmark} &  \\
    Zhou et al. \cite{zhou2024continual} & \multicolumn{1}{l}{A survey on continual Learning with pre-trained models.} & \color[RGB]{3,191,61}{\Checkmark} & \color[RGB]{3,191,61}{\Checkmark} &       &  \\
    \rowcolor[rgb]{ .949,  .949,  .949} Wang et al. \cite{wang2024comprehensive} & \multicolumn{1}{l}{A survey on settings, foundations, methods and applications of continual learning.} &       & \color[RGB]{3,191,61}{\Checkmark} &       &  \\
    Biesialska et al. \cite{biesialska2020continual} & \multicolumn{1}{l}{A survey on continual lifelong learning in natural language processing.} &       & \color[RGB]{3,191,61}{\Checkmark} & \color[RGB]{3,191,61}{\Checkmark} &  \\
    \rowcolor[rgb]{ .949,  .949,  .949} Parisi et al. \cite{parisi2019continual} & \multicolumn{1}{l}{A review on continual lifelong learning with neural networks.} &       & \color[RGB]{3,191,61}{\Checkmark} &       &  \\
    Wang et al. \cite{wang2024survey} & \multicolumn{1}{l}{A survey on LLM based autonomous agents.} & \color[RGB]{3,191,61}{\Checkmark} &       &       & \color[RGB]{3,191,61}{\Checkmark} \\
    \rowcolor[rgb]{ .949,  .949,  .949} Xi et al. \cite{xi2023rise} & \multicolumn{1}{l}{A survey on LLM based agents.} & \color[RGB]{3,191,61}{\Checkmark} &       &       & \color[RGB]{3,191,61}{\Checkmark} \\
    Li et al. \cite{li2024personal} & \multicolumn{1}{l}{A survey on personal large language model agents.} & \color[RGB]{3,191,61}{\Checkmark} &       &       & \color[RGB]{3,191,61}{\Checkmark} \\
    \rowcolor[rgb]{ .949,  .949,  .949} Cheng et al. \cite{cheng2024exploring} & \multicolumn{1}{l}{An overview of LLM based intelligent agents within single-agent and multi-agent systems.} & \color[RGB]{3,191,61}{\Checkmark} &       &       & \color[RGB]{3,191,61}{\Checkmark} \\
    Li et al. \cite{li2024survey} & \multicolumn{1}{l}{A survey on LLM based multi-agent systems.} & \color[RGB]{3,191,61}{\Checkmark} &       &       & \color[RGB]{3,191,61}{\Checkmark} \\
    \rowcolor[rgb]{ .949,  .949,  .949} Zhang et al. \cite{zhang2024large} & \multicolumn{1}{l}{A survey on LLM-brained GUI agents.} & \color[RGB]{3,191,61}{\Checkmark} &       &       & \color[RGB]{3,191,61}{\Checkmark} \\
    \midrule
    \textbf{Our Work} & \textbf{A survey on lifelong learning of LLM based agents.} & \color[RGB]{3,191,61}{\Checkmark} & \color[RGB]{3,191,61}{\Checkmark} & \color[RGB]{3,191,61}{\Checkmark} & \color[RGB]{3,191,61}{\Checkmark} \\
    \bottomrule
    \end{tabular}%
    }
  \label{tab:memory_relatedwork}%
\end{table*}%

\subsection{Surveys on Lifelong Learning}
\label{sec:related_work:surveys_on_lifelong_learning}
In recent years, the rapid growth of lifelong learning has received a great deal of academic attention. In order to summarize the research in this area, researchers have written a large number of surveys. In this subsection, we will briefly review these surveys and summarize their main contributions in order to better highlight our research, as illustrated in Table \ref{tab:memory_relatedwork}.

Some surveys involve presenting research advances in lifelong learning in the field of natural language processing. For instance, Zheng et al. \cite{zheng2024towards} systematically summarize lifelong learning methods for LLMs for the first time by considering them from the perspective of 12 scenarios, which are classified into two categories: internal and external knowledge. In the section on internal knowledge, this survey presents scenarios about continual finetuning in natural language processing. Ke and Liu \cite{ke2022continual} first introduce the settings and learning modes of continual learning and summarize the natural language processing problems in continual learning. Then, they propose that existing continual learning techniques are mainly used to solve the two challenges of catastrophic forgetting and knowledge transfer. On this basis, this survey analyzes approaches for catastrophic forgetting prevention and knowledge transfer. Besides, Biesialska et al. \cite{biesialska2020continual} classify continual learning methods into three main categories: rehearsal, regularization, and architectural as well as a few hybrid categories. This survey also explores the application of continual learning to several tasks in natural language processing, including question answering, sentiment analysis, and text classification.

Other surveys present progress in lifelong learning from a variety of perspectives. For instance, Wu et al. \cite{wu2024continual} describe the different training stages of LLMs, including continual pre-training, continual instruction tuning, and continual alignment. In addition, this survey presents benchmarks and evaluations on continual learning. Shi et al. \cite{shi2024continual} present the general picture of continually learning LLMs from vertical continuity and horizontal continuity. In addition, this survey introduces the evaluation protocols and datasets for continually learning large language models. Starting from the motivation of solving the learning problem, Zhou et al. \cite{zhou2024continual} group pre-trained models-based continual learning studies into three categories, i.e., prompt-based methods, representation-based methods, and model mixture-based methods. In addition, this survey analyzes the experimental results of ten algorithms of the above three categories of methods on seven benchmark datasets. Wang et al. \cite{wang2024comprehensive} classify continual learning methods into five main categories: regularization-based approach, replay-based approach, optimization-based approach, representation-based approach and architecture-based approach, where each category is analyzed in detail for its typical implementations and empirical properties. From the biological perspective of lifelong learning, Parisi et al. \cite{parisi2019continual} analyze lifelong learning and catastrophic forgetting in neural networks. In addition, they summarize well-established and emerging neural network approaches driven by interdisciplinary research introducing findings from neuroscience, psychology, and cognitive sciences for the development of lifelong learning autonomous agents.

However, prior to this paper, little work has focused specifically on the rapidly emerging and highly promising field about lifelong learning of LLM-based Agents. In this survey, we collate the extensive literature about lifelong learning of LLM-based Agents, covering their construction, application, and evaluation processes.

\subsection{Surveys on Large Language Model-based Agent}
\label{sec:related_work:survyes_on_large_language_model_based_agent}
With the boom of LLM-based Agents, a wide variety of surveys have emerged that systematically summarize the research in this area. In this subsection, we briefly review these surveys, presenting their main contributions, as illustrated in Table \ref{tab:memory_relatedwork}.

From the perspective of agent architecture design, Wang et al. \cite{wang2024survey} propose a unified framework including four parts, namely profiling module, memory module, planning module and action module for the first time. In addition, this survey summarizes the applications, evaluations and challenges of LLM-based autonomous agent. Similarly, Xi et al. \cite{xi2023rise} introduce the three main components of the agent from the perspective of construction: brain, perception and action. In addition, this survey also investigates the practical applications of agents in different scenarios and the personal and social behaviors of agents.

Additionally, different focus and categorization produce a diverse understanding of the field. Some other papers review specific aspects of LLM-based agents. For example, Li et al. \cite{li2024personal} present the background and technological advances of intelligent personal assistants and introduce the concept of personal LLM agents. This survey delves into the issues of fundamental capabilities and efficiency of personal LLM agents. Finally, it analyzes the challenges regarding security and privacy. Cheng et al. \cite{cheng2024exploring} consider the LLM-based agent system framework and analyze the single agent system and multi-agent system in detail. Similarly, this survey discusses the performance evaluation and prospect applications of LLM-based agent. Li et al. \cite{li2024survey} propose a unified framework of the general multi-agent system, which includes five modules, namely profile, perception, self-action, mutual interaction and evolution. In addition, this survey also analyzes the application and some key open issues of multi-agent systems. Besides, Zhang et al. \cite{zhang2024large}define the concept of LLM-brained GUI agents and analyze the architecture and progress of LLM-brained GUI agents.

These surveys analyze important progress in building LLM-based agents, covering everything from personal LLM agents to multi-agent systems. However, they fall short in their connection to lifelong learning. Lifelong learning emphasizes the ability of agents to continuously learn and adapt in changing environments, whereas these surveys mainly focus on current technology implementations and application scenarios. In contrast to them, our survey provides an in-depth summary of how to equip LLM-based agents with the ability to learn and evolve in the long term, and how to optimize their performance as they continue to accumulate experience.

\section{Building Lifelong Learning LLM Agents}
\label{sec:building_lifelong_learning_llm_agents}

We begin by presenting the formal definition of lifelong learning in LLM agents, followed by an overview of its historical development, and conclude with a detailed description of the overall architecture for lifelong learning in LLM agents.

\subsection{Formal Definition of Lifelong Learning for LLM-based Agents}
\label{sec:building_lifelong_learning_llm_agents:formal_definition_of_lifelong_learning_for_llm_based_agents}

\begin{definition}[Environment of LLM Agents]\label{def:environment}
We model the environment of an LLM-based agent as a goal-conditional partially observable Markov decision process (POMDP). Formally, a POMDP is defined as an 8-tuple:
\begin{equation}
\mathcal{E} = (\mathcal{S}, \mathcal{A}, \mathcal{G}, T, R, \Omega, O, \gamma),
\end{equation}
where:
\begin{itemize}[]
    \item $\mathcal{S}$ is a set of states. Each $s \in \mathcal{S}$ can include multimodal information such as textual descriptions, images, or structured data (e.g., a product page on an e-commerce website containing text, images, and product specifications).
    \item $\mathcal{A}$ is a set of actions. Each $a \in \mathcal{A}$ represents an instruction or command that the agent can issue, often expressed in natural language (e.g., “add this item to the cart”).
    \item $\mathcal{G}$ is a set of possible goals. Each $g \in \mathcal{G}$ specifies a particular objective (e.g., “purchase a laptop”).
    \item $T(s' \mid s,a)$ is the state transition probability function. For each state-action pair $(s,a)$, $T$ defines a probability distribution over next states $s' \in \mathcal{S}$. For example, if the agent clicks on a product link, $T$ models the probability of transitioning to the product’s detail page.
    \item $R : \mathcal{S} \times \mathcal{A} \times \mathcal{G} \to \mathcal{R}$ is the goal-conditional reward function. For each triplet $(s,a,g)$, $R$ may return a numeric value or textual feedback (e.g., “Good job!”) indicating how well the action $a$ taken in state $s$ advances the objective $g$. This allows the environment to interactively provide user feedback aligned with the chosen goal.
    \item $\Omega$ is a set of observations. Each $o' \in \Omega$ can be textual, visual, or a combination thereof. Observations represent the agent’s partial view of the underlying state (e.g., the content visible on a webpage).
    \item $O(o' \mid s',a)$ is the observation probability function. Given that the environment transitioned to state $s'$, and action $a$ was taken, $O$ defines the probability of receiving observation $o'$. For instance, upon navigating to a product page, $O$ might model the probability of observing a particular product image and description.
    \item $\gamma \in [0,1)$ is the discount factor, which balances immediate versus long-term rewards. In the context of LLM agents, the discount factor is used only when the reward is numeric.
\end{itemize}

This framework captures the complexities of real-world scenarios. For instance, a virtual shopping assistant (the agent) interacts with a complex, multimodal website (the environment) to achieve a goal such as completing a purchase. The assistant receives partial observations (e.g., product listings), takes actions (e.g., clicking a link), and obtains feedback (numeric or textual) reflecting progress toward the goal.

\begin{definition}[LLM-based Agent]\label{def:agent}
An LLM-based agent is an agent whose policy and decision-making process rely on a large language model. The agent interacts with an environment $\mathcal{E} = (\mathcal{S}, \mathcal{A}, \mathcal{G}, T, R, \Omega, O, \gamma)$ as defined above. The agent's policy $\pi$ is a mapping from its observations to actions, where $\pi(o_t) \in \mathcal{A}$ represents the action selected at each time step based on the observation $o_t \in \Omega$. At each time step $t$:
\begin{itemize}
    \item The agent receives an observation $o_t \in \Omega$, which may include text, images, or other structured data.
    \item The agent selects an action $a_t \in \mathcal{A}$, typically by generating a textual command or query through its LLM-based policy $\pi(o_t)$.
    \item The environment returns a reward $r_t = R(s_t, a_t, g)$, potentially numeric or textual, guiding the agent towards achieving the goal $g$.
\end{itemize}
\end{definition}

The agent’s policy $\pi$ maps histories of observations and actions to a probability distribution over $\mathcal{A}$. By employing language understanding and generation capabilities, the agent can handle complex, real-world tasks where purely numeric feedback is insufficient.
\end{definition}

\begin{definition}[Task]\label{def:task}
A task $\mathcal{T}^{(i)}$ is given by:
\begin{equation}
\mathcal{T}^{(i)} = \langle \mathcal{E}^{(i)}, o_0^{(i)}, g^{(i)} \rangle,
\end{equation}
where:
\begin{itemize}
    \item $\mathcal{E}^{(i)} = (\mathcal{S}^{(i)}, \mathcal{A}^{(i)}, \mathcal{G}^{(i)}, T^{(i)}, R^{(i)}, \Omega^{(i)}, O^{(i)}, \gamma^{(i)})$ is the environment of the $i$-th task.
    \item $o_0^{(i)} \in \Omega^{(i)}$ is the initial observation. For example, the agent might start on a homepage of an e-commerce site.
    \item $g^{(i)} \in \mathcal{G}^{(i)}$ is the specific goal for this task (e.g., “buy a laptop under \$1000”).
\end{itemize}

To solve $\mathcal{T}^{(i)}$, the agent must select actions that lead from an initial observation toward states fulfilling $g^{(i)}$, with rewards computed according to $R^{(i)}(s,a,g^{(i)})$.
\end{definition}

\begin{definition}[Trajectory]\label{def:trajectory}
For a given task $\mathcal{T}^{(i)}$, a trajectory of length $T_i$ is:
\begin{equation}
\xi_{T_i}^{(i)} = \langle o_0, a_0, r_0, o_1, a_1, r_1, \dots, o_{T_i}, a_{T_i}, r_{T_i} \rangle,
\end{equation}
where $o_t \in \Omega^{(i)}$, $a_t \in \mathcal{A}^{(i)}$, and $r_t = R^{(i)}(s_t,a_t,g^{(i)})$. Each trajectory captures one sequence of interactions, starting from the initial observation and continuing until termination.
\end{definition}

\begin{definition}[Trial]\label{def:trial}
A \emph{trial} is one complete attempt by the agent to solve a given task $\mathcal{T}^{(i)}$. Each trial corresponds to one trajectory $\xi_{T_i}^{(i)}$. In a real-world setting, a trial might represent one entire session of trying to complete a purchase. Multiple trials may differ due to stochastic environment dynamics or policy variability.
\end{definition}

\begin{definition}[Trajectory-Level and Step-Level Rewards]\label{def:reward_levels}
Rewards can be offered at different granularities:
\begin{itemize}[label=$\blacksquare$]
    \item \emph{Trajectory-Level Rewards}: Only the final outcome yields informative feedback. For example, the agent receives a positive reward (or a textual confirmation) only upon successfully completing the purchase.
    \item \emph{Step-Level Rewards}: Intermediate actions also yield meaningful feedback (e.g., navigating to a more relevant product page might produce a positive textual hint).
\end{itemize}

Because $R^{(i)} : \mathcal{S}^{(i)} \times \mathcal{A}^{(i)} \times \mathcal{G}^{(i)} \to \mathcal{R}^{(i)}$, each reward depends on the current goal $g^{(i)}$. In practical applications, this allows the environment to provide goal-aligned feedback, guiding the agent’s actions in pursuit of that objective.
\end{definition}

\begin{definition}[Lifelong Learning Tasks]\label{def:lifelong_tasks}
Lifelong learning considers a set of tasks:
\begin{equation}
\mathcal{U} = \{\mathcal{T}^{(1)}, \mathcal{T}^{(2)}, \dots, \mathcal{T}^{(n)}\}.
\end{equation}
Tasks may vary in terms of states, actions, goals, and rewards. Lifelong learning can be:
\begin{itemize}[label=$\blacksquare$]
    \item \emph{Intra-environment}: All tasks share the same underlying environment structure, but differ in initial conditions or goals (e.g., different products to purchase on the same website).
    \item \emph{Inter-environment}: Tasks arise from distinct environments, requiring the agent to adapt and transfer knowledge across different domains (e.g., switching from an e-commerce site to a travel booking platform).
\end{itemize}

In real-world scenarios, agents continuously face new tasks and must accumulate knowledge without forgetting previous solutions, improving efficiency and competence over time.
\end{definition}

\begin{definition}[Objective of Lifelong Learning]\label{def:objective}
Define a numeric mapping $\tilde{R}^{(i)}(o_t,a_t,g^{(i)})$ that converts the (possibly textual) reward $R^{(i)}(s_t,a_t,g^{(i)})$ into a scalar value. The performance of a policy $\pi$ on task $\mathcal{T}^{(i)}$ is the expected cumulative reward:
\begin{equation}
J(\mathcal{T}^{(i)},\pi) = \mathbb{E}_{\xi_{T_i}^{(i)} \sim \pi}\left[\sum_{t=0}^{T_i} \tilde{R}^{(i)}(o_t,a_t,g^{(i)})\right].
\end{equation}

The objective of lifelong learning is to find a policy $\pi$ that maximizes the expected performance across all tasks in $\mathcal{U}$:
\begin{equation}
\max_{\pi} \sum_{i=1}^n J(\mathcal{T}^{(i)},\pi).
\end{equation}

This formulation encourages the agent to improve continuously, leveraging past experience and knowledge to excel at current and future tasks, much like a human learner facing increasingly challenging goals.
\end{definition}

\begin{figure*}[!t]
    \centering
    \includegraphics[width=0.99\linewidth]{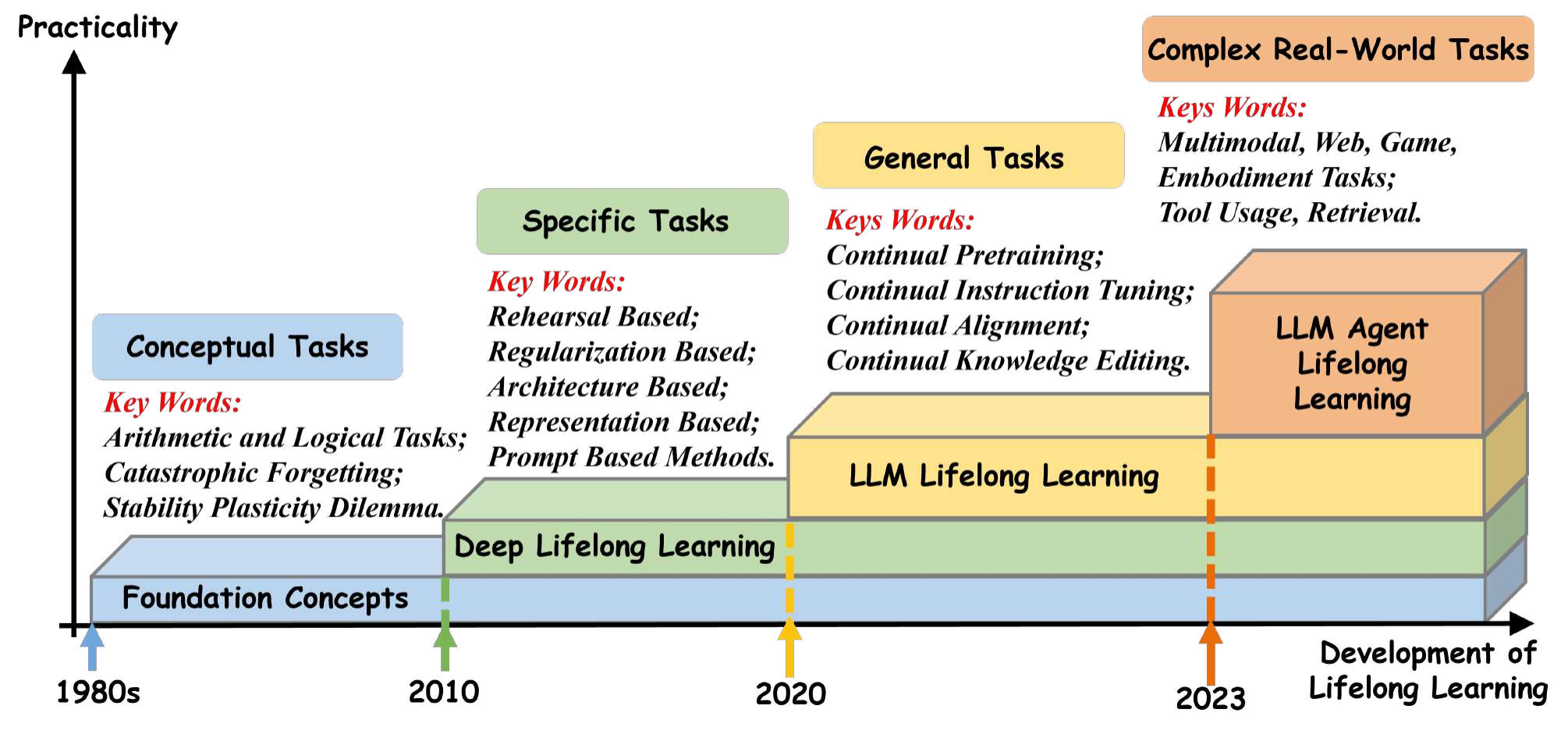}
    \caption{Development of lifelong learning for AI systems, highlighting four key stages: (1) Establishment of foundational concepts starting in the 1980s, (2) Advancements in deep lifelong learning from 2010 to the present, (3) Integration of lifelong learning into large language models from 2020 onwards, and (4) the latest developments in lifelong learning for LLM agents. The practicality of lifelong learning has significantly improved alongside its development, enabling more versatile and adaptive AI systems in diverse real-world applications.}
    \label{fig:history_lifelong_learning}
\end{figure*}

\subsection{Background and History of Lifelong Learning for AI Systems}

Lifelong learning, also referred to as continual or incremental learning, is grounded in the idea that intelligent systems should continually acquire, refine, and retain knowledge over extended periods—much like humans. Unlike traditional machine learning approaches that assume access to a fixed, stationary dataset, lifelong learning frameworks confront the reality that data and tasks evolve over time, and that models must adapt without forgetting previously mastered skills. A illustration of the development of lifelong learning is provided in Figure \ref{fig:history_lifelong_learning}.

\textbf{Human and Neuroscience Perspectives:}  
The principles of lifelong learning draw inspiration from human cognitive development. Humans do not train on a fixed dataset; instead, we accumulate knowledge from diverse and changing experiences \cite{carpenter1988art}. Memory consolidation in the human brain, involving complex interactions between the hippocampus and neocortex, ensures that new learning does not completely overwrite old memories. Studies of synaptic plasticity and learning in neuroscience help inform algorithms that attempt to preserve and integrate previously acquired representations while adapting to new information. Incorporating such principles into neural networks has been a longstanding challenge and motivation for research in lifelong learning \cite{french1999catastrophic, mcclelland1995there, hassabis2017neuroscience}.

\subsubsection{Foundation Concepts (1980s–Present)}

The origins of lifelong learning research can be traced back to early adaptive control and incremental learning studies in the 1980s and 1990s. Researchers recognized that many real-world environments are non-stationary: the distribution of data changes over time, and models must adapt accordingly. During this period, foundational concepts such as catastrophic forgetting and the stability-plasticity dilemma were first identified. Catastrophic forgetting describes the tendency of neural networks to lose previously acquired knowledge when trained on new data \cite{mccloskey1989catastrophic, ratcliff1990connectionist}, while the stability-plasticity dilemma \cite{carpenter1988art,grossberg1987competitive} addresses the need to balance retaining old knowledge (stability) with acquiring new information (plasticity).

Throughout the 1990s and into the 2000s, catastrophic forgetting became a central research focus. Early attempts to mitigate forgetting relied on replaying previously encountered examples, adjusting network architectures to accommodate new tasks, or using heuristics to preserve existing representations \cite{robins1995catastrophic}. As neural networks became more widely adopted across domains, more sophisticated methods for lifelong learning were developed. Researchers introduced structured approaches, such as dynamically expanding architectures that allocate new network capacity for novel tasks, dual-memory frameworks inspired by the interplay between short-term and long-term memory in the brain, and early forms of regularization and parameter isolation techniques to reduce interference with older knowledge \cite{thrun1998lifelong, parisi2019continual}.

\subsubsection{Deep Lifelong Learning (2010–Present)}

The rise of deep learning in the early 2010s significantly expanded the capabilities of neural networks. However, these initial breakthroughs often assumed a stationary training dataset. Incorporating incremental learning into deep networks required new strategies. Techniques like Elastic Weight Consolidation (EWC) \cite{kirkpatrick2017overcoming} and Learning without Forgetting (LwF) \cite{li2017learning} became popular, using careful parameter adjustments to preserve previously learned representations. These methods \cite{chen2018lifelong} leveraged deep learning’s powerful representational ability while addressing the core issue of catastrophic forgetting, pushing lifelong learning closer to practical real-world applicability.

During this period, various approaches were categorized based on their methodologies: 
\emph{Rehearsal-Based Methods:} Involve replaying or retraining on a subset of previously seen data to retain past knowledge \cite{rolnick2019experience}. 
\emph{Regularization-Based Methods:} Apply constraints or penalties to the loss function to prevent significant changes to important parameters \cite{kirkpatrick2017overcoming}.
\emph{Architecture-Based Methods:} Dynamically modify the network architecture to accommodate new tasks without interfering with existing ones \cite{mallya2018packnet}.
\emph{Representation-Based Methods:} Focus on learning robust and transferable representations that facilitate knowledge retention and transfer \cite{mehta2023empirical}.
\emph{Prompt-Based Methods:} Utilize prompts or auxiliary inputs to guide the model in retaining and utilizing past knowledge \cite{razdaibiedinaprogressive}.

\subsubsection{LLM Lifelong Learning (2020–Present)}

With the advent of large-scale pre-training, especially in language models, the landscape of AI has been reshaped. LLMs like GPT-3 \cite{brown2020language} introduced context-aware word representations, enabling models to perform a wide range of Natural Language Processing (NLP) tasks with high efficiency. Lifelong learning in LLMs initially focused on conventional NLP tasks such as text classification, machine translation, and instruction following. Techniques developed during this period include parameter-efficient fine-tuning \cite{houlsby2019parameter, brown2020language}, retrieval-augmented learning \cite{lewis2020retrieval}, and prompt-based adaptation \cite{brown2020language}. These approaches allow LLMs to continuously integrate new linguistic knowledge and adapt to evolving language patterns without extensive retraining, thereby enhancing their performance on established NLP benchmarks.

Key applications and methods in this period include:
\emph{Continual Pretraining:} Continual pretraining of models on domain-specific corpora or newly available datasets, such as Wikipedia, to incorporate the most up-to-date knowledge or adapt to additional languages \cite{jin2022lifelong}.
\emph{Continual Instruction Tuning:} Incremental fine-tuning of the model to improve its ability to follow a diverse set of instructions, enhancing its performance across various tasks such as summarization, translation, and instruction following \cite{luo2023empirical, zhang2023citb}.
\emph{Continual Alignment:} Ensuring that the model’s outputs remain aligned with human values, ethical guidelines, and user preferences as it learns new tasks, thereby maintaining trustworthiness and relevance \cite{shen2023large}.
\emph{Continual Knowledge Editing:} Addressing outdated or incorrect information within LLMs, effectively mitigating the risk of hallucinated or inaccurate outputs \cite{jang2021towards, wang2023knowledge}.

\subsubsection{LLM-based Agents Lifelong Learning (2023–Present)} 

Starting around 2023, the focus of lifelong learning has expanded from conventional NLP tasks to more realistic and complex applications embodied by LLM-based agents. Unlike LLMs, which primarily handle tasks such as text generation and classification, LLM-based agents are designed to interact with dynamic environments and perform intricate tasks like online shopping, household management, operating system operations, and more. These agents require advanced lifelong learning capabilities to manage multimodal inputs, execute sequential decision-making processes, and maintain coherent performance across diverse and evolving tasks.

Key advancements in this period include:
\emph{Dynamic Task Adaptation:} Developing models that can seamlessly switch between varied tasks without compromising performance on previously learned ones \cite{liu2023agentbench, zhou2024webarena, drouin2024workarena}.
\emph{Multimodal Integration:} Enhancing agents' abilities to process and integrate information from multiple modalities (e.g., text, images, and sensor data) to perform complex real-world tasks \cite{jang2024videowebarena, koh2024visualwebarena, liu2024visualagentbench}.
\emph{Memory and Knowledge Management:} Implementing sophisticated memory systems that allow agents to retain and efficiently retrieve past experiences, facilitating better decision-making and knowledge transfer \cite{zhong2024memorybank, zhang2024survey, lu2023memochat}.
\emph{Reinforcement Learning Integration:} Combining lifelong learning with reinforcement learning techniques to enable agents to learn optimal policies in interactive environments continuously \cite{liu2024autoglm, zhang2024large, shinn2024reflexion}.
\emph{External Knowledge Integration:} Enabling agents to utilize external tools and databases to enhance their capabilities, as well as integrating retrieval mechanisms to access relevant information on-the-fly \cite{schick2024toolformer, lewis2020retrieval}.
\emph{Broader Real-world Applications:} Developing chatbots and interactive agents that can sustain long-term dialogues, adapt to user preferences, and perform tasks within web or gaming environments \cite{hu2023chatdb, wang2023voyager, zhu2023ghost}.

These advancements empower LLM-based agents to operate in more interactive and unpredictable settings, reflecting real-world complexities. For example, an online shopping assistant must not only understand and generate natural language but also navigate product databases, handle user preferences, and adapt to new product categories over time. Similarly, a household management agent must integrate visual inputs from cameras, interpret voice commands, and learn to manage various home devices, all while retaining knowledge from previous interactions.

\subsubsection{Summary}

In summary, the evolution of lifelong learning in AI systems is a story of increasing sophistication—moving from foundational concepts aimed at mitigating forgetting to advanced, principled methods informed by cognitive science and neuroscience. The advent of LLMs has further propelled the field, initially enhancing conventional NLP tasks and now expanding into the realm of LLM-based agents that tackle complex, real-world applications. As LLMs and other large-scale models become integral to real-world applications, lifelong learning stands at the forefront, driving innovation toward systems that improve continuously and gracefully over their operational lifetimes.

\begin{figure*}[!t]
    \centering
    \includegraphics[width=0.99\linewidth]{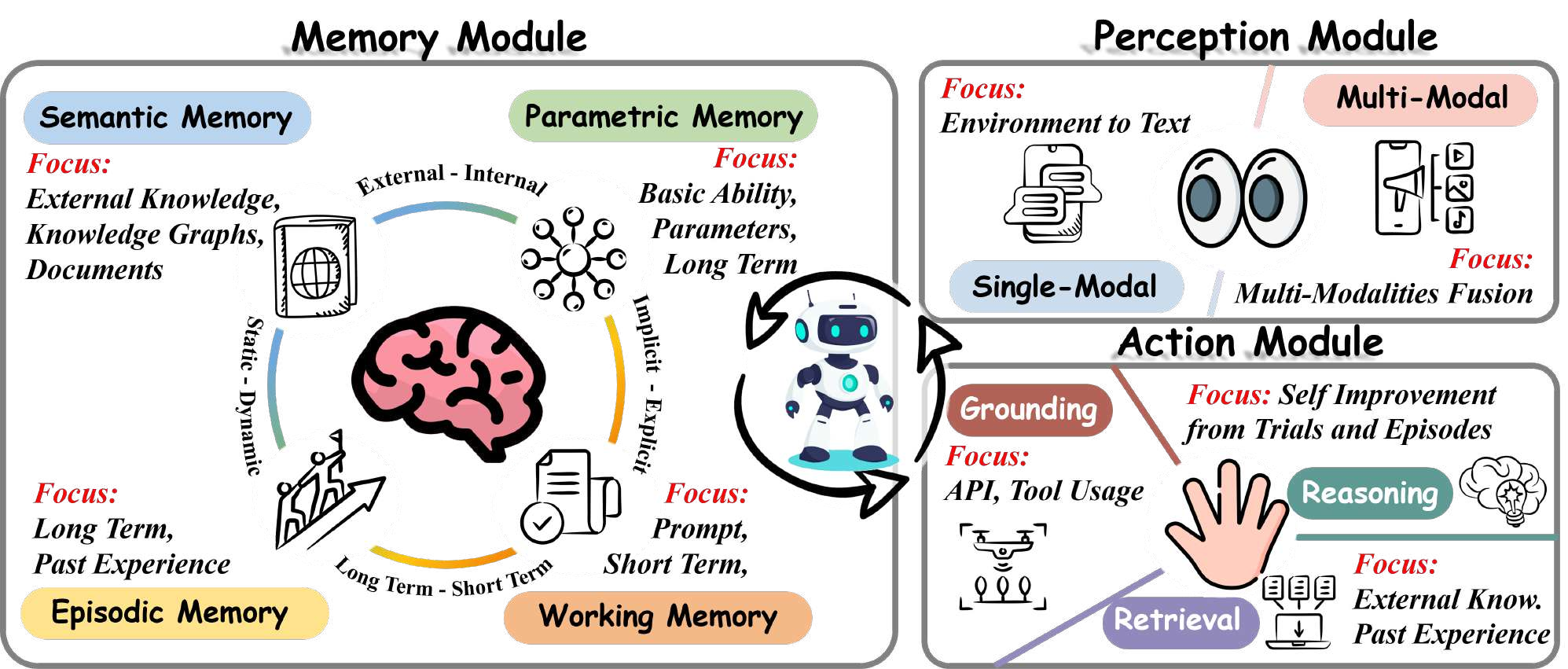}
    \caption{Overall architecture of a lifelong learning LLM-based agent, comprising three key modules—Perception, Memory, and Action. The Perception module continuously gathers and integrates information from the environment. The Memory module stores and manages knowledge from past experiences. The Action module guides interactions and decision-making based on the agent’s current goals and stored knowledge. Together, these modules form a dynamic feedback loop, allowing the agent to adapt, learn, and optimize its behavior across diverse tasks and environments. This design replaces the “Brain” module from previous frameworks, offering a clearer decomposition of functionalities necessary for lifelong learning.}
    \label{fig:illustration_overall_architecture_each_component}
\end{figure*}

\begin{figure*}[!t]
\centering
\tikzset{
        my node/.style={
            draw,
            align=center,
            thin,
            text width=2.5cm, 
            rounded corners=3,
        },
        my leaf/.style={
            draw,
            align=left,
            thin,
            text width=4.5cm, 
            % text height=1cm, 
            % minimum height=0.5cm,
            rounded corners=3,
        }
}
\forestset{
  every leaf node/.style={
    if n children=0{#1}{}
  },
  every tree node/.style={
    if n children=0{minimum width=1em}{#1}
  },
}
\begin{forest}
    for tree={%
        % my node,
        every leaf node={my leaf, font=},
        every tree node={my node, font=\small, l sep-=4.5pt, l-=1.pt},
        anchor=west,
        inner sep=2pt,
        % l = 10pt,
        l sep=10pt, % control leaf to parent nodes gaps (horizontal)
        s sep=5pt, % control node gaps (vertical)
        fit=tight,
        grow'=east,
        edge={ultra thin},
        parent anchor=east,
        child anchor=west,
        if n children=0{tier=last}{},
        edge path={
            \noexpand\path [draw, \forestoption{edge}] (!u.parent anchor) -- +(5pt,0) |- (.child anchor)\forestoption{edge label};
        },
        if={isodd(n_children())}{
            for children={
                if={equal(n,(n_children("!u")+1)/2)}{calign with current}{}
            }
        }{}
    }
    [\cref{sec:building_lifelong_learning_llm_agents} Building Lifelong Learning LLM Agents, draw=gray, color=gray!100, fill=gray!15, very thick, text=black, text width=3.5cm 
        [\cref{sec:building_lifelong_learning_llm_agents:overall_architecture:perception} Perception Design, color=cyan!100, fill=cyan!15, very thick, text=black, text width=3.5cm
            [\cref{sec:single_modal_perception} Single-Modal Perception, color=cyan!100, fill=cyan!15, very thick, text=black
                [\cref{sec:single_modal_perception:web_pages_and_charts_environment} Web Pages and Charts Environment, color=cyan!100, fill=cyan!15, very thick, text=black, tier=Task, text width=6.8cm
                ]
                [\cref{sec:single_modal_perception:game_environment} Game Environment, color=cyan!100, fill=cyan!15, very thick, text=black, tier=Task, text width=6.8cm
                ]
            ]
            [\cref{sec:multi_modal_perception} Multi-Modal Perception, color=cyan!100, fill=cyan!15, very thick, text=black
                [\cref{sec:multi_modal_perception:new_knowledge_perception} New Knowledge Perception, color=cyan!100, fill=cyan!15, very thick, text=black, tier=Task, text width=6.8cm
                ]
                [\cref{sec:multi_modal_perception:old_knowledge_perception} Old Knowledge Perception, color=cyan!100, fill=cyan!15, very thick, text=black, tier=Task, text width=6.8cm
                ]
            ]
        ]
        [\cref{sec:building_lifelong_learning_llm_agents:overall_architecture:memory} Memory Design, color=lightcoral!100, fill=lightcoral!15, very thick, text=black, text width=3.5cm
            [\cref{sec:working_memory} Working Memory, color=lightcoral!100, fill=lightcoral!15, very thick, text=black
                [\cref{sec:working_memory:prompt_compression} Prompt Compression, color=lightcoral!100, fill=lightcoral!15, very thick, text=black, tier=Task, text width=6.8cm
                ]
                [\cref{sec:working_memory:long_context_comprehension} Long Context Comprehension, color=lightcoral!100, fill=lightcoral!15, very thick, text=black, tier=Task, text width=6.8cm
                ]
                [\cref{sec:working_memory:role_playing} Role Playing, color=lightcoral!100, fill=lightcoral!15, very thick, text=black, tier=Task, text width=6.8cm
                ]
                [\cref{sec:working_memory:self_correction} Self Correction, color=lightcoral!100, fill=lightcoral!15, very thick, text=black, tier=Task, text width=6.8cm
                ]
                [\cref{sec:working_memory:prompt_optimization} Prompt Optimization, color=lightcoral!100, fill=lightcoral!15, very thick, text=black, tier=Task, text width=6.8cm
                ]
            ]
            [\cref{sec:episodic_memory} Episodic Memory, color=lightcoral!100, fill=lightcoral!15, very thick, text=black
                [\cref{sec:episodic_memory:data_replay_and_feature_replay} Data Repaly and Feature Replay, color=lightcoral!100, fill=lightcoral!15, very thick, text=black, tier=Task, text width=6.8cm
                ]
                [\cref{sec:episodic_memory:continual_reinforcement_learning} Continual Reinforcement Learning, color=lightcoral!100, fill=lightcoral!15, very thick, text=black, tier=Task, text width=6.8cm
                ]
                [\cref{sec:episodic_memory:self_experience} Self Experience, color=lightcoral!100, fill=lightcoral!15, very thick, text=black, tier=Task, text width=6.8cm
                ]
            ]
            [\cref{sec:semantic_memory} Semantic Memory, color=lightcoral!100, fill=lightcoral!15, very thick, text=black
                [\cref{sec:semantic_memory:continual_knowledge_graph_learning} Continual Knowledge Graph Learning, color=lightcoral!100, fill=lightcoral!15, very thick, text=black, tier=Task, text width=6.8cm
                ]
                [\cref{sec:semantic_memory:continual_document_learning} Continual Document Learning, color=lightcoral!100, fill=lightcoral!15, very thick, text=black, tier=Task, text width=6.8cm
                ]
            ]
            [\cref{sec:parametric_memory} Parametric Memory, color=lightcoral!100, fill=lightcoral!15, very thick, text=black
                [\cref{sec:parametric_memory:continual_instruction_tuning} Continual Instruction Tuning, color=lightcoral!100, fill=lightcoral!15, very thick, text=black, tier=Task, text width=6.8cm
                ]
                [\cref{sec:parametric_memory:continual_knowledge_editing} Continual Knowledge Editing, color=lightcoral!100, fill=lightcoral!15, very thick, text=black, tier=Task, text width=6.8cm
                ]
                [\cref{sec:parametric_memory:continual_alignment} Continual Alignment, color=lightcoral!100, fill=lightcoral!15, very thick, text=black, tier=Task, text width=6.8cm
                ]
            ]
        ]
        [\cref{sec:building_lifelong_learning_llm_agents:overall_architecture:action} Action Design, color=darkpastelgreen!100, fill=darkpastelgreen!15, very thick, text=black, text width=3.5cm 
            [\cref{sec:grounding_actions} Grounding Actions, color=darkpastelgreen!100, fill=darkpastelgreen!15, very thick, text=black
                [\cref{sec:grounding_actions:challenges_of_grounding_actions} Challenges of Grounding Actions, color=darkpastelgreen!100, fill=darkpastelgreen!15, very thick, text=black, tier=Task, text width=6.8cm
                ]
                [\cref{sec:grounding_actions:solutions_for_different_environments} Solution for Different Environments, color=darkpastelgreen!100, fill=darkpastelgreen!15, very thick, text=black, tier=Task, text width=6.8cm
                ]
            ]
            [\cref{sec:retrieval_actions} Retrieval Actions, color=darkpastelgreen!100, fill=darkpastelgreen!15, very thick, text=black
                [\cref{sec:retrieval_actions:retrieval_from_semantic_memory} Retrieval from Semantic Memory, color=darkpastelgreen!100, fill=darkpastelgreen!15, very thick, text=black, tier=Task, text width=6.8cm
                ]
                [\cref{sec:retrieval_actions:retrieval_from_episodic_memory} Retrieval from Episodic Memory, color=darkpastelgreen!100, fill=darkpastelgreen!15, very thick, text=black, tier=Task, text width=6.8cm
                ]
            ]
            [\cref{sec:reasoning_actions} Reasoning Actions, color=darkpastelgreen!100, fill=darkpastelgreen!15, very thick, text=black
                [\cref{sec:reasoning_actions:intra-episodic_reasoning_actions} Intra Episodic Reasoning Actions, color=darkpastelgreen!100, fill=darkpastelgreen!15, very thick, text=black, tier=Task, text width=6.8cm
                ]
                [\cref{sec:reasoning_actions:inter-episodic_reasoning_actions} Inter Episodic Reasoning Actions, color=darkpastelgreen!100, fill=darkpastelgreen!15, very thick, text=black, tier=Task, text width=6.8cm
                ]
            ]
        ]
    ]
\end{forest}
\caption{
Hierarchical organization of the survey sections, highlighting each component of the lifelong learning LLM-based agent architecture and its key functionalities.
}
\label{fig:survey_structure}
\end{figure*}
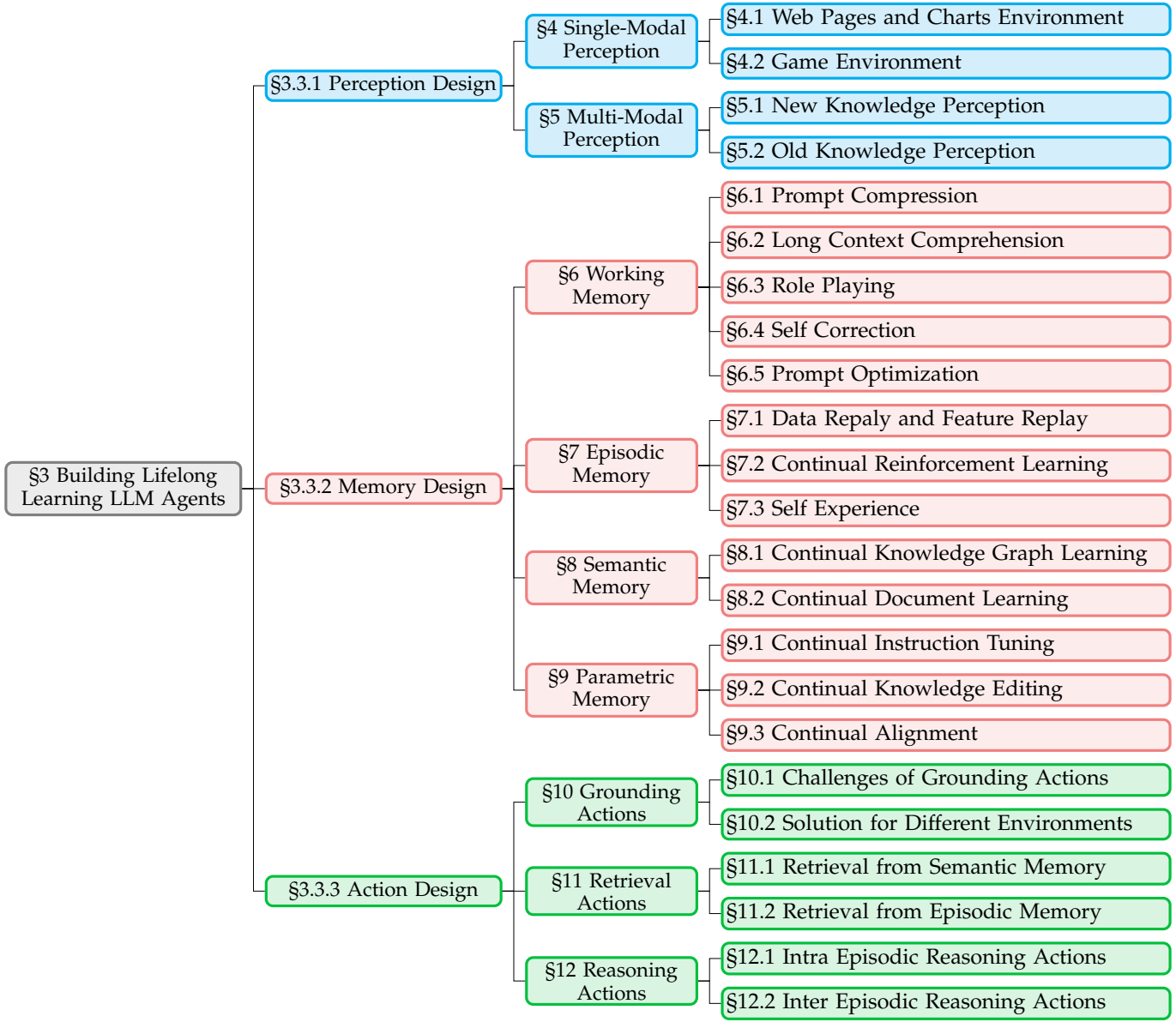

\subsection{Overall Architecture}

The architecture of a lifelong learning LLM-based agent is designed to continually adapt, integrate, and optimize its behavior across a range of tasks and environments. In this subsection, we identify three essential modules—Perception, Memory, and Action—that enable lifelong learning. This division follows the framework introduced in prior work \cite{xi2023rise}, with one notable difference: rather than retaining the “Brain” module, we adopt a “Memory” module as proposed in \cite{xi2023rise}, offering clearer functionality and improved modularity.

Each module interacts with the others to ensure that the agent can process new information, retain valuable knowledge, and select contextually appropriate actions. The rationale for these three modules stems from the agent’s need to (i) perceive and interpret evolving data, (ii) store and manage knowledge from past experiences, and (iii) perform tasks that adapt to changing circumstances.

These modules form a dynamic feedback loop: the Perception module delivers new information to the Memory module, where it is stored and processed. The Memory module then guides the Action module, influencing the environment and informing future perception. Through this continuous cycle, agents progressively refine their knowledge and improve their adaptability, ultimately enhancing their performance in complex, dynamic environments.

In what follows, we describe each module in detail, examining how its design contributes to the agent’s lifelong learning capabilities. Figure \ref{fig:illustration_overall_architecture_each_component} provides an illustration of this overall architecture, while Figure \ref{fig:survey_structure} summarizes the organization of the subsequent sections.

\subsubsection{Perception Module}
\label{sec:building_lifelong_learning_llm_agents:overall_architecture:perception}

The perception module of a lifelong learning LLM-based agent is responsible for acquiring and integrating information from the environment. Similar to humans, who continuously update their understanding through sensory input, agents must perceive and process diverse data sources to remain effective across varying tasks. This module plays a crucial role in adapting the agent’s behavior based on new or evolving contexts.

We divide the perception module into two primary categories: \textbf{single-modal perception} and \textbf{multimodal perception}. 
\emph{Single-modal perception:} This refers to the agent's ability to continuously learn from a single modality, typically textual information. This allows the agent to develop deep domain-specific knowledge, such as understanding webpages or engaging in textual interactions within games or other specialized tasks.
\emph{Multimodal perception:} This expands the agent’s ability to integrate information from multiple modalities—such as visual, auditory, and text data—into a unified understanding of the environment. By combining these different sensory inputs, multimodal perception enables the agent to form a more comprehensive, robust understanding, which is essential for tasks requiring context from various sources (e.g., robotic control, multimedia analysis).

The integration of both single-modal and multimodal perception allows LLM-based agents to continuously learn and adapt to diverse, ever-changing environments.

\subsubsection{Memory Module}
\label{sec:building_lifelong_learning_llm_agents:overall_architecture:memory}

The memory module in a lifelong learning LLM agent allows the agent to store, retain, and recall information—essential for both learning from past experiences and improving decision-making. Memory is the foundation for an agent’s ability to develop coherent long-term behavior, make informed decisions, and interact meaningfully with other agents or humans. The memory module thus supports the agent’s ability to learn from experience, avoid catastrophic forgetting, and enable collaborative behaviors.

We categorize the memory module into four key types: \textbf{Working Memory}, \textbf{Episodic Memory}, \textbf{Semantic Memory}, and \textbf{Parametric Memory}. These four types work together to provide a comprehensive memory system:
\emph{Working Memory:} This represents the agent’s short-term memory, responsible for handling immediate context, such as prompts, user inputs, and relevant workspace information. It enables the agent to act upon the present context in real-time, providing the foundation for short-term reasoning and decision-making.
\emph{Episodic Memory:} This type of memory stores long-term experiences and events, such as user interactions, previous task outcomes, or multi-turn dialogues. Episodic memory helps the agent recall past experiences to improve its future actions, while maintaining consistency in long-term behavior and learning.
\emph{Semantic Memory:} This type of memory functions as an external knowledge store, helping the agent acquire and update world knowledge. Through mechanisms like continual knowledge graph learning and continual document learning, semantic memory facilitates the integration of new knowledge into the agent’s internal framework. By leveraging external databases such as knowledge graphs or dynamic document corpora, semantic memory ensures the agent can stay current with evolving information, improving both its ability to answer queries and enhance long-term learning.
\emph{Parametric Memory:} Unlike the explicit memory of past events, parametric memory resides within the internal parameters of the model. Changes in these parameters—such as through fine-tuning or training updates—reflect long-term knowledge and contribute to the agent’s general knowledge base. This memory type allows the agent to retain knowledge across tasks without storing explicit event details.

The synergy between these memory types supports the agent’s ability to continuously learn, adapt, and avoid catastrophic forgetting, making it capable of lifelong learning.

\subsubsection{Action Module}
\label{sec:building_lifelong_learning_llm_agents:overall_architecture:action}

The action module enables the agent to interact with its environment, make decisions, and execute behaviors that shape the course of its learning. In a lifelong learning framework, actions are crucial for closing the feedback loop: actions influence the environment, and the environment provides feedback that is used to refine future actions. 

We categorize actions into three main types: \textbf{Grounding Actions}, \textbf{Retrieval Actions}, and \textbf{Reasoning Actions}:
\emph{Grounding Actions:} These are the primary means by which the agent interacts with the environment. Grounding actions involve physically or digitally affecting the environment, whether through manipulating objects, generating text, or triggering specific behaviors. The effects of these actions may persist, influencing future behavior and feedback loops.
\emph{Retrieval Actions:} These actions enable the agent to access and retrieve relevant information from its memory, whether from semantic memory (e.g., general knowledge) or episodic memory (e.g., past experiences). Retrieval actions help the agent maintain consistency, acquire new insights, and enhance decision-making.
\emph{Reasoning Actions:} These involve the agent using its working memory, past experiences, and external data to perform complex reasoning, planning, or decision-making tasks. Reasoning actions are essential for tasks that require long-term planning, multi-step decision-making, or the integration of diverse sources of information.
As illustrated in Figure \ref{fig:action_evolution}, these three action types allow the agent to continuously interact with, learn from, and adapt to its environment, supporting a lifelong process of improvement and behavioral evolution. 

\section{Perception Design: Single-modal Perception}
\label{sec:single_modal_perception}

The single-modal perception of LLM-based agents primarily involves the reception of textual information. Throughout the process of lifelong learning, the sources of textual information that agents encounter may originate from various structures and environments. In natural text environments, current LLM-based systems have demonstrated the fundamental capability to communicate with humans through text input and output \cite{touvron2023llama,achiam2023gpt}. Building on this foundation, the agent needs to acquire text information from non-natural text environments to better simulate information perception in the real world. Therefore, in this section, we present methods for single-modality perception in environments such as \emph{web} and \emph{game} environments. 

\subsection{Web Pages and Charts Environment}
\label{sec:single_modal_perception:web_pages_and_charts_environment}

Certain methods have been developed to extract structured text that adheres to standardized formats \cite{yang2024agentoccam,zheng2023synapse,gur2023real,deng2024mind2web,zhou2024webarena}, thereby transforming complex information into a format accessible to LLM-based agents. The mainstream approaches can be broadly classified into two categories: \emph{HTML manipulation} and \emph{screenshot}. For instance, Synapse \cite{zheng2023synapse} and AgentOccam \cite{yang2024agentoccam} simplify HTML elements from web pages and selectively incorporate them into prompts, while WebAgent \cite{gur2023real} summarizes HTML documents and breaks instructions down into multiple sub-instructions, proposing an enhanced planning strategy. Additionally, to effectively harness the information provided by images, several studies \cite{shaw2023pixels,furuta2023multimodal,hong2024cogagent,pan2024autonomous,zheng2024gpt} have converted screenshots into textual formats for alignment with LLMs.

\subsection{Game Environment}
\label{sec:single_modal_perception:game_environment}

LLM-based agents can perceive their surroundings through textual mediums \cite{wang2023jarvis,dongetal2024villageragent,wang2023describe}, recognizing elements such as characters, time, location, events, and emotions \cite{urbanek2019learning}, and can perform corresponding actions based on the feedback from these gaming elements using textual instructions. For example, JARVIS-1 \cite{wang2023jarvis} improves its understanding of the environment through self-reflection and self-explanation, incorporating previous plans into its prompts. VillagerAgent \cite{dongetal2024villageragent} utilizes a dedicated state manager module to filter task-relevant environmental information from the global context. By interfacing with graphical user interfaces (GUIs), LLM-based agents can enhance their ability to extract textual information from visual data \cite{furuta2023multimodal,zheng2024gpt,li2020mapping,zhang2023you}, thus facilitating a better understanding of the graphics and elements within user interfaces.

In conclusion, a human-like LLM-based agent should possess strong text perception and adaptability across a variety of complex environments. As related research continues to grow, exploring how agents perceive text input in broader and more diverse environments holds significant promise for future advancements.

\begin{figure*}[!t]
    \centering
    \subfloat[Agents Perceiving Multimodalies]{
        \includegraphics[width=0.41\linewidth]{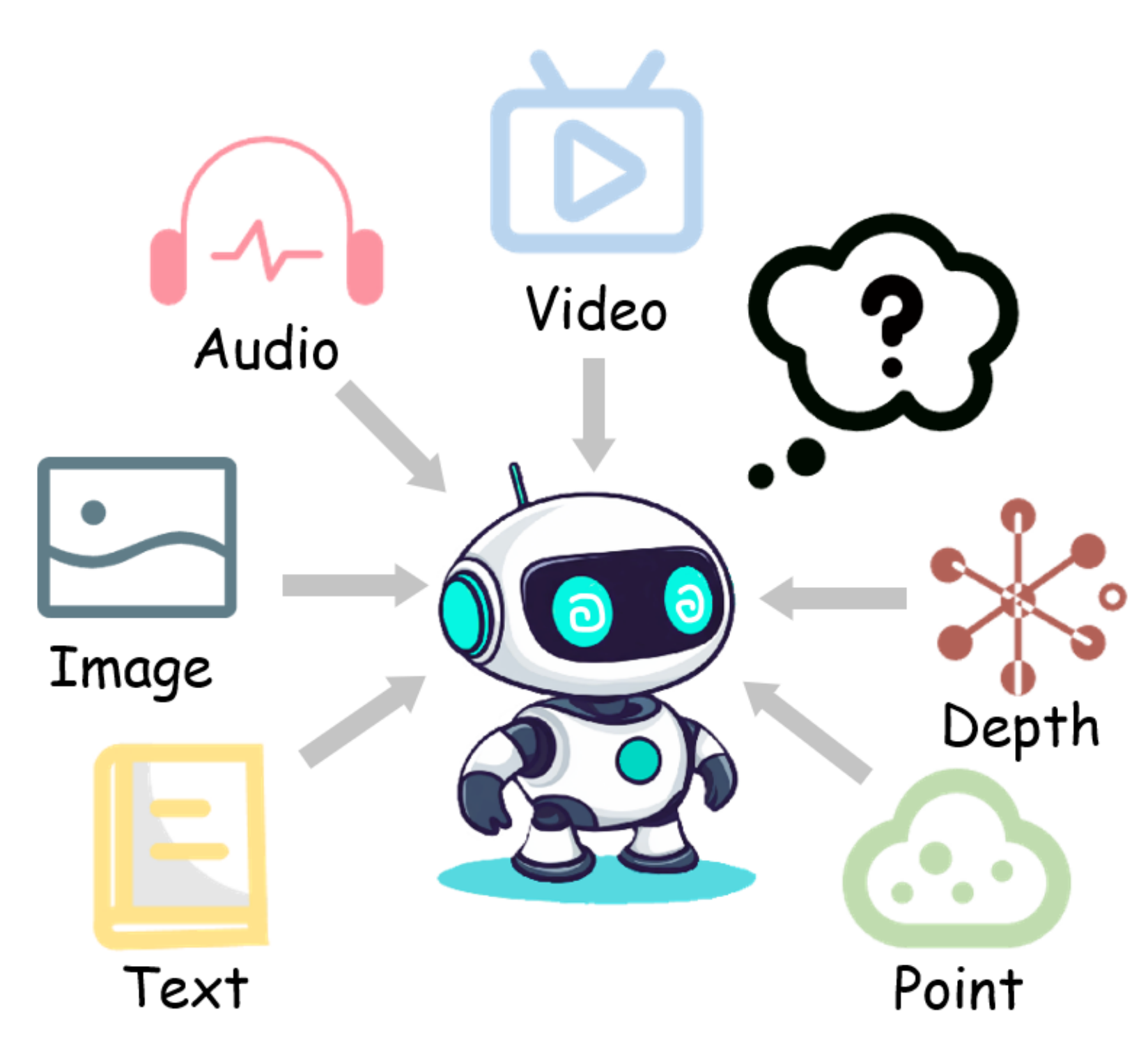}
    }
    \subfloat[Agents Perceiving New Modalities Incrementally]{
        \includegraphics[width=0.57\linewidth]{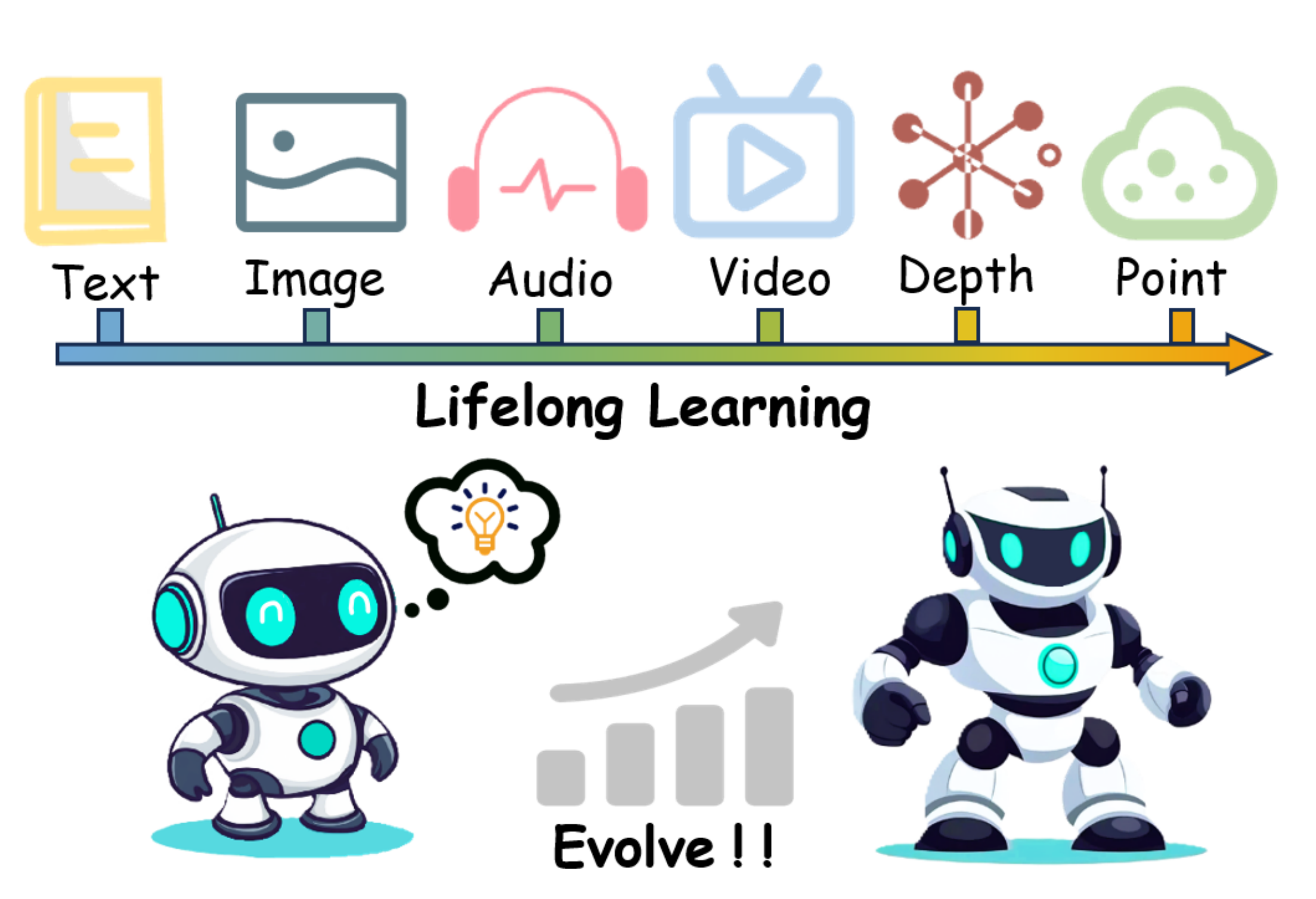}
    }
    \caption{Comparisons of agents to perceive multimodal information: (a) traditional agents require cross-modal data for joint training; (b) lifelong learning agents incrementally learning new modality information without joint-modal retraining.}
    \label{fig:Multimodal_Perception}
\end{figure*}

\renewcommand{\arraystretch}{1.5} 
\begin{table*}[htbp]
  \centering
    \caption{Multimodal perception of Lifelong agents. "Focus": \emph{MA} means processing Modality-Agnostic inputs, \emph{CKT} means crossModal knowledge transfer, \emph{DA} means dynamic adaptation to multimodal, \emph{MCF} means Mitigating catastrophic forgetting; "Modality":{\color[RGB]{3,191,61}{\Checkmark}} indicates the modality involved in the method.}
  \resizebox{0.99\linewidth}{!}{
        \begin{tabular}{c|c|l|c|l|ccccc|c}
    \toprule
    \multicolumn{3}{c|}{
    \multirow{2}[2]{*}{Method}} & 
    \multirow{2}[2]{*}{Focus} & 
    \multicolumn{1}{c|}{
    \multirow{2}[2]{*}{Contribution}} & 
    \multicolumn{5}{c|}{ Modality }       & 
    \multirow{2}[2]{*}{ Code } \\
    \multicolumn{3}{c|}{} &       &       & Language & Vision & Audio & Video & Other &  \\
    \midrule
    \multirow{19}[4]{*}{\rotatebox{90}{New Knowledge Perception}} & 
    \multirow{12}[2]{*}{\rotatebox{90}{Modality-Complete}} & Perceiver\cite{jaegle2021perceiver} & MA   & Utilize asymmetric attention mechanisms to process multimodal data. & \color[RGB]{3,191,61}{\Checkmark} & \color[RGB]{3,191,61}{\Checkmark} & \color[RGB]{3,191,61}{\Checkmark} & \color[RGB]{3,191,61}{\Checkmark} & \color[RGB]{3,191,61}{\Checkmark} & \href{https://github.com/lucidrains/perceiver-pytorch}{Link} \\
          &       & \cellcolor[rgb]{ .949,  .949,  .949}VATT\cite{akbari2021vatt} & \cellcolor[rgb]{ .949,  .949,  .949}MA & \cellcolor[rgb]{ .949,  .949,  .949}Utilize self-supervised multimodal learning methods. & \cellcolor[rgb]{ .949,  .949,  .949}\color[RGB]{3,191,61}{\Checkmark} & \cellcolor[rgb]{ .949,  .949,  .949} & \cellcolor[rgb]{ .949,  .949,  .949}\color[RGB]{3,191,61}{\Checkmark} & \cellcolor[rgb]{ .949,  .949,  .949}\color[RGB]{3,191,61}{\Checkmark} & \cellcolor[rgb]{ .949,  .949,  .949} & \cellcolor[rgb]{ .949,  .949,  .949}\href{https://github.com/google-research/google-research/tree/master/vatt}{Link} \\
          &       & Omnivore\cite{girdhar2022omnivore} & MA   & Introduce a multimodal feature aggregation mechanism. & \color[RGB]{3,191,61}{\Checkmark} & \color[RGB]{3,191,61}{\Checkmark} & \color[RGB]{3,191,61}{\Checkmark} &       &       & \href{https://github.com/facebookresearch/omnivore}{Link} \\
          &       & \cellcolor[rgb]{ .949,  .949,  .949}ModaVerse\cite{wang2024modaverse} & \cellcolor[rgb]{ .949,  .949,  .949}MA & \cellcolor[rgb]{ .949,  .949,  .949}Build a modal conversion framework for rapid multimodal integration. & \cellcolor[rgb]{ .949,  .949,  .949}\color[RGB]{3,191,61}{\Checkmark} & \cellcolor[rgb]{ .949,  .949,  .949}\color[RGB]{3,191,61}{\Checkmark} & \cellcolor[rgb]{ .949,  .949,  .949}\color[RGB]{3,191,61}{\Checkmark} & \cellcolor[rgb]{ .949,  .949,  .949}\color[RGB]{3,191,61}{\Checkmark} & \cellcolor[rgb]{ .949,  .949,  .949} & \cellcolor[rgb]{ .949,  .949,  .949}\href{https://github.com/xinke-wang/ModaVerse}{Link} \\
          &       & Vokenization\cite{tan2020vokenization} & CKT   & Integrate visual information into the training of language models. & \color[RGB]{3,191,61}{\Checkmark} & \color[RGB]{3,191,61}{\Checkmark} &       &       &       & \href{https://github.com/airsplay/vokenization}{Link} \\
          &       & \cellcolor[rgb]{ .949,  .949,  .949}VidLanKD\cite{tang2021vidlankd} & \cellcolor[rgb]{ .949,  .949,  .949}CKT & \cellcolor[rgb]{ .949,  .949,  .949}Utilize cross-modal KD to achieve cross-modal knowledge transfer. & \cellcolor[rgb]{ .949,  .949,  .949}\color[RGB]{3,191,61}{\Checkmark} & \cellcolor[rgb]{ .949,  .949,  .949} & \cellcolor[rgb]{ .949,  .949,  .949}\color[RGB]{3,191,61}{\Checkmark} & \cellcolor[rgb]{ .949,  .949,  .949} & \cellcolor[rgb]{ .949,  .949,  .949} & \cellcolor[rgb]{ .949,  .949,  .949}\href{https://github.com/zinengtang/VidLanKD}{Link} \\
          &       & Gupta et al.\cite{gupta2016cross} & CKT   & Transfer the features learned from one modality as supervision signals to another. & \color[RGB]{3,191,61}{\Checkmark} & \color[RGB]{3,191,61}{\Checkmark} & \color[RGB]{3,191,61}{\Checkmark} &       &       & \href{https://github.com/s-gupta/fast-rcnn/tree/distillation}{Link} \\
          &       & \cellcolor[rgb]{ .949,  .949,  .949}ZSCL\cite{zheng2023preventing} & \cellcolor[rgb]{ .949,  .949,  .949}CKT, MCF & \cellcolor[rgb]{ .949,  .949,  .949}Match image-text similarity distributions through KD. & \cellcolor[rgb]{ .949,  .949,  .949}\color[RGB]{3,191,61}{\Checkmark} & \cellcolor[rgb]{ .949,  .949,  .949}\color[RGB]{3,191,61}{\Checkmark} & \cellcolor[rgb]{ .949,  .949,  .949} & \cellcolor[rgb]{ .949,  .949,  .949} & \cellcolor[rgb]{ .949,  .949,  .949} & \cellcolor[rgb]{ .949,  .949,  .949}\href{https://github.com/Thunderbeee/ZSCL}{Link} \\
          &       & SoundSpaces\cite{chen2020soundspaces} & CKT   & Integrate audio and visual information to facilitate navigation in 3D environments. &       & \color[RGB]{3,191,61}{\Checkmark} & \color[RGB]{3,191,61}{\Checkmark} &       & \color[RGB]{3,191,61}{\Checkmark} & \href{http://vision.cs.utexas.edu/projects/audio_visual_navigation}{Link} \\
          &       & \cellcolor[rgb]{ .949,  .949,  .949}LMC\cite{chen2023continual} & \cellcolor[rgb]{ .949,  .949,  .949}CKT, MCF & \cellcolor[rgb]{ .949,  .949,  .949}Build a multimodal knowledge graph to enhance information between modalities. & \cellcolor[rgb]{ .949,  .949,  .949}\color[RGB]{3,191,61}{\Checkmark} & \cellcolor[rgb]{ .949,  .949,  .949}\color[RGB]{3,191,61}{\Checkmark} & \cellcolor[rgb]{ .949,  .949,  .949}\color[RGB]{3,191,61}{\Checkmark} & \cellcolor[rgb]{ .949,  .949,  .949}\color[RGB]{3,191,61}{\Checkmark} & \cellcolor[rgb]{ .949,  .949,  .949} & \cellcolor[rgb]{ .949,  .949,  .949}\href{https://github.com/zjunlp/ContinueMKGC}{Link} \\
          &       & EPIC-Fusion\cite{kazakos2019epic} & CKT   & Enhance egocentric action recognition using audio-visual temporal binding. &       & \color[RGB]{3,191,61}{\Checkmark} & \color[RGB]{3,191,61}{\Checkmark} &       & \color[RGB]{3,191,61}{\Checkmark} & \href{https://github.com/ekazakos/temporal-binding-network}{Link} \\
          &       & \cellcolor[rgb]{ .949,  .949,  .949}Zhang et al.\cite{zhang2021repetitive} & \cellcolor[rgb]{ .949,  .949,  .949}CKT & \cellcolor[rgb]{ .949,  .949,  .949}Combine visual and auditory cues to improve the counting of repetitive activities. & \cellcolor[rgb]{ .949,  .949,  .949} & \cellcolor[rgb]{ .949,  .949,  .949}\color[RGB]{3,191,61}{\Checkmark} & \cellcolor[rgb]{ .949,  .949,  .949}\color[RGB]{3,191,61}{\Checkmark} & \cellcolor[rgb]{ .949,  .949,  .949} & \cellcolor[rgb]{ .949,  .949,  .949} & \cellcolor[rgb]{ .949,  .949,  .949}\href{https://github.com/xiaobai1217/RepetitionCounting}{Link} \\
\cmidrule{2-11}          & \multirow{7}[2]{*}{\rotatebox{90}{Modality-Incomplete}} & SMIL\cite{ma2021smil} & DA    & Adopt an adaptive modality weighting mechanism to enhance robustness. & \color[RGB]{3,191,61}{\Checkmark} & \color[RGB]{3,191,61}{\Checkmark} & \color[RGB]{3,191,61}{\Checkmark} &       &       & \href{https://github.com/mengmenm/SMIL}{Link} \\
          &       & 
          \cellcolor[rgb]{ .949,  .949,  .949}PathWeave\cite{yu2024llms} & \cellcolor[rgb]{ .949,  .949,  .949}DA, CKT & \cellcolor[rgb]{ .949,  .949,  .949}Introduce a novel 'Adapter in Adapter' framework to achieve modality alignment. & \cellcolor[rgb]{ .949,  .949,  .949}\color[RGB]{3,191,61}{\Checkmark} & \cellcolor[rgb]{ .949,  .949,  .949}\color[RGB]{3,191,61}{\Checkmark} & \cellcolor[rgb]{ .949,  .949,  .949}\color[RGB]{3,191,61}{\Checkmark} & \cellcolor[rgb]{ .949,  .949,  .949} & \cellcolor[rgb]{ .949,  .949,  .949} & \cellcolor[rgb]{ .949,  .949,  .949}\href{https://github.com/JiazuoYu/PathWeave}{Link} \\
          &       & DDAS\cite{yu2024boosting} & DA, CKT & Hand modality data for in-distribution and out-of-distribution by routing. & \color[RGB]{3,191,61}{\Checkmark} & \color[RGB]{3,191,61}{\Checkmark} &       &       &       & \href{https://github.com/JiazuoYu/MoE-Adapters4CL}{Link} \\
          &       & \cellcolor[rgb]{ .949,  .949,  .949}MMIN\cite{zhao2021missing} & \cellcolor[rgb]{ .949,  .949,  .949}DA & \cellcolor[rgb]{ .949,  .949,  .949}Generate imagined features of missing modalities using conditional GANs. & \cellcolor[rgb]{ .949,  .949,  .949}\color[RGB]{3,191,61}{\Checkmark} & \cellcolor[rgb]{ .949,  .949,  .949}\color[RGB]{3,191,61}{\Checkmark} & \cellcolor[rgb]{ .949,  .949,  .949}\color[RGB]{3,191,61}{\Checkmark} & \cellcolor[rgb]{ .949,  .949,  .949} & \cellcolor[rgb]{ .949,  .949,  .949} & \cellcolor[rgb]{ .949,  .949,  .949}\href{https://github.com/AIM3RUC/MMIN}{Link} \\
          &       & DiCMoR\cite{wang2023distribution} & DA    & Use a generative model to minimize the feature difference for modality recovery. & \color[RGB]{3,191,61}{\Checkmark} & \color[RGB]{3,191,61}{\Checkmark} & \color[RGB]{3,191,61}{\Checkmark} &       &       & \href{https://github.com/mdswyz/DiCMoR}{Link} \\
          &       & \cellcolor[rgb]{ .949,  .949,  .949}Zhang et al.\cite{zhang2024learning} & \cellcolor[rgb]{ .949,  .949,  .949}DA & \cellcolor[rgb]{ .949,  .949,  .949}Generalize unseen modality combinations via projection and pseudo-supervision. & \cellcolor[rgb]{ .949,  .949,  .949}\color[RGB]{3,191,61}{\Checkmark} & \cellcolor[rgb]{ .949,  .949,  .949}\color[RGB]{3,191,61}{\Checkmark} & \cellcolor[rgb]{ .949,  .949,  .949}\color[RGB]{3,191,61}{\Checkmark} & \cellcolor[rgb]{ .949,  .949,  .949} & \cellcolor[rgb]{ .949,  .949,  .949}\color[RGB]{3,191,61}{\Checkmark} & \cellcolor[rgb]{ .949,  .949,  .949}\href{https://xiaobai1217.github.io/Unseen-Modality-Interaction/}{Link} \\
          &       & ShaSpec\cite{wang2023multi} & DA    & Hand missing modalities using shared and specific feature representations. &       & \color[RGB]{3,191,61}{\Checkmark} & \color[RGB]{3,191,61}{\Checkmark} &       & \color[RGB]{3,191,61}{\Checkmark} & \href{https://github.com/billhhh/ShaSpec}{Link} \\
          &       & \cellcolor[rgb]{ .949,  .949,  .949}MissModal\cite{lin2023missmodal} & \cellcolor[rgb]{ .949,  .949,  .949}DA & \cellcolor[rgb]{ .949,  .949,  .949}Introduce constraints to align representations of missing modality data. & \cellcolor[rgb]{ .949,  .949,  .949}\color[RGB]{3,191,61}{\Checkmark} & \cellcolor[rgb]{ .949,  .949,  .949}\color[RGB]{3,191,61}{\Checkmark} & \cellcolor[rgb]{ .949,  .949,  .949}\color[RGB]{3,191,61}{\Checkmark} & \cellcolor[rgb]{ .949,  .949,  .949} & \cellcolor[rgb]{ .949,  .949,  .949} & \cellcolor[rgb]{ .949,  .949,  .949}\href{https://github.com/RH-Lin/MissModal}{Link} \\
          &       & Ma et al.\cite{ma2022multimodal} & DA    & Improve robustness by automatically finding the optimal data fusion strategy. & \color[RGB]{3,191,61}{\Checkmark} & \color[RGB]{3,191,61}{\Checkmark} &       &       &       & - \\
    \midrule
    \multirow{12}[3]{*}{\rotatebox{90}{Old Knowledge Perception}} & \multirow{4}[2]{*}{\rotatebox{90}{Regularization}} & \cellcolor[rgb]{ .949,  .949,  .949}TIR\cite{he2023continual} & \cellcolor[rgb]{ .949,  .949,  .949}MCF & \cellcolor[rgb]{ .949,  .949,  .949}Adapt weight regularization based on the similarity between old and new tasks. & \cellcolor[rgb]{ .949,  .949,  .949}\color[RGB]{3,191,61}{\Checkmark} & \cellcolor[rgb]{ .949,  .949,  .949}\color[RGB]{3,191,61}{\Checkmark} & \cellcolor[rgb]{ .949,  .949,  .949} & \cellcolor[rgb]{ .949,  .949,  .949} & \cellcolor[rgb]{ .949,  .949,  .949} & \cellcolor[rgb]{ .949,  .949,  .949}\href{https://github.com/Thunderbeee/ZSCL}{Link} \\
          &       & Model Tailor\cite{zhu2024model} & MCF   & Identifying key parameters for Combine regularization. & \color[RGB]{3,191,61}{\Checkmark} & \color[RGB]{3,191,61}{\Checkmark} &       &       &       & - \\
          &       & \cellcolor[rgb]{ .949,  .949,  .949}Mod-X\cite{ni2023continual} & \cellcolor[rgb]{ .949,  .949,  .949}CKT, MCF & \cellcolor[rgb]{ .949,  .949,  .949}Distill on contrastive matrix to preserve spatial distribution between modalities. & \cellcolor[rgb]{ .949,  .949,  .949}\color[RGB]{3,191,61}{\Checkmark} & \cellcolor[rgb]{ .949,  .949,  .949}\color[RGB]{3,191,61}{\Checkmark} & \cellcolor[rgb]{ .949,  .949,  .949} & \cellcolor[rgb]{ .949,  .949,  .949} & \cellcolor[rgb]{ .949,  .949,  .949} & \cellcolor[rgb]{ .949,  .949,  .949}- \\
          &       & CTP\cite{zhu2023ctp} & CKT, MCF & Utilize KD to maintain the similarity distributions between image and text. & \color[RGB]{3,191,61}{\Checkmark} & \color[RGB]{3,191,61}{\Checkmark} &       &       &       & \href{https://github.com/KevinLight831/CTP}{Link} \\
\cmidrule{2-11}          & \multirow{8}[1]{*}{\rotatebox{90}{Replay}} & \cellcolor[rgb]{ .949,  .949,  .949}Vqacl\cite{zhang2023vqacl} & \cellcolor[rgb]{ .949,  .949,  .949}MCF & \cellcolor[rgb]{ .949,  .949,  .949}Randomly Replay past task samples to maintain old task performance. & \cellcolor[rgb]{ .949,  .949,  .949}\color[RGB]{3,191,61}{\Checkmark} & \cellcolor[rgb]{ .949,  .949,  .949}\color[RGB]{3,191,61}{\Checkmark} & \cellcolor[rgb]{ .949,  .949,  .949} & \cellcolor[rgb]{ .949,  .949,  .949} & \cellcolor[rgb]{ .949,  .949,  .949} & \cellcolor[rgb]{ .949,  .949,  .949}\href{https://github.com/zhangxi1997/VQACL}{Link} \\
          &       & SAMM\cite{sarfraz2024beyond} & CKT, MCF & Store randomly selected training samples in a memory buffer to avoid forgetting. &       & \color[RGB]{3,191,61}{\Checkmark} & \color[RGB]{3,191,61}{\Checkmark} &       &       & \href{https://github.com/NeurAI-Lab/MultiModal-CL}{Link} \\
          &       & \cellcolor[rgb]{ .949,  .949,  .949}TAM-CL\cite{cai2023task} & \cellcolor[rgb]{ .949,  .949,  .949}CKT, MCF & \cellcolor[rgb]{ .949,  .949,  .949}Combine KD and replay to transfer knowledge from old tasks. & \cellcolor[rgb]{ .949,  .949,  .949}\color[RGB]{3,191,61}{\Checkmark} & \cellcolor[rgb]{ .949,  .949,  .949}\color[RGB]{3,191,61}{\Checkmark} & \cellcolor[rgb]{ .949,  .949,  .949} & \cellcolor[rgb]{ .949,  .949,  .949} & \cellcolor[rgb]{ .949,  .949,  .949} & \cellcolor[rgb]{ .949,  .949,  .949}\href{https://github.com/YuliangCai2022/TAM-CL}{Link} \\
          &       & KDR\cite{yang2023knowledge} & MCF   & Combine KD to replay the image-text similarity matrix output by the old task model. & \color[RGB]{3,191,61}{\Checkmark} & \color[RGB]{3,191,61}{\Checkmark} &       &       &       & - \\
          &       & \cellcolor[rgb]{ .949,  .949,  .949}IncCLIP\cite{yan2022generative} & \cellcolor[rgb]{ .949,  .949,  .949}CKT, MCF & \cellcolor[rgb]{ .949,  .949,  .949}Generate negative text replay and multimodal knowledge distillation. & \cellcolor[rgb]{ .949,  .949,  .949}\color[RGB]{3,191,61}{\Checkmark} & \cellcolor[rgb]{ .949,  .949,  .949}\color[RGB]{3,191,61}{\Checkmark} & \cellcolor[rgb]{ .949,  .949,  .949} & \cellcolor[rgb]{ .949,  .949,  .949} & \cellcolor[rgb]{ .949,  .949,  .949} & \cellcolor[rgb]{ .949,  .949,  .949}- \\
          &       & SGP\cite{lei2023symbolic} & MCF   & replay past knowledge through randomly sampled scene graph relationships. & \color[RGB]{3,191,61}{\Checkmark} & \color[RGB]{3,191,61}{\Checkmark} &       &       &       & \href{https://github.com/showlab/CLVQA}{Link} \\
          &       & \cellcolor[rgb]{ .949,  .949,  .949}FGVIRs\cite{he2024continual} & \cellcolor[rgb]{ .949,  .949,  .949}MCF & \cellcolor[rgb]{ .949,  .949,  .949}Store old knowledge to create pseudo-representations for training classifier. & \cellcolor[rgb]{ .949,  .949,  .949} & \cellcolor[rgb]{ .949,  .949,  .949}\color[RGB]{3,191,61}{\Checkmark} & \cellcolor[rgb]{ .949,  .949,  .949} & \cellcolor[rgb]{ .949,  .949,  .949} & \cellcolor[rgb]{ .949,  .949,  .949}\color[RGB]{3,191,61}{\Checkmark} & \cellcolor[rgb]{ .949,  .949,  .949}- \\
          &       & AID[38]\cite{cheng2024vision} & MCF   & Retain old knowledge by Enhance the model's representational capacity. &       & \color[RGB]{3,191,61}{\Checkmark} &       &       & \color[RGB]{3,191,61}{\Checkmark} & - \\
          \bottomrule
    \end{tabular}%
    }
  \label{tab:multimodal_perception_agent}%
\end{table*}%

\section{Perception Design: Multimodal Perception}
\label{sec:multi_modal_perception}

The real world comprises diverse data modalities, making unimodal perception methods insufficient to address its complexity. With the explosive growth of image, text, and video content on online platforms, developing LLM-based agents capable of continuously perceiving multimodal information has become crucial. These agents must effectively integrate information from different modalities while maintaining the accumulation and adaptation of prior knowledge. This enables them to better emulate the lifelong learning process of humans in multimodal environments, thereby enhancing their overall perceptual and cognitive capabilities. In this section, we categorize the lifelong learning methods for the agent's perception of multimodal information into \emph{new knowledge perception} and \emph{old knowledge perception}. A comparison and summarization of the related methods in provided in Table \ref{tab:multimodal_perception_agent}.

\subsection{New Knowledge Perception}
\label{sec:multi_modal_perception:new_knowledge_perception}

In lifelong learning scenarios involving the perception of various modalities, agents must focus on the interaction between different modalities and the perception and processing of new modalities to better handle the rapidly evolving forms of information in the real world. Many studies explore how an agent can maintain stability on tasks involving existing modalities while improving its capability to handle new tasks when encountering those with new modalities. We categorize new knowledge perception into the following two scenarios: \emph{Modality-Complete Learning} and \emph{Modality-Incomplete Learning}. 

\subsubsection{Modality-Complete Learning}

Modality-Complete Learning assumes that all data has the same modality during both the training and inference stages. In this scenario, the agent’s lifelong learning for multimodal perception focuses on how to perceive data from multiple modalities and achieve cross-modal knowledge transfer in new tasks. 

Some studies \cite{jaegle2021perceiver,akbari2021vatt,girdhar2022omnivore,wang2024modaverse} have explored Modality-Agnostic Models, which can perceive any modality as input. Perceiver \cite{jaegle2021perceiver} utilizes an iterative attention mechanism to dynamically encode any modality of data and generate a unified high-dimensional representation. VATT \cite{akbari2021vatt} adopts a shared transformer architecture and learns cross-modal correlations effectively from raw video, audio, and text through multimodal self-supervised learning tasks. Omnivore \cite{girdhar2022omnivore} introduces a multimodal feature aggregation mechanism, using shared convolutional layers and an adaptive feature fusion strategy to achieve a unified representation of various visual modalities such as images, videos, and depth maps. ModaVerse \cite{wang2024modaverse} leverages LLMs as the core to build an efficient modality transformer framework, enabling fast information fusion across different modalities and improving the processing efficiency of multimodal tasks. 

On the other hand, Cross-Modal Knowledge Transfer refers to the effective application of knowledge acquired in one modality to improve performance on tasks involving another modality \cite{tan2020vokenization,tang2021vidlankd,jin2022leveraging,kazakos2019epic,zhang2021repetitive,chen2020soundspaces}. Some methods achieve knowledge transfer between different modalities by feeding connected unimodal features into linear layers \cite{kazakos2019epic,zhang2021repetitive,chen2020soundspaces}. In visual-language tasks, some studies \cite{tan2020vokenization,tang2021vidlankd,jin2022leveraging} focus on transferring visual knowledge into language models. For example, Vokenization \cite{tan2020vokenization} introduces "visual tokens" and performs text-to-image retrieval using contextual images, embedding visual information into the training of language models to enhance language understanding. VidLanKD \cite{tang2021vidlankd} uses contrastive learning to train a teacher model on video datasets and distills the extracted visual features into the language model, improving its performance on text tasks related to video. 

Additionally, by combining knowledge distillation \cite{gou2021knowledge,gupta2016cross} and regularization methods, some studies \cite{tang2021vidlankd,zheng2023preventing,zhu2023ctp,ni2023continual,chen2023continual}  leverage the high-performance capability of the teacher model in one modality to transfer its knowledge to the student model through constraint-based mapping. For example, the CTP \cite{zhu2023ctp} method employs a distillation strategy that incorporates compatible momentum contrast and topology-preserving techniques, achieving a balance between old and new task knowledge in visual-language continual pretraining, effectively alleviating catastrophic forgetting. Mod-X \cite{ni2023continual} uses contrastive matrix alignment of off-diagonal information and spatial distribution distillation to maintain alignment of the representation space across old data domains from different modalities.

\subsubsection{Modality-Incomplete Learning}

Modality-Incomplete Learning involves the challenge of how the agent can dynamically adapt to effectively learn and infer when encountering incomplete or missing modality information during lifelong learning (See Figure \ref{fig:Multimodal_Perception}). By utilizing the MoE \cite{masoudnia2014mixture} module, PathWeave \cite{yu2024llms} introduces a novel Adapter in Adapter (AnA) framework, enabling the seamless integration of single-modal and cross-modal adapters, allowing for incremental learning of new modal knowledge. Similarly, DDAS \cite{yu2024boosting} can automatically route inputs within and outside the training data distribution to the MoE adapter and the original CLIP model, respectively, to achieve efficient dynamic modality adaptation. 

Some studies \cite{ma2021smil,zhao2021missing,wang2023distribution,zhang2024learning} have utilized available modality information to predict the representation of the missing modality. Specifically, SMIL \cite{ma2021smil} proposes an adaptive modality weighting mechanism, enhancing robustness to severely missing modalities through multimodal fusion. MMIN \cite{zhao2021missing} uses conditional generative adversarial networks to generate imagined features for the missing modality, improving performance in sentiment recognition tasks. DiCMoR \cite{wang2023distribution} introduces a generative model to minimize the difference between available modalities and the true feature distribution, thereby enabling accurate modality recovery. The latest research \cite{zhang2024learning} projects the multi-dimensional features of different modalities into a shared space and uses pseudo-supervision to indicate the reliability of modality prediction, enabling generalization to unseen modality combinations. 

Other studies improve the representation of input data by learning shared and modality-specific features \cite{wang2023multi,lin2023missmodal}, which helps demonstrate better robustness when handling missing modalities. Similarly, Ma et al. \cite{ma2022multimodal}, based on the sensitivity of transformer models to missing modalities, constructed a robust transformer optimized through multi-task learning. They also developed an algorithm to automatically search for the optimal fusion strategy across different datasets to optimize test-time robustness. 

In summary, new knowledge perception emphasizes the incremental perception of multimodal information by LLM-based agents in real-world complex environments. The agent perceives new modality inputs and integrating the new knowledge with existing modality experiences to enable lifelong learning perception. 

\subsection{Old Knowledge Perception}
\label{sec:multi_modal_perception:old_knowledge_perception}

In the process of lifelong learning, the agent not only needs to leverage existing modality experiences to complete tasks involving new modalities, but also must maintain the stability of old knowledge after receiving new information. In this section, we present a classification of representative multimodal perception lifelong learning methods aimed at addressing the problem of catastrophic forgetting.

\subsubsection{Regularization-based Approach}

The regularization-based Approach aims to mitigate the phenomenon of catastrophic forgetting by introducing regularization terms that limit the changes in model parameters during the learning of new tasks. Based on the method of constraint application, the Regularization-based Approach can be subdivided into two directions: \emph{weight Regularization} and \emph{function Regularization}.

Weight Regularization directly applies penalty terms to the model's weights, restricting their changes when learning new tasks. For example, TIR \cite{he2023continual} and Model Tailor \cite{zhu2024model} utilize existing importance measurement methods like EWC \cite{kirkpatrick2017overcoming} to calculate the importance of parameters for old tasks, thereby applying penalties.

Function Regularization focuses on constraining the intermediate or final outputs of the model, ensuring that the model retains the output features of old tasks while learning new tasks. Function Regularization often uses knowledge distillation strategies, where the previously learned model serves as the teacher, and the model learning the new task serves as the student \cite{zheng2023preventing,zhu2023ctp,ni2023continual ,chen2023continual} For example, the CTP \cite{ni2023continual} method achieves a balance between old and new task knowledge in visual-language continual pretraining through momentum contrast and topology-preserving distillation strategies, effectively reducing catastrophic forgetting. Mod-X \cite{chen2023continual} aligns non-diagonal information in the contrastive matrix and distills spatial distribution to maintain the alignment of the representation space across modalities for old data domains.

\subsubsection{Replay-based Approach}

The Replay-based Approach is a method that alleviates catastrophic forgetting by preserving and reusing previously learned experiences. In multimodal continual perception learning, depending on the specific content of the replay, the method can be divided into \emph{Experience Replay} and \emph{Generative Replay}.

Due to storage space limitations, Experience Replay focuses on how to store more representative old training samples in a limited memory space. Some studies \cite{zhang2023vqacl,sarfraz2024beyond} randomly select multimodal samples for replay. TAM-CL \cite{cai2023task} and KDR \cite{yang2023knowledge} combine experience replay with knowledge distillation strategies to constrain distribution shifts, thus reducing catastrophic forgetting.
 
Generative Replay requires training an additional generative model to replay generated data \cite{yan2022generative,lei2023symbolic,he2024continual,cheng2024vision}, which effectively reduces storage requirements. IncCLIP \cite{yan2022generative } enhances continuous visual-language pretraining by generating negative sample texts, thus improving the model's robustness when facing new tasks. SGP \cite{lei2023symbolic}  uses scene graphs as prompts and integrates them with language models to enhance continual learning in visual question answering tasks.

Existing studies also include projection-based and architecture-based methods. Projection-based methods map data from different modalities (e.g., images, text, and audio) to a unified feature space \cite{song2020real,zhang2024learning,zhao2021missing,han2024onellm,moon2024anymal,zheng2024beyond}, enabling the model to process information from multiple modalities. The architecture-based approach is a strategy that supports lifelong learning by adjusting the model's structure. This approach divides the model into task-shared and task-specific components, ensuring relative isolation between tasks to reduce the impact of learning new tasks on the retention of old knowledge \cite{del2020ratt}.

In conclusion, old knowledge perception focuses on the agent's ability to retain and perceive old knowledge as it continually receives multimodal information during lifelong learning. By incorporating existing lifelong learning methods, the agent can significantly mitigate the issue of catastrophic forgetting.

\section{Memory Design: Working Memory}
\label{sec:working_memory}

\begin{figure}[!t]
    \centering
    \includegraphics[width=0.99\linewidth]{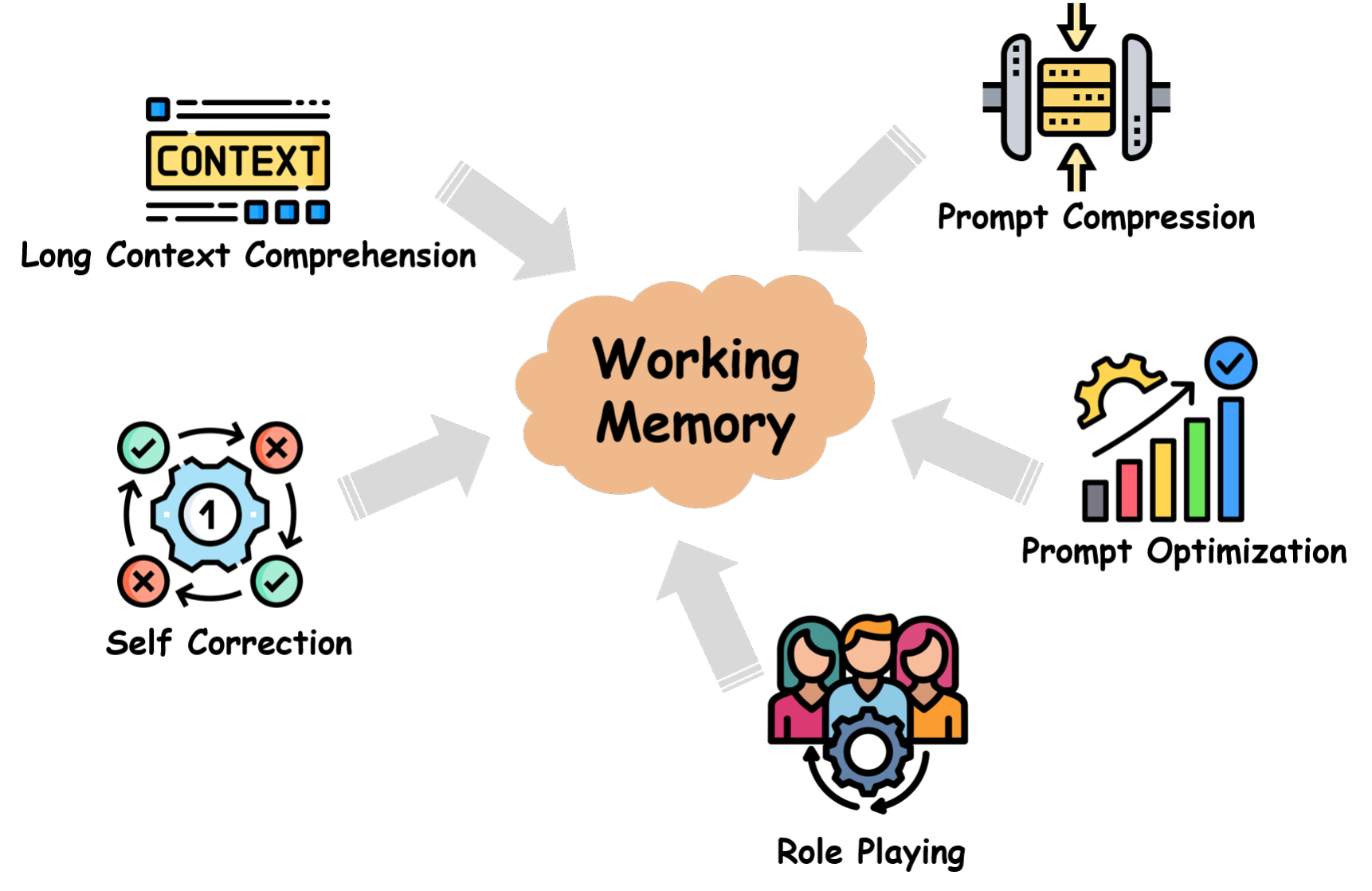}
    \caption{Potential techniques for working memory.}
    \label{fig:memory_working}
\end{figure}

Working Memory is primarily manifested as the short-term memory of the agent, encompassing elements such as prompts, workspace memory, and context provided by users. Agents utilize these prompts or contextual inputs as their working memory to generate responses or to proceed with planning and actions. Working memory serves as the operational memory of the agent, facilitating real-time interactions and decision-making processes. In short, working memory is the active workspace of the agent, where immediate information is processed and manipulated to produce responses or to guide subsequent actions. We discuss working memory from five main perspectives: \textbf{prompt compression}, \textbf{long context comprehension}, \textbf{role playing}, \textbf{self correction}, and \textbf{prompt optimization}, as illustrated in Figure \ref{fig:memory_working}.

\subsection{Prompt Compression}
\label{sec:working_memory:prompt_compression}

In terms of working memory, agents can include more contextual content within the limited length of prompts by compressing prompt words that users input. This process not only increases the efficiency of information processing, but also allows agents to effectively avoid catastrophic forgetting of historical information when integrating old prompts into new ones. In this way, agents can retain memories of previous experiences while continuously learning new information, achieving the goal of lifelong learning. The techniques for prompt compression are classified mainly into two categories: \textbf{soft compression} and \textbf{hard compression}. 

\emph{Soft compression} learns alternative forms of embeddings by optimizing a small number of soft prompt tokens to compress the original prompts, preserving the abstract sentiment or key information. AutoCompressors \cite{chevalier2023adapting} can process the prompts input to the model by recursively generating summary vectors that are passed as soft prompts to all subsequent prompt segments. It is ingenious to fine-tune the existing model structure by increasing the vocabulary and utilizing \emph{summary tokens} and \emph{summary vectors} to refine a large amount of prompt information. Mu et al. \cite{mu2024learning} propose \emph{gisting} to compress prompts into a smaller set of \emph{gist tokens} by training the language model. Considering that gist tokens are much shorter than the length of the original input prompts, these tokens can be cached and reused, thus improving computational efficiency.

In contrast, \emph{hard compression} focuses on directly compressing prompts by filtering out redundant or non-essential tokens based on various scores computed by a pre-trained model. Selective Context \cite{li2023compressing} employs a mini-language model to compute the self-information values of individual lexical units (e.g., sentences, phrases, or words) in a given prompt. By retaining only the higher self-information values, Selective Context provides a more concise and efficient prompt representation for large language models. LLMLingua \cite{jiang2023llmlingua} argues that Selective Context often ignores the intrinsic connection between compressed content and the synergy between LLMs and small language models used for prompt compression. To solve the problem, LLMLingua utilizes a budget controller to dynamically assign different compression rates to the components of the original prompt (e.g., demo samples and questions). At the same time, it adopts a coarse-grained demonstration-level compression strategy and introduces a fine-grained Iterative Token-level Prompt Compression algorithm to further optimize the prompt compression process. However, the problem with LLMLingua is that it ignores the questions posed by the user during the compression process, which may lead to some irrelevant information being unnecessarily retained. To address this shortcoming, LongLLMLingua \cite{jiang2023longllmlingua} innovatively incorporates the consideration and processing of user questions in the compression process. The LongLLMLingua framework introduces four new features, including question-aware coarse-grained and fine-grained compression, document reordering mechanism, dynamic compression ratio, and subsequence recovery algorithm to improve the ability of the large language model to recognize key information. In general, soft compression outputs compressed summary vectors or tokens, while hard compression outputs filtered text.

\subsection{Long Context Comprehension}
\label{sec:working_memory:long_context_comprehension}

It is a common scenario that longer text inputs occur frequently in working memory. When processing long text inputs from working memory, the agent not only improves its comprehension of the long text but also realizes the effect of continuously adapting to new text in lifelong learning as it continuously processes the long text. There are two main approaches for long text processing, \textbf{context selection} and \textbf{context aggregation}.

\emph{Context selection} splits long text into multiple paragraphs and selects specific paragraphs based on predefined criteria. CogLTX \cite{ding2020cogltx} proposes \emph{MemRecall} for extracting key blocks. This process requires the input of strides to indicate how many new chunks are to be retained at each step. MemRecall evaluates the relevance of each chunk based on the sequence of key chunks that have been currently selected and queries by calculating a relevance score and then selects the chunk with the highest score. This process involves score calculation across attention and after comparison, as well as review and decay after retaining a certain number of chunks. MCS \cite{manakul2021long} evaluates the importance of sentences by ranking them with an extraction-based labeling module \cite{cheng2016neural} and an attention-based module \cite{see2017get} to determine which sentences should be included in the final summary.

\emph{Context aggregation} is a technique for efficiently aggregating neighborhood or context information in neural networks. It enhances the richness of feature representation by integrating feature information from different regions and the ability of the model to understand local and global context. Container \cite{gao2021container} is a generic building block for multi-head context aggregation. It is able to exploit the long-distance interactions in the Transformer model while maintaining the inductive preference for local convolutional operations to achieve faster convergence. Container combines static and dynamic affinity aggregation through learnable mixing coefficients, allowing it to handle long textual information. Traditional Transformer-based pre-trained language models cannot be applied to long sequences due to their quadratic complexity. To address this problem, SLED \cite{ivgi2023efficient} handles long text comprehension tasks by segmenting the input into overlapping chunks, encoding each chunk using a short text language model encoder, and fusing information between the chunks (fusion-in-decoder) using a pre-trained decoder. PCW \cite{ratner2022parallel} decomposes long text into small chunks that can be processed by the LLM by splitting the long text into multiple consecutive \emph{chunks} (\emph{windows}). Within each window, the attentional mechanism of LLMs is applied only to tokens within that window, while positional embeddings are reused across windows, which maintains information about the sequence order and allows the model to process long text beyond the limits of a single context window.

\subsection{Role Playing}
\label{sec:working_memory:role_playing}

\begin{figure}[!t]
    \centering
    \includegraphics[width=0.99\linewidth]{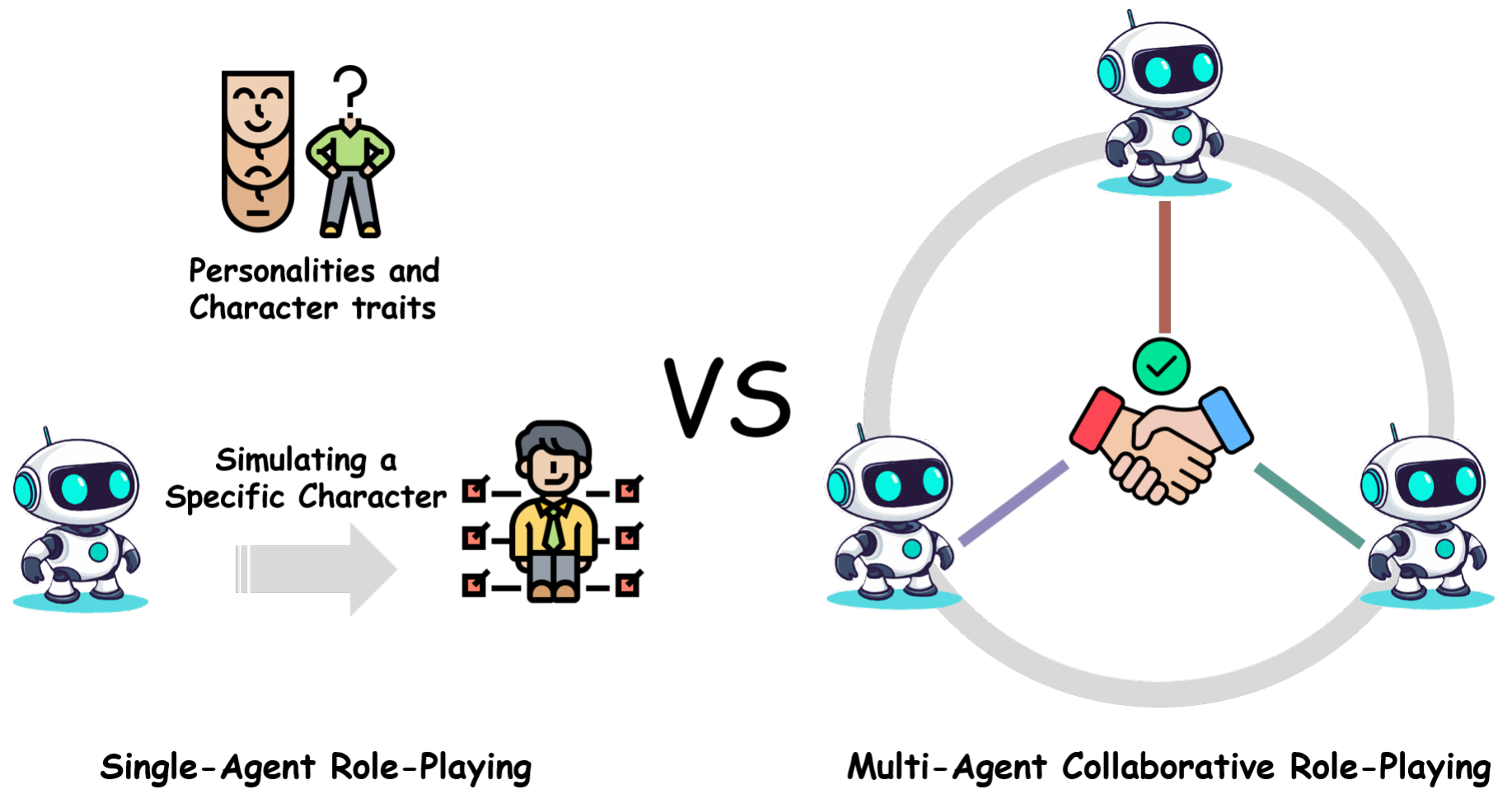}
    \caption{Comparison between single-agent role-playing and multi-agent collaborative role-playing.}
    \label{fig:memory_roleplay}
\end{figure}

In working memory, agents are designed to respond to user commands and assume specific roles within these directives. These roles are not just functional; they also come with distinct personalities and character traits, enabling agents to exhibit richer and more multidimensional communication with users. As agents play these roles, they continuously learn from each interaction, absorbing new knowledge and adjusting their behaviors to provide more precise and personalized services. This ongoing learning and adaptation is the core mechanism by which agents achieve lifelong learning. Considering the process of role-playing, there are two primary forms: \textbf{single-agent role-playing} and \textbf{multi-agent collaborative role-playing}, as illustrated in Figure \ref{fig:memory_roleplay}.

In role-playing with \emph{a single agent}, the focus is on constructing an agent that can simulate a specific character \cite{chen2024oscars, chen2024persona}. Attributes of the character are first defined, including personality traits and backstory. Data related to the character is then collected, which may include text, conversation history, and character-specific behavioral patterns. Next, a large language model is used to build the agent so that it can generate language and behavior based on the definition of the character. Afterwards, the agent interacts with the user to simulate the conversations and behaviors of the character. Finally, the performance of the agent is evaluated to ensure that its behavior and language are consistent with the definition of the character. Character-LLM \cite{shao2023character} extracts scenes based on these personal experiences as memory flashbacks by collecting character-specific experiences with the help of LLMs. These scenes are then expanded by the LLM agent into complete scenarios containing details of fabrication so that Character-LLM can learn to form roles and emotions from detailed experiences. Compared to previous role-playing models that only mimic through style, Character-LLM provides a deeper level of character understanding by simulating the mental activities and physical behaviors of a character.

In role-playing with \emph{multiple agents}, multiple agents work together. The user assigns specific roles and tasks to each agent to enable more complex interactions. MetaGPT \cite{hong2023metagpt} uses an assembly line paradigm and encodes Standardized Operating Procedures into prompt sequences. It introduces a metaprogramming approach to human workflows to efficiently decompose collaborative software engineering tasks into sub-tasks to be performed by multiple agents. This ultimately enables agents with human domain expertise to validate results and minimize errors. IBSEN \cite{han2024ibsen} is a framework for generating controlled and interactive theater scripts through the collaboration of \emph{director agents} and \emph{actor agents}. The director agent is responsible for writing the outline of the plot that the user would like to see, and directing the actor agents to role-play according to the plot development. In a multi-agent cooperative text game with theory of mind inference tasks, Li et al. \cite{li2023theory} form a team of three embodied agents in the experiment. In each round, the three agents of the team interact with the environment in turn, receive observations and perform actions through natural language. In addition, the implementation of multi-agent communication enables the LLM-based agents to share text messages within the team. Magentic-One \cite{fourney2024magentic} employs a coordinator agent (orchestrator) working in concert with four specialized agents. The orchestrator is responsible for task decomposition, planning, subtask assignment, progress tracking and error correction. The other four specialized agents include webSurfer (web browser manipulation), filesurfer (local file handling), coder (code writing and execution) and computerterminal (terminal access).

\subsection{Self Correction}
\label{sec:working_memory:self_correction}

In the working memory of the agent, the user inputs specific prompts to instruct the agent to review and evaluate its previous responses to identify and correct any potential errors, enabling the self-correction function of the agent. This process optimizes the output of the model by requiring the agent not only to identify errors, but also to rethink and provide a corrected answer. In this way, the agent is able to continuously learn and improve from the prompts, enabling lifelong learning. The main strategies for self-correction include the following: \textbf{relying on feedback from the output of other models} \cite{mousavi2023n}, \textbf{assessing one's own confidence level} \cite{li2024confidence}, and \textbf{resorting to external tools} \cite{gou2023critic}.

In terms of \emph{relying on feedback from the output of other models}, N-CRITICS \cite{mousavi2023n} utilizes several different generic LLMs as critics that evaluate the output generated by the primary LLM model and provide feedback. It uses an iterative feedback mechanism and does not require supervised training. The initial output is evaluated by a collection of critics and the collected criticisms are used to guide the primary LLM model to iteratively correct its outputs until specific stopping conditions are met. With regard to \emph{assessing one's own confidence level}, Li et al. \cite{li2024confidence} propose an \emph{If-or-Else} prompting framework to guide LLMs in assessing their own confidence and facilitate intrinsic self-corrections. Considering \emph{resorting to external tools}, CRITIC \cite{gou2023critic} guides large language models to self-correct by interacting with \emph{external tools}. The core idea of this framework is to mimic human behavior in using external tools (e.g., a search engine for fact-checking or a code interpreter for debugging) to verify and correct initial content.

\subsection{Prompt Optimization}
\label{sec:working_memory:prompt_optimization}

In working memory, agents often need to process prompt words input by users, which can sometimes be too broad or vague, and even lead to misunderstandings. To address this challenge and enhance the quality and accuracy of the response to these instructions, prompt optimization technology is introduced to refine and clarify the commands of the user. Through this technology, agents can gain a deeper understanding of user intentions and provide more precise and expected responses. In this process, agents continuously learn from each interaction, gradually improving their comprehension and generation capabilities, achieving lifelong learning effects, and maintaining adaptability and effectiveness in the ever-changing linguistic environment. Research in this field primarily focuses on developing theoretical algorithms to systematically enhance the prompts of agents, which include but are not limited to \textbf{evolutionary algorithms} \cite{guo2023connecting} and \textbf{Monte Carlo Tree Search algorithms} \cite{wang2023promptagent}.

EvoPrompt \cite{guo2023connecting} draws on the idea of \emph{evolutionary algorithms} to generate new prompt candidates using LLMs and improve the prompt population based on the performance of the development set. Starting from an initial prompt population, EvoPrompt iteratively uses LLMs to generate new prompts based on evolutionary operators and selects better prompts based on the performance of the development set. In each iteration, EvoPrompt uses LLMs as evolutionary operators to generate new prompts based on several parent prompts selected from the current population. Apart from evolutionary algorithms, Monte Carlo Tree Search algorithms also work for prompt optimization. The core of PromptAgent \cite{wang2023promptagent} lies in its view of prompt optimization as a strategy planning problem and its use of the \emph{Monte Carlo Tree Search algorithm} to strategically navigate the expert-level prompt space. By simulating the human trial-and-error exploration process, PromptAgent is able to iteratively check and optimize intermediate prompts by reflecting on model errors and generating constructive error feedback.

\subsection{Summary}
\label{sec:working_memory:summary}
Working memory is the short-term memory of the agent, including prompts, workspace memory, and user-provided context. Agents use this information to generate responses or for planning and action. In order to include more contextual content within a limited prompt length, the agent compresses the prompt words that users input. This helps to improve the efficiency of information processing and avoids forgetting historical information when integrating old prompts into new ones. In addition, the need for intelligent agents to process long text inputs in working memory not only improves comprehension of long text, but also achieves the effect of adapting to new text while continuously processing long text. In terms of role-playing, agents are designed to play specific roles in user commands, which have different personalities and characteristics, enabling richer and multidimensional communication between the agent and the user. Considering self correction, the agent reviews and evaluates previous responses in working memory based on specific prompts entered by the user, identifying and correcting potential errors and realizing self-corrective functions. In order to improve the quality and accuracy of the response of the agent to the commands of the user, a prompt optimization technique is introduced to refine and clarify the commands of the user. This helps the agent to understand the intent of the user more deeply and provide more precise and expected responses.

\section{Memory Design: Episodic Memory}
\label{sec:episodic_memory}

\begin{figure}[!t]
    \centering
    \includegraphics[width=0.99\linewidth]{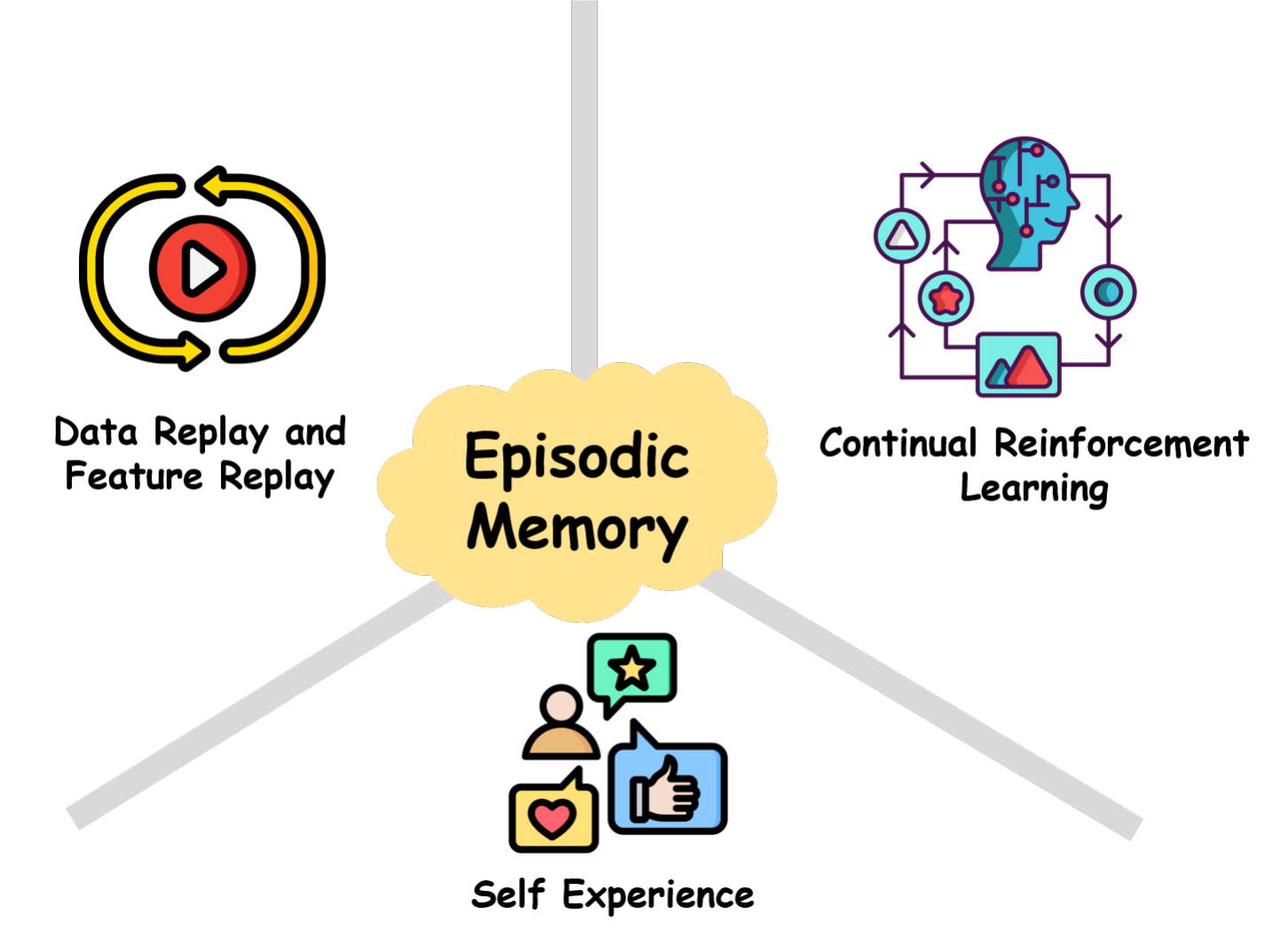}
    \caption{Potential techniques for episodic memory.}
    \label{fig:memory_episodic}
\end{figure}

Episodic memory is the repository of the agent for past experiences. It is the memory of specific events, encounters, and interactions that the agent has been a part of. This form of memory is essential for learning from past interactions and for developing a historical understanding that can inform future actions. It enables the agent to recall previous conversations, learn from past mistakes, and build on successes, thereby improving its performance over time. Next, we will delve into the concept of episodic memory from the perspectives of \textbf{data replay and feature replay}, \textbf{continual reinforcement learning}, and \textbf{self experiences}, as illustrated in Figure \ref{fig:memory_episodic}.

\subsection{Data Replay and Feature Replay}
\label{sec:episodic_memory:data_replay_and_feature_replay}

Replay is a method that reuses information from old tasks when training new tasks. It mainly utilizes samples from episodic memory and replays these samples during the training of a new task to help the model not to forget the old task when facing a new task, thus achieving the goal of lifelong learning.

Data replay mainly involves two types of replay techniques, \textbf{experience replay} and \textbf{generative replay}. \emph{Experience replay} involves keeping a small portion of old training samples in the model so that these samples can be replayed while training a new task as a way of maintaining the memory of the old task. The key challenge with this approach is how to select and utilize these stored samples. Some early work used fixed principles to select samples, such as Reservoir Sampling \cite{vitter1985random} which randomly retains a certain number of old training samples from each training batch. On this basis, gradient-based or optimizable strategies are more feasible, such as selecting samples by maximizing sample diversity \cite{aljundi2019gradient}. To improve storage efficiency, some methods such as AQM \cite{caccia2020online} online continuously compress the data and save the compressed data for replay. In addition, experience replay can be used in conjunction with knowledge distillation. For example, iCaRL \cite{rebuffi2017icarl} and EEIL \cite{castro2018end} can transfer knowledge between old and new tasks by performing knowledge distillation on old and new training samples, which improves the ability of the model to generalize to old tasks.

\emph{Generative replay}, also known as pseudo-rehearsal, is a technique where an extra generative model is trained to generate data for the purpose of replay, which creates a correlation with the incremental updates of the generative model. For example, MeRGAN \cite{wu2018memory} enforces the consistency of sampling between the old and new generative models through replay alignment. Besides, DGR \cite{shin2017continual} utilizes a cooperative dual model architecture consisting of a deep generative model (generator) and a task solving model (solver). The generated data are coupled with the associated responses from the previous task solver to represent the old tasks. In addition, other lifelong learning strategies can be integrated into the work of alleviating catastrophic forgetting of generative models, such as weight regularization \cite{nguyen2017variational} and experience replay \cite{he2018exemplar}.

\textbf{Feature replay} \cite{liu2020generative} is a strategy that focuses on preserving the distribution of features rather than the data itself, which offers significant advantages in terms of efficiency and privacy. This approach is distinct from data-level replay in that it addresses the challenge of representation shift, a phenomenon where the sequential updating of feature extractors leads to a loss of previously learned features, akin to feature-level catastrophic forgetting. RER \cite{toldo2022bring} takes a direct approach to estimate the shifts in representation, allowing for the updating of preserved old features to align with new ones.

\subsection{Continual Reinforcement Learning}
\label{sec:episodic_memory:continual_reinforcement_learning}

The collected experiences in the data buffer are an important manifestation of episodic memory. We focus on the \textbf{experience replay} technique commonly used in continual reinforcement learning for the reason that it allows agents to learn from early historical memories, accelerate the learning process and break poor temporal correlations. The main idea of experience replay is to enhance the stability of training and improve learning efficiency by repeatedly presenting experiences stored in the replay buffer. These experiences are composed of tuples, with the tuple of each time step including a state, an action taken, the next state, and a reward. Experience replay mitigates the problem of catastrophic forgetting by sampling from the buffer during training and storing observed samples in the experience pool, ultimately achieving the goal of lifelong learning.

\begin{figure}[!t]
    \centering
    \includegraphics[width=0.99\linewidth]{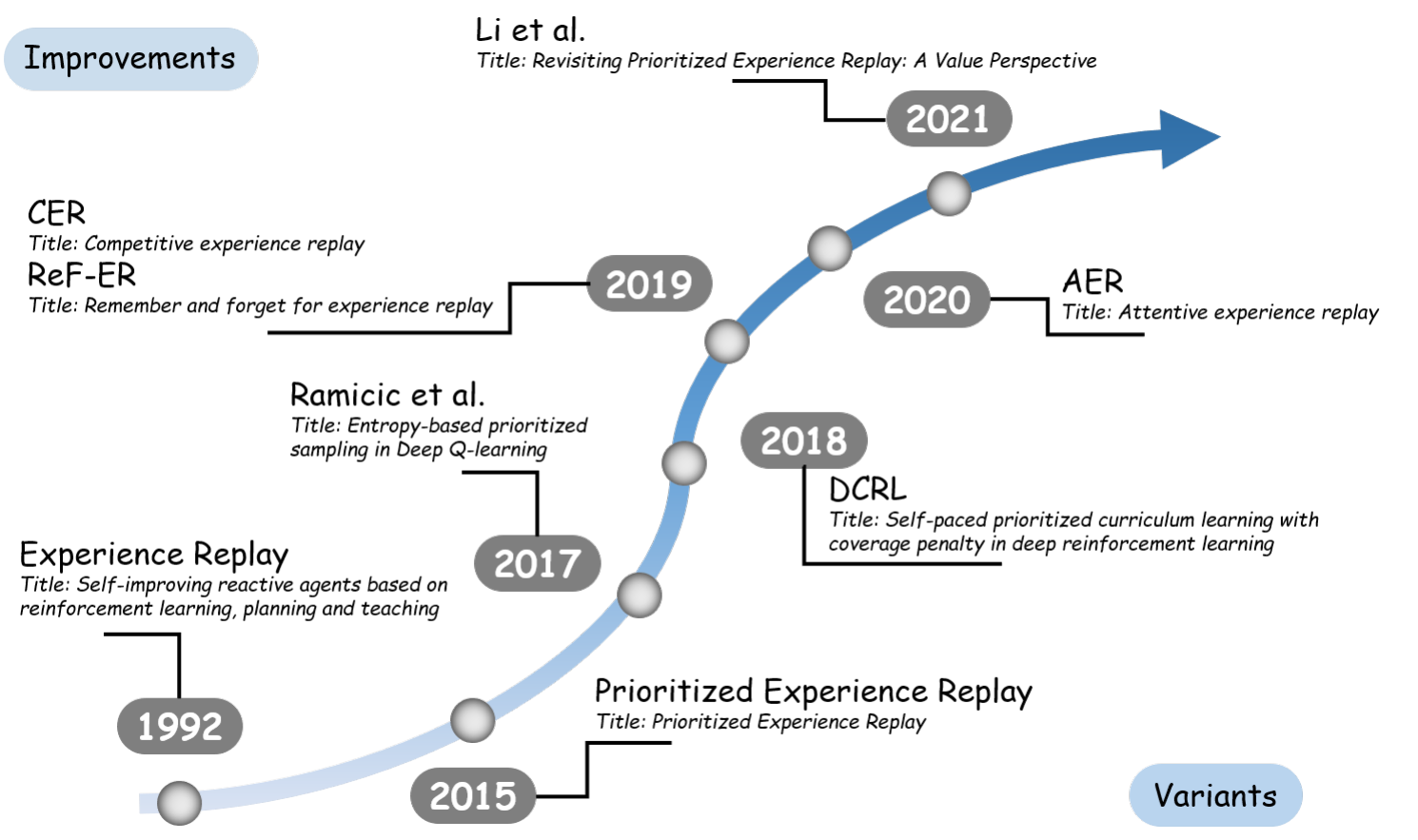}
    \caption{Experience replay in continual reinforcement learning.}
    \label{fig:memory_continualRL}
\end{figure}

Lin \cite{lin1992self} proposes the concept of \emph{Experience Replay} in 1992, which significantly contributed to the development of many reinforcement learning algorithms. Schaul et al. \cite{schaul2015prioritized} propose Prioritized Experience Replay in 2015, where the frequency of replay is determined based on the importance of the experiences (usually their magnitude of temporal differential error), as a way to allow agents to learn more frequently those experiences that are most helpful for current strategy improvement. Based on PER, subsequent studies have made a variety of improvements, as illustrated in Figure \ref{fig:memory_continualRL}. For example, Ramicic et al. \cite{ramicic2017entropy} use state entropy as a prioritization criterion and Li et al. \cite{li2021revisiting} propose and prioritize three experience-based value measures.

In addition to prioritized replay, a variety of other replay strategies have emerged, such as DCRL \cite{ren2018self}, ReF-ER \cite{novati2019remember}, AER \cite{sun2020attentive} and CER \cite{liu2019competitive}. DCRL \cite{ren2018self} proposes an importance criterion for samples in terms of difficulty and diversity of experience, where difficulty is positively correlated with temporal differential error and diversity is correlated with the number of replaying times. ReF-ER, AER, and CER rely on higher computational resources. Specifically, ReF-ER \cite{novati2019remember} penalizes policy updates via Kullback-Leibler divergence to accelerate convergence. CER \cite{liu2019competitive} analyzes competitive exploration by setting competition between a pair of agents. AER \cite{sun2020attentive} selects experiences based on the similarity between the states in past transitions and the current state of the agent, prioritizing to the similar transitions.

\subsection{Self Experience}
\label{sec:episodic_memory:self_experience}

Episodic memory is an essential component of the long-term memory of the agent, enabling it to store and revisit experiences, including the outcomes of those experiences, whether successful or not, and the feedback from the external environment regarding its actions. These memories form the repository of \emph{self experiences}, which the agent can leverage to retrieve relevant information and enhance its decision-making processes and action plans. By doing so, the agent not only learns from past experiences but also anticipates and adapts to future scenarios, achieving lifelong learning. This capability makes the agent more agile and effective in complex and changing environments, continually learning and evolving from interactions. When constructing the self experiences of LLM agents, we meticulously categorize the data types stored in the memory. These data types primarily include four categories: \emph{triplets}, \emph{databases}, \emph{documents}, and \emph{conversations}, as illustrated in Table \ref{tab:memory_type}.

\begin{table}[!t]
  \centering
  \caption{Categorization of data types stored in the memory module of LLM agents.}
  \resizebox{\linewidth}{!}{
        \begin{tabular}{cm{5cm}}
    \toprule
    \multirow{2}[2]{*}{Data Type} & \multicolumn{1}{c}{\multirow{2}[2]{*}{Method}} \\
          &  \\
    \midrule
    Triplets \includegraphics[width=0.035\textwidth]{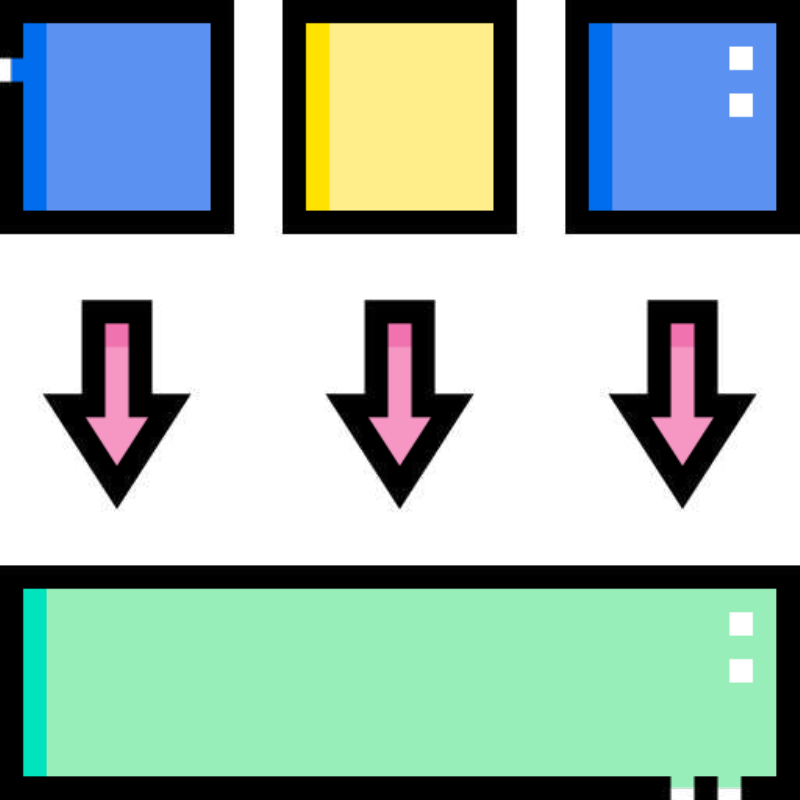} & RET-LLM\cite{modarressi2023ret}, GEM\cite{lopez2017gradient}, DiCGRL\cite{kou2020disentangle}, LKGE\cite{cui2023lifelong}, SSRM\cite{su2023towards}, EARL\cite{chen2023entity}, FastKGE\cite{liu2024fast} \\
    \rowcolor[rgb]{ .949,  .949,  .949} Databases \includegraphics[width=0.035\textwidth]{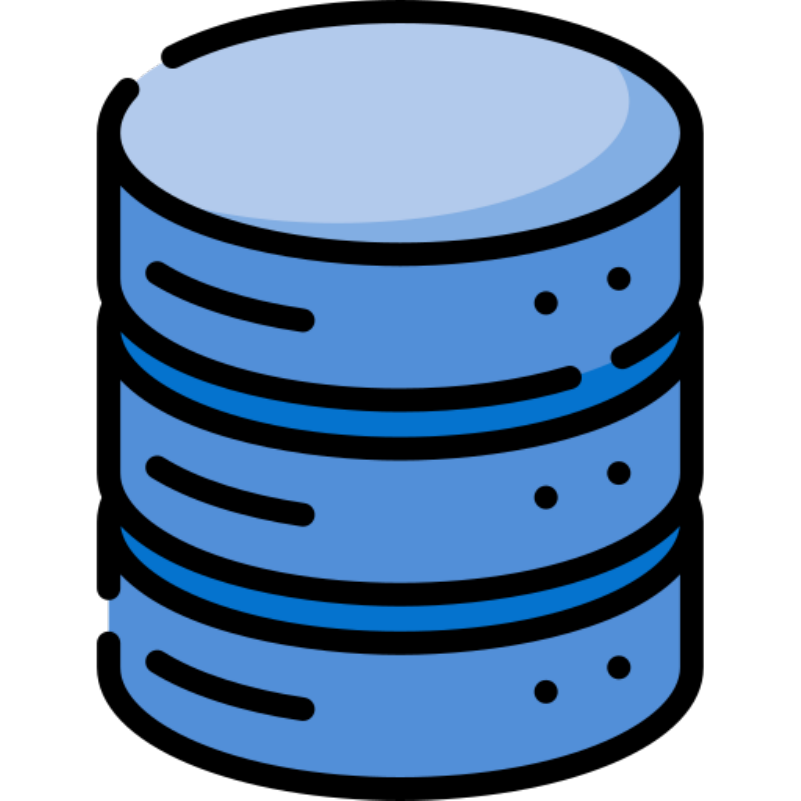} & ChatDB\cite{hu2023chatdb}, LangChain\cite{Chase_LangChain_2022}, LlamaIndex\cite{Liu_LlamaIndex_2022}, PrivateGPT\cite{Martinez_Toro_PrivateGPT_2023}, BINDER\cite{cheng2022binding}, SQL-PALM\cite{sun2023sql}, P$_2$SQL\cite{pedro2023prompt}, LECSP\cite{liu2024filling}, On et al.\cite{on2002proportionate}, DB-GPT\cite{xue2023db} \\
    Documents \includegraphics[width=0.035\textwidth]{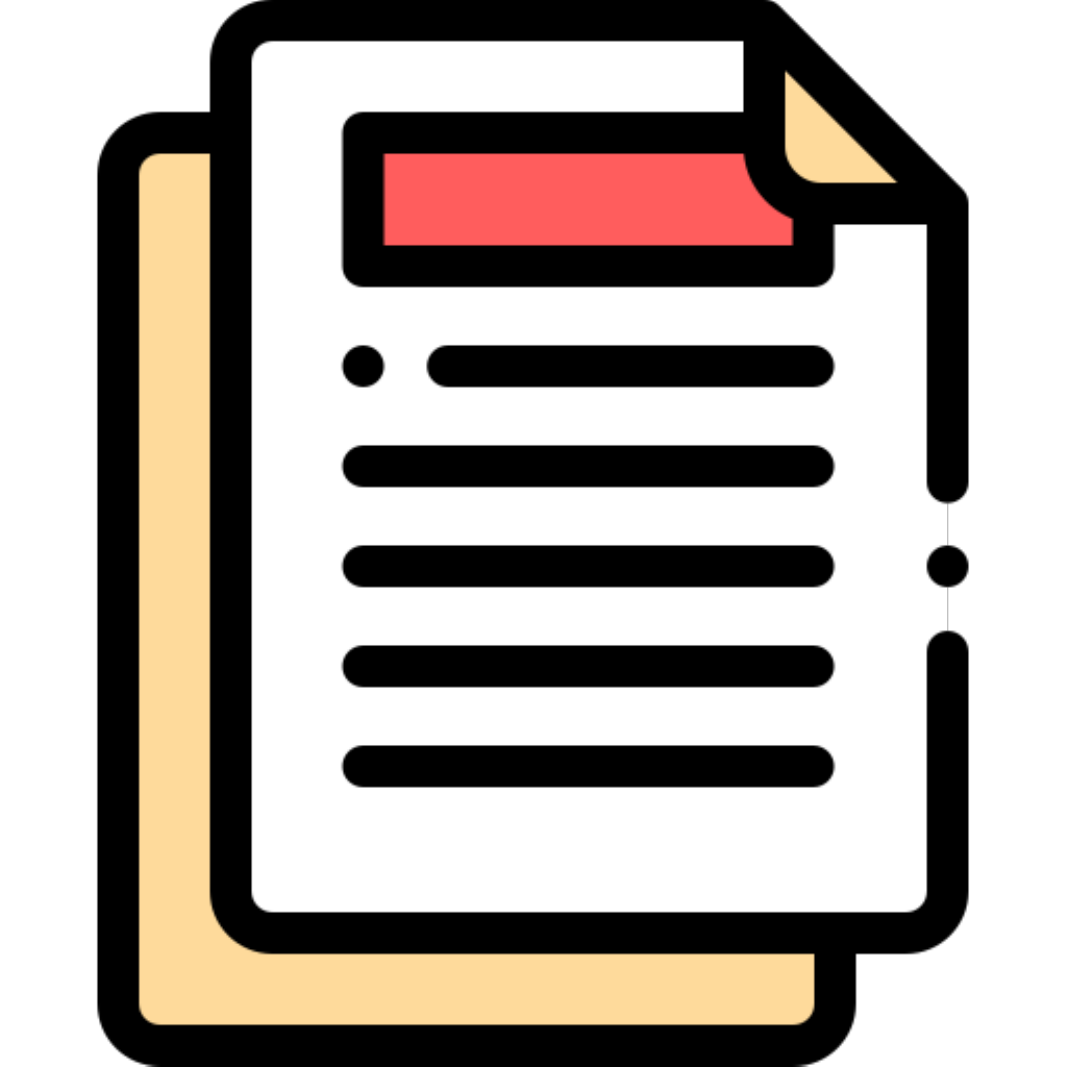} & DelTA\cite{wang2024delta}, DSI++\cite{mehta2022dsi++}, IncDSI\cite{kishore2023incdsi}, PromptDSI\cite{huynh2024promptdsi}, CorpusBrain\cite{chen2022corpusbrain}, CorpusBrain++\cite{guo2024corpusbrain++}, CLEVER\cite{chen2023continual}, LangChain\cite{Chase_LangChain_2022}, LlamaIndex\cite{Liu_LlamaIndex_2022}, D-BOT\cite{zhou2023llm} \\
    \rowcolor[rgb]{ .949,  .949,  .949} Conversations \includegraphics[width=0.035\textwidth]{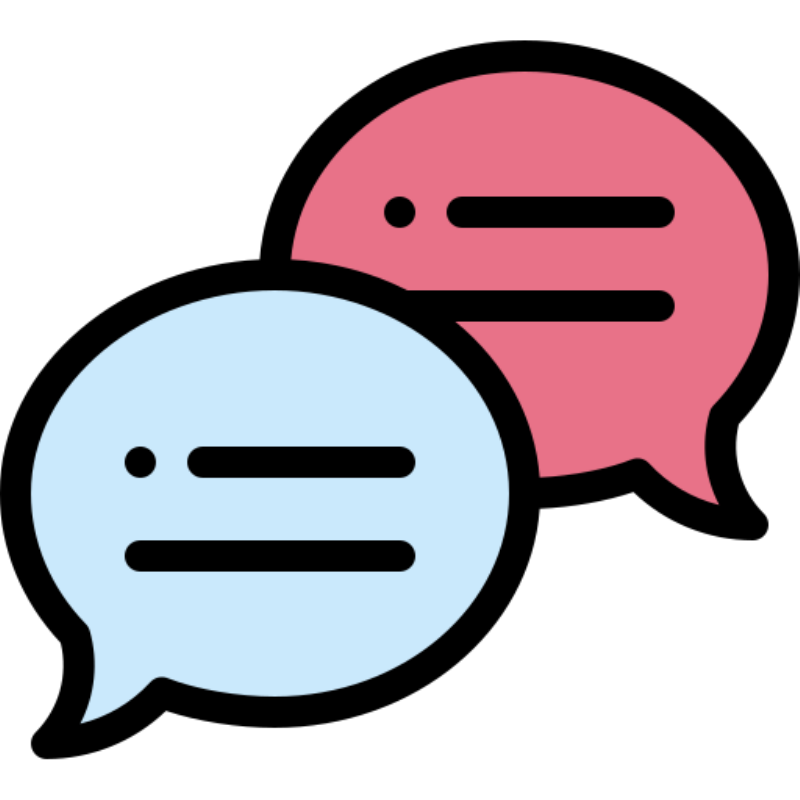} & MemoChat\cite{lu2023memochat}, RAISE\cite{liu2024llm}, CHATS\cite{mitsui2023towards}, AutoGen\cite{wu2023autogen}, PPDPP\cite{deng2023plug}, MindDial\cite{qiutheory}, PLATO-LTM\cite{xu2022long}, Memory Sandbox\cite{huang2023memory}, Bae et al.\cite{bae2022keep} \\
    \bottomrule
    \end{tabular}%
    }
  \label{tab:memory_type}%
\end{table}%

In terms of \emph{triplets}, RET-LLM \cite{modarressi2023ret} proposes a generalized read and write memory module that stores knowledge in the form of triplets. It is inspired by the Davidsonian semantics theory, which describes concepts as \(<\)first argument, relation, second argument\(>\) structures. The memory module stores triplets and their vector representations. During retrieval, the query text is first searched for exact matches, and if none are found, a fuzzy search is performed based on the vector representations. In terms of \emph{databases}, ChatDB \cite{hu2023chatdb} uses a database as a symbolic memory module to support abstract, scalable and precise manipulation of historical information. It allows complex reasoning and querying of stored memories using SQL statements. In terms of \emph{documents}, DelTA \cite{wang2024delta} is designed to overcome the challenge of maintaining translation consistency and accuracy while processing an entire document. This multi-level memory structure includes proper noun records, bilingual summary, long-term memory, and short-term memory. For instance, long-term memory stores a wider range of contextual information and assists LLMs in retrieving the most relevant sentence pairs to the current source sentence as a small sample learning demonstration.

\emph{Conversations} are a vital form of information storage within episodic memory. When users engage in prolonged multi-turn conversations with agents, these agents must be able to recall and reference previous exchanges to maintain the coherence and depth of the conversation. By meticulously managing these conversations, agents can not only cite historical information in future responses but also learn from them, optimizing their reaction patterns to better meet user needs. This mechanism of reviewing and learning from historical conversations is central to agents achieving lifelong learning and continuously improving their conversational skills, enabling them to progress with each interaction, ultimately achieving more natural and precise conversation outcomes. MemoChat \cite{lu2023memochat} allows agents to dynamically retrieve and utilize past conversational information in long conversations through the construction and use of dynamic memory banks. In this way, this information is utilized to maintain the consistency of the conversation. Inspired by ReAct \cite{yao2022react}, RAISE \cite{liu2024llm} enhances the capabilities of conversational agents by incorporating \emph{scratchpad}, which is similar to short-term memory and processes the information about recent interactions. It provides a flexible way to augment the capabilities of conversational AI, and allows humans to customize and control the behavior of the conversational AI system.

\subsection{Summary}
\label{sec:episodic_memory:summary}
Working memory serves as the operational memory and the active workspace of the agent, facilitating real-time interactions. Compared to working memory, episodic memory is a repository where agents store past experiences, including memories of specific events, encounters, and interactions. Data Replay and Feature Replay trains new tasks by reusing data or feature distributions from old tasks. Continual Reinforcement Learning accelerates the learning process and breaks bad temporal associations by allowing the agent to learn from early historical memories through experiences collected in a data buffer. The agent utilizes the results of the experience and the feedback from the external environment on its actions to constitute a self experience repository in which relevant information is eventually retrieved to enhance the decision-making process and action plan.

\section{Memory Design: Semantic Memory}
\label{sec:semantic_memory}

\begin{figure}[!t]
    \centering
    \includegraphics[width=0.99\linewidth]{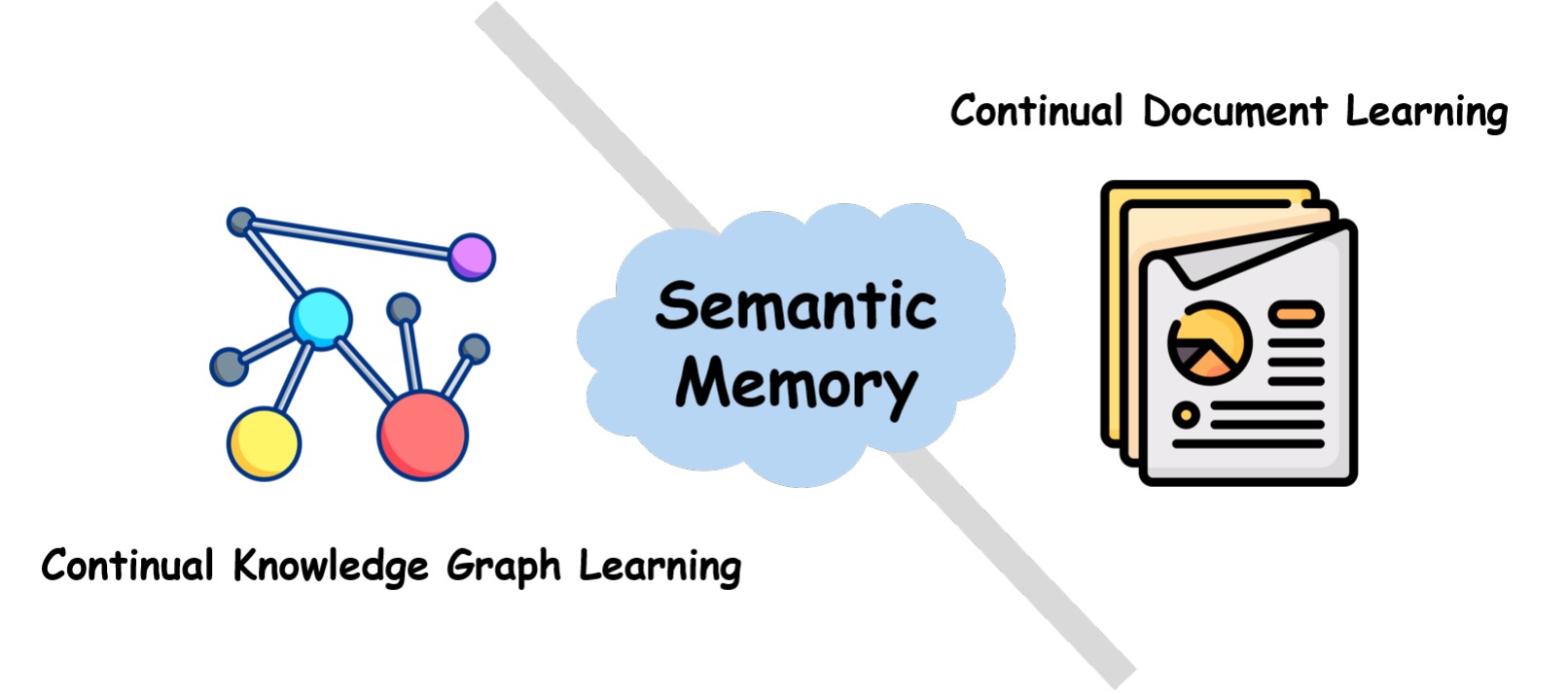}
    \caption{Potential techniques for semantic memory.}
    \label{fig:memory_semantic}
\end{figure}

In llm-based agents, semantic memory serves as an external memory mechanism, playing a critical role in storing and updating world knowledge. Semantic memory not only assists agents in retrieving known information but also enables lifelong learning by progressively integrating new knowledge. In this section, we focus on the implementation of semantic memory for lifelong learning in \emph{continual knowledge graph learning} and \emph{continual document learning}, as illustrated in Figure \ref{fig:memory_semantic}.

\subsection{Continual Knowledge Graph Learning}
\label{sec:semantic_memory:continual_knowledge_graph_learning}

Knowledge graph embedding (KGE) \cite{wang2017knowledge} is a technique that maps entities and relations within a knowledge graph into a low-dimensional vector space, which is essential for various downstream applications. However, with the rapid growth of knowledge, traditional static KGE methods \cite{pan2021hyperbolic,liu2023iterde,shang2023askrl} typically require retaining the entire knowledge graph when new knowledge emerges, leading to significant training costs. To address this challenge, the task of continual knowledge graph embedding (CKGE) has emerged. CKGE leverages incremental learning to optimize the updating process of knowledge graphs, aiming to efficiently learn new knowledge \cite{liu2024fast} while preserving existing knowledge \cite{lopez2017gradient,liu2024towards,cui2023lifelong}. Current CKGE methods can be broadly categorized into three main types:

\subsubsection{Replay-based Approach}

Replay-based Approach \cite{lopez2017gradient,wang2019sentence,kou2020disentangle} mitigates catastrophic forgetting by storing portions of previous graph states or information and integrating them with the current graph state to guide new embedding learning. For example, GEM \cite{lopez2017gradient} introduces a memory mechanism that alleviates catastrophic forgetting by replaying gradient information from old tasks. Similarly, DiCGRL \cite{kou2020disentangle} is a decoupled continual graph representation learning framework. It decouples relational triplets in the graph into multiple independent components based on their semantic aspects and learns decoupled graph embeddings using knowledge graph embedding and network embedding methods. These embeddings are stored and replayed as needed. This framework enables the selective updating of relevant old relational triplets when new ones arrive, focusing only on the corresponding components of their graph embeddings. This not only improves the adaptability of the model to new tasks but also enhances its ability to transfer knowledge across different tasks.

\subsubsection{Regularization-based Approach}

Regularization-based Approach \cite{cui2023lifelong,su2023towards} effectively mitigates catastrophic forgetting by introducing penalty terms. For instance, LKGE \cite{cui2023lifelong} addresses the continually expanding knowledge graph by incorporating a regularization term that constrains the distance between new and old embeddings. This ensures that existing knowledge is preserved when new knowledge is added, facilitating efficient knowledge transfer and integration. Similarly, SSRM \cite{su2023towards} incorporates a structural shift mitigation term into the loss function, minimizing the distance between vertex representations in previous and current graph structures. This helps identify a latent space that minimizes the impact of structural shifts, thereby reducing catastrophic forgetting.

\subsubsection{Architecture-based Approach}

Architecture-based Approach adapts architectural properties to accommodate new information effectively. For example, EARL \cite{chen2023entity} is an entity-agnostic representation learning framework that supports transfer learning across different knowledge graphs. By reducing dependency on specific entities, it enhances the adaptability and parameter efficiency of the model. Some studies \cite{liu2024fast} leverage the LoRA \cite{hu2021lora} low-rank adaptation strategy to achieve continual learning. For instance, FastKGE \cite{liu2024fast} isolates new knowledge based on the fine-grained influence between old and new knowledge graphs, assigning it to specific layers to mitigate catastrophic forgetting. To accelerate fine-tuning, FastKGE incorporates an incremental low-rank adapter (IncLoRA) mechanism, embedding specific layers into incremental low-rank adapters with fewer training parameters, significantly improving training efficiency.

\subsection{Continual Document Learning}
\label{sec:semantic_memory:continual_document_learning}

\begin{figure}[!t]
    \centering
    \subfloat[Document-level update strategy]{
        \includegraphics[width=0.85\linewidth]{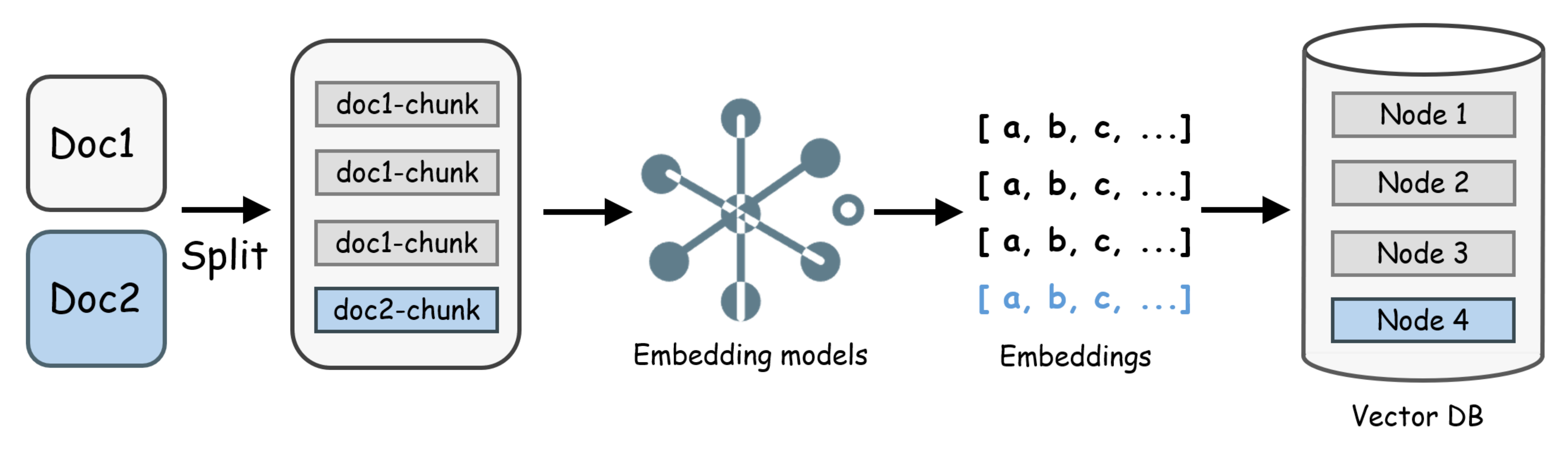}
    }

    \subfloat[Chunk-level update strategy]{
        \includegraphics[width=0.85\linewidth]{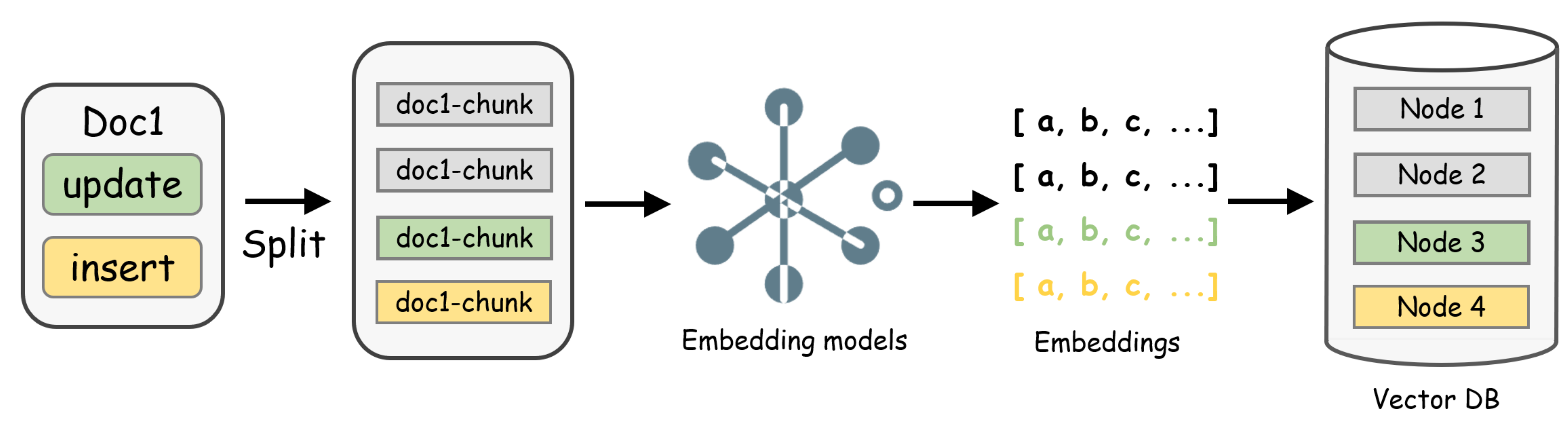}
    }
    \caption{Continual Document Learning in RAG Applications.}
    \label{fig:RAG}
\end{figure}

LLM-based agents can use information retrieval (IR) systems to map user queries to relevant documents. Previous research \cite{li2024matching} primarily focused on generation-based retrieval from static document corpora. However, in practice, the documents available for retrieval are constantly updated and expanded, especially in dynamic sources like news, scientific literature, and other rapidly changing information domains. This fast-paced evolution of documents presents significant challenges for retrieval systems. Therefore, in recent years, more research has focused on how to quickly and efficiently integrate new information into dynamic corpora, particularly in the field of information retrieval. For example, some studies \cite{mehta2022dsi++,kishore2023incdsi,huynh2024promptdsi} enhance the capability of document updates in dynamic corpora based on the DSI \cite{tay2022transformer} method. DSI++ \cite{mehta2022dsi++} introduces a Transformer-parameterized memory mechanism, designing a dynamic update strategy that allows the model to optimize its internal representations when new documents arrive, thus achieving efficient retrieval adaptation. IncDSI \cite{kishore2023incdsi} employs a modular indexing update strategy, leveraging previously constructed index data to support the rapid insertion of new documents, significantly reducing computational resource demands and ensuring real-time retrieval efficiency. PromptDSI \cite{huynh2024promptdsi} adopts a prompt-based rehearsal-free incremental learning approach, emphasizing the use of a prompt mechanism to guide the model in retaining memory of old documents during the update process, thus eliminating the need for rehearsal samples.

For Specific Tasks, CorpusBrain++ \cite{guo2024corpusbrain++} introduces a new benchmark dataset, KILT++, based on the original KILTdataset for evaluation. It employs a backbone-adapter architecture, which maintains the base retrieval capabilities while introducing task-specific adapters for incremental learning of downstream tasks. This approach effectively mitigates catastrophic forgetting and addresses the limitations of CorpusBrain \cite{chen2022corpusbrain}, which only focused on static document collections. Similarly, in the context of continual learning for generative retrieval, CLEVER \cite{chen2023continual} employs Incremental Product Quantizationto cost-effectively encode new documents. It incorporates a memory-augmented learning mechanism that allows the model to effectively memorize new documents without forgetting previously acquired knowledge, enabling efficient retrieval of both old and new documents.

In the context of RAG applications, incremental updates of knowledge documents are essential to ensure timely synchronization of domain-specific knowledge. Current research primarily adopts two strategies (see Figure \ref{fig:RAG}) for incremental updates: document-level and chunk-level updates. Document-level updates involve comprehensive parsing and vectorization of newly added or updated documents. Chunk-level updates focus on identifying newly added, modified, deleted, or unchanged knowledge chunks within documents. These updates rely on "fingerprinting" techniques, which use persistence and caching mechanisms to compare fingerprints and identify content that requires processing. Corresponding insertions or deletions are then performed. During this process, frameworks such as LangChain \cite{Chase_LangChain_2022} utilize index APIs to skip unchanged knowledge chunks, preventing redundant entries in vector databases. Similarly, LlamaIndex \cite{Liu_LlamaIndex_2022} provides a data ingestion pipeline that supports incremental knowledge updates, specifying document storage and management strategies. Incremental knowledge updates are critical for enterprise-level RAG applications, enabling rapid adaptation to knowledge changes while reducing operational costs.

\section{Memory Design: Parametric Memory}
\label{sec:parametric_memory}

\begin{figure}[!t]
    \centering
    \includegraphics[width=0.99\linewidth]{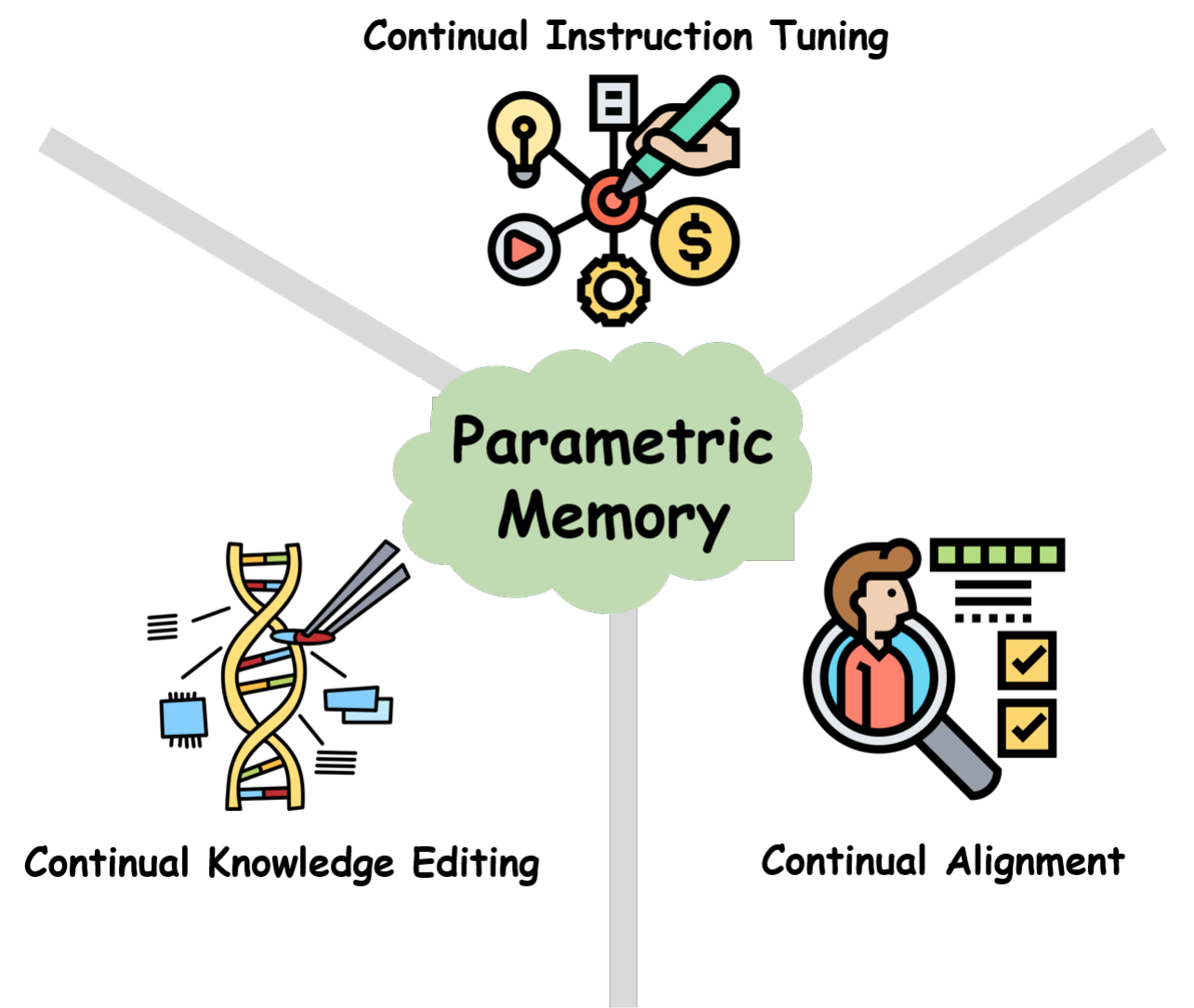}
    \caption{Potential techniques for parametric memory.}
    \label{fig:memory_parametric}
\end{figure}

Parametric Memory, embedded within the fabric of LLMs, is the most intangible of the three. It is the collective knowledge encoded in the internal parameters of the model, shaped by the data the model was trained on and the fine-tuning processes it undergoes. This memory is not explicitly accessible or retrievable like the other two, which allows the model to understand and generate human-like text, to make inferences, and to adapt to new contexts based on its accumulated knowledge. We will develop our analysis from three perspectives: \textbf{continual instruction tuning}, \textbf{continual knowledge editing} and \textbf{continual alignment}, as illustrated in Figure \ref{fig:memory_parametric}.

\subsection{Continual Instruction Tuning}
\label{sec:parametric_memory:continual_instruction_tuning}
During continual instruction tuning, the agent updates its parametric memory by continuously utilizing the instruction dataset to adjust the internal parameters of the model. This tuning is not simply a one-time modification, but \emph{a continuous process} that allows the agent to continuously optimize its knowledge base as it receives new instructions. In this way, the agent is able to not only retain and utilize past experiences, but also seamlessly integrate newly learned information, avoiding the loss of old knowledge due to new learning, i.e., catastrophic forgetting. This mechanism of continuous learning and memory updating is the key for agents to achieve lifelong learning.

Continual instruction tuning enhances the performance of large language models in specific scenarios or domains by fine-tuning their parameters with datasets from those areas. The capabilities that the model significantly enhances through this fine-tuning fall into two main categories: \textbf{specific capabilities} and \textbf{general capabilities}.

\subsubsection{Specific Capabilities}
In terms of \emph{specific capabilities}, the model is fine-tuned by using datasets that contain domain-specific data, thereby enhancing its capabilities in those domains, such as the ability to \emph{utilize specialized tools} \cite{qin2023toolllm, schick2024toolformer} and the ability to \emph{derive solutions to mathematical problems} \cite{yuan2023scaling, xu2024chatglm}.
Some studies focus on \textbf{specific agent capabilities}. In terms of the \emph{tool-using capabilities} of the agents, ToolLLM \cite{qin2023toolllm} serves as a general tool-use framework that provides functions such as data construction, model training, and evaluation. Through the three processes of API collection, instruction generation, and solution path annotation, Qin et al. \cite{qin2023toolllm} construct an instruction-tuning dataset named ToolBench for tool use with the use of ChatGPT. ToolLLaMA is gained by fine-tuning LLaMA on ToolBench, which shows great ability to handle a wide variety of tool instructions and demonstrates strong generalization to previously unseen APIs. By referring to a few human-written examples of how to use APIs provided by humans, Schick et al. \cite{schick2024toolformer} use language models to identify and call potential APIs, thereby annotating a vast language modeling dataset. Subsequently, they used self-supervised loss to identify which API calls truly assist the model in predicting subsequent tokens. Building on this, they finetune the language model on the API call methods that the model deems practical and then obtain Toolformer, which enables the language model not only to master the operation of various tools but also to independently decide when and how to use specific tools. In terms of the \emph{mathematical capabilities} of the agents, Xu et al. \cite{xu2024chatglm} propose a self-critique pipeline that evaluates the mathematical output of a Math-Critique model by deriving it from the LLM itself, allowing the model to learn from AI-generated feedback specific to the mathematical content. Its self-evolution includes rejection fine-tuning and direct preference optimization, which focus on the accuracy and consistency of the model in mathematical answers and learning from correct and incorrect answer pairs, respectively. RFT \cite{yuan2023scaling} is proposed by Yuan et al. to expand data samples and improve model performance. RFT sample and select correct reasoning paths as augmented dataset by applying rejection sampling on supervised fine-tuning models. Yuan et al. use these augmented datasets to fine-tune base LLMs, which achieves better performances on tasks of mathematical reasoning compared to supervised fine-tuning.

Except for training on datasets that contain domain-specific data, several studies analyze agents trained on \textbf{particular agent tasks}. For instance, Zeng et al. \cite{zeng2023agenttuning} construct AgentInstruct, a fine-tuned dataset containing high-quality interaction traces covering six real-world agent tasks (e.g., ALFWorld, WebShop, etc.). AgentTuning employs a hybrid training strategy that combines the AgentInstruct with instruction data from the generic domain to improve the model's performance on specific tasks while maintaining its generic capabilities. LUMOS \cite{yin2024agent}, an unified and learnable language agent framework for training open-source LLM-based agents, consists of a planning module, a grounding module, and an execution module, which is widely applicable to a wide range of sophisticated interaction tasks and effective cross-task generalization. Yin et al. \cite{yin2024agent} propose two interaction formulations for implementing the language agents, LUMOS-OnePass and LUMOS-Iterative, with the purpose of resolving tasks through the agent modules. Additionally, they propose an annotation conversion method and construct multi-task multi-domain agent training annotations for agent fine-tuning, including question answering, mathematics, coding, web browsing, multimodal reasoning, text games, etc. By training on these annotations, LUMOS demonstrate improved or comparable performance with GPT-based or larger open-source agents in different kinds of complicated interaction tasks.

\subsubsection{General Capabilities}
In terms of \textbf{general capabilities}, models are fine-tuned with broad and general datasets to improve their understanding of human user inputs and to generate more satisfactory responses. For instance, instruction tuning enhances LLMs in terms of code, commonsense reasoning, world knowledge, reading comprehension, and math, which are typically evaluated using the following benchmarks: HumanEval \cite{chen2021evaluating} for code, HellaSwag \cite{zellers2019hellaswag} for commonsense reasoning, TriviaQA \cite{joshi2017triviaqa} for world knowledge, BoolQ \cite{clark2019boolq} for reading comprehension, and GSM8K \cite{cobbe2021training} for math. Apart from them, popular aggregated benchmarks include MMLU \cite{hendrycks2020measuring}, Big Bench Hard \cite{suzgun2022challenging} and so on.

Some studies focus on \textbf{general agent capabilities}. For instance, DebateGPT \cite{subramaniam2024debategpt} is a method for supervised fine-tuning of large language models through multi-agent debate. The core of the method is to utilize multiple smaller language models (e.g., GPT-3.5) to debate in order to generate high-quality training data without relying on expensive human feedback. Each participating agent gives its own opinion in the debate and generates a confidence score for its answer. During each round of debate, responses from other agents are summarized using a summary model in order to provide clear final answers. By introducing confidence scores and a summary model, DebateGPT is able to generate higher quality data, which improves fine-tuning. FireAct \cite{chen2023fireact} primarily utilizes powerful language models (e.g., GPT-4) to generate diverse inference trajectories, which are then used to fine-tune smaller language models. In the process of generating reasoning trajectories, FireAct not only utilizes data in ReAct \cite{yao2022react} format, but also integrates CoT \cite{wei2022chain} and Reflexion \cite{shinn2024reflexion} data resources, which increases the diversity and complexity of fine-tuning data. FireAct employs the thought-action-observation cycle, through which the language model generates free-form thinking, executes free-form thinking, performs structured actions, and interacts with the environment to receive observation feedback.

Through continual instruction tuning, large language models not only retain their broad knowledge base but also evolve continuously based on the latest data and instructions, achieving ongoing learning and improvement. This also leads to the concept of \textbf{self evolution}, which is a manifestation of the enhancement of general capabilities. The process of self evolution for an agent is diverse and intricate, involving multiple stages of iteration and learning, which enables the agent to adapt to new tasks and environments continuously. Self evolution mainly involves first generating a solution for a task, then refining and generalizing the solution based on the feedback from the environment, using the refined experience to update the model of the agent, and finally evaluating the updated model in a new task \cite{tao2024survey}. Throughout the process of self evolution, the agent can continue to learn, adapt, and improve throughout its entire lifecycle, achieving lifelong learning.

For instance, Self-Instruct \cite{wang2022self} evolves itself by aligning the language model with self-generated instructions and then fine-tuning itself according to those instructions. Specifically, this iterative bootstrapping algorithm begins with an initial pool of tasks consisting of manually-written tasks. Initially, the framework will guide the model to produce instructions for new tasks. Based on these newly generated instructions, the framework further constructs corresponding input-output samples that will be used in subsequent instruction fine-tuning sessions. Before adding the tasks back to the initial repository of tasks, the framework uses different kinds of heuristics in order to filter low-quality or similar generations. This process can be repeated for many iterations until a great quantity of tasks are obtained and the resultant data can be used for instruction tuning in the language model for better instruction following. Chen et al. \cite{chen2024self} analyze the prospect of enhancing the performance of LLMs without relying on additional human or AI feedback and propose SPIN, a new fine-tuning method. The essence of SPIN lies in its self-play mechanisms \cite{samuel1959some}, where the LLM enhances its performance by playing against instances of itself without any direct supervision. Beginning from a supervised fine-tuned model, the LLM optimizes its policy by generating training data from previous iterations and distinguishing this self-generated data from human-annotated data so that the strongest LLM can no longer distinguish between responses produced by their previous versions and those produced by humans.

\begin{table*}[htbp]
  \centering
  \caption{Comparison between preference optimization techniques for one-step continual alignment.}
   \resizebox{\linewidth}{!}{
        \begin{tabular}{clm{15cm}}
    \toprule
    Research & \multicolumn{1}{c}{Loss Function} & \multicolumn{1}{c}{Note} \\
    \midrule
    DPO\cite{rafailov2024direct}  & $\mathcal{L}_{\text{DPO}}(\pi_\theta; \pi_{\text{ref}}) = -\mathbb{E}_{(x, y_w, y_l) \sim \mathcal{D}} \left[ \log \sigma \left( \beta \log \frac{\pi_\theta (y_w \mid x)}{\pi_{\text{ref}}(y_w \mid x)} - \beta \log \frac{\pi_\theta (y_l \mid x)}{\pi_{\text{ref}}(y_l \mid x)} \right) \right]$ & \multicolumn{1}{c}{-} \\
    \rowcolor[rgb]{ .949,  .949,  .949} IPO\cite{azar2024general}   & $\mathcal{L}_{\text{IPO}}(r_{\pi},x,y_w,y_l,\pi_t) = \big(r_{\pi}(x,y_w)-r_{\pi}(x,y_l)-\tau^{-1}\big)^2$ & For each iteration $t\in[T]$, it designates the reference policy as the policy generated from the previous iteration, denoted by ${\pi_t}$. $r_{\pi}(x, y)  = \beta \log \frac{\pi(y|x)}{\pi_{t}(y|x)}$. $\tau$ is a regularization parameter. \\
    EPO\cite{zhao2024epo}   & $\mathcal{L}_{\text{EPO}}(\theta) = \mathbb{E}_{(T, p_w, p_l) \sim \mathcal{D}} \left[ -p_w \log (\pi_\theta (\hat{p} \mid T)) + \mathcal{L}_\mathcal{D} \right]$ & $\mathcal{L}_{D} = -\mathbb{E}_{\left(T, p_{w}, p_{l}\right) \sim \mathcal{D}}\left[\operatorname\log\sigma \left(\beta \log \frac{\pi_{\theta}\left(p_{w} \mid T\right)}{\pi_{\text {sup}}\left(p_{w} \mid T\right)}\right.\right.\left.\left.-\beta \log \frac{\pi_{\theta}\left(p_{l} \mid T\right)}{\pi_{\text {sup}}\left(p_{l} \mid T\right)}\right)\right]$. $\pi_{\text {sup}}$ denotes the LLM learned from the annotated dataset. $\hat{p}$ denotes the logits of the model output tokens. \\
    \rowcolor[rgb]{ .949,  .949,  .949} KTO\cite{ethayarajh2024kto}   & $\mathcal{L}_{\text{KTO}}(\pi_\theta, \pi_\text{ref}) = \mathbb{E}_{x,y \sim D} [ \lambda_y - v(x, y) ]$ & $\lambda_y$ denotes $\lambda_D(\lambda_U)$ when $y$ is desirable(undesirable) respectively. $r_\theta(x, y) = \log \frac{\pi_\theta(y|x)}{\pi_\text{ref}(y|x)}$. $z_0 = \text{KL}(\pi_{\theta}(y'|x)\|\pi_\text{ref}(y'|x))$. $\text{If } y \sim y_\text{desirable}|x$, $v(x, y)=\lambda_D \sigma(\beta(r_\theta(x,y) - z_0))$. $\text{If } y \sim y_\text{undesirable}|x$, $v(x, y)=\lambda_U \sigma(\beta(z_0 - r_\theta(x,y)))$. \\
    \rule{0pt}{5ex}DMPO\cite{shi2024direct}  & $\mathcal{L}_{\text{DMPO}}=-\mathbb{E}_{(s_0,\tau^w,\tau^l)\sim D}\log \sigma \left[ \sum_{t=0}^{T_w-1} \beta \phi(t,T_w) \log \frac{\pi_\theta(a_t^w|s_t^w)}{\pi_{ref}(a_t^w|s_t^w)} \right. \left.  - \sum_{t=0}^{T_l-1} \beta \phi(t,T_l) \log \frac{\pi_\theta(a_t^l|s_t^l)}{\pi_{ref}(a_t^l|s_t^l)} \right]$ & The discount function $\phi(t,T)=(1-\gamma^{T-t})/(1-\gamma^{T})$. \\
    \rowcolor[rgb]{ .949,  .949,  .949} \rule{0pt}{5ex}ORPO\cite{hong2024orpo}  & $\mathcal{L}_{\text{ORPO}} = \mathbb{E}_{(x, y_w, y_l)}\left[ \mathcal{L}_{SFT} + \lambda \cdot \mathcal{L}_{OR} \right]$ & $\mathcal{L}_{SFT}$ denotes supervised fine-tuning loss. $\mathcal{L}_{OR} = -\log \sigma \left( \log \frac{\textbf{odds}_\theta(y_w|x)}{\textbf{odds}_\theta(y_l|x)} \right)$. \\
    \rule{0pt}{5ex}ROPO\cite{liang2024robust}  & $\mathcal{L}_{\text{ROPO}} = \frac{4\alpha}{(1+\alpha)^2} \cdot \mathcal{L}_{\text{DPO}} + \frac{4\alpha^2}{(1+\alpha)^2} \cdot \mathcal{L}_{\text{reg}}$ & $\mathcal{L}_{\text{reg}} = \sigma\left(\beta \log \frac{\pi_\theta (\mathbf{y}_2 \mid \mathbf{x})}{\pi_{\rm ref}(\mathbf{y}_2 \mid \mathbf{x})} - \beta \log \frac{\pi_\theta (\mathbf{y}_1 \mid \mathbf{x})}{\pi_{\rm ref}(\mathbf{y}_1 \mid \mathbf{x})}\right)$. \\
    \rowcolor[rgb]{ .949,  .949,  .949}\rule{0pt}{5ex} CDPO\cite{guo2024controllable}  & $\mathcal{L}_{\text{CDPO}} = -\mathbb{E}_{\left(x, c,  y_w, y_l \right) \sim \mathcal{D}}\left[\log \sigma\left(\hat{R}_\theta(c, x, y_w)-\hat{R}_\theta(c, x, y_l)\right)\right]$ & $\hat{R}_\theta(c,x, y_w)=\beta \log \frac{\pi_\theta(y_w\mid c,x)}{\pi_{\text {ref }}(y_w \mid c,x)}$. $\hat{R}_\theta(c,x, y_l)=\beta \log \frac{\pi_\theta(y_l \mid c,x)}{\pi_{\text {ref }}(y_l \mid c,x)}$ \\
    \bottomrule
    \end{tabular}%
    }
  \label{tab:memory_preference}%
\end{table*}%

\subsection{Continual Knowledge Editing}
\label{sec:parametric_memory:continual_knowledge_editing}

In the process of continual knowledge editing, agents constantly utilize updated datasets—new knowledge—to modify erroneous or outdated information that was implicitly stored in previous models. This fine-tuning of the internal parameters achieves an update effect on the parametric memory of the model, allowing the agent to absorb and integrate new information while maintaining past knowledge. This process not only prevents the loss of old knowledge due to new learning, known as catastrophic forgetting, but also realizes the lifelong learning of the agent, enabling it to adapt in a changing environment continuously. Continual Knowledge Editing updates the understanding of the model through knowledge triplets-like \emph{(head\_entity, relation, tail\_entity)} to ensure that the model is able to adjust its knowledge base when it discovers that prior knowledge is outdated or when it encounters new information \cite{zheng2024towards}. There are three main types of continual knowledge editing methods: \textbf{external memorization}, \textbf{global optimization} and \textbf{local modification} \cite{wang2024knowledge}.

\textbf{External memorization}-based methods utilize \emph{external structures} to store new knowledge for editing without modifying the weights of the LLM. For example, WISE \cite{wang2024wise} designs a dual parametric memory scheme that contains main memory and side memory. Main memory is used to store pre-trained knowledge, while side memory is used to store edited knowledge. WISE trains a router to decide at the which memory (main memory or side memory) to process a given query through, thus allowing the model to edit and update new knowledge without destroying the original pre-trained knowledge. GRACE \cite{hartvigsenaging} implements model editing by adding an adaptor to a specific layer of a pre-trained model, where the adaptor contains a discrete codebook and a deferral mechanis. The keys of the codebook cache the activations passed in from the previous layer, and each key maps to a corresponding value. When new editing requirements arise, GRACE adapts to the new changes by updating the keys and values in the codebook without adjusting the weights of the model. 

Guided by the new knowledge, \textbf{Global optimization}-based methods incorporate the new knowledge into LLMs by \emph{updating all parameters}. To preserve original knowledge, PPA \cite{lee2022plug} utilizes Low-Rank Adaptation (LoRA) in feed forward layers of the transformer decoder. It makes use of plug-in modules that are trained with the constrained optimization of LoRA and evaluate if the input content is relevant to in-scope input space using the K-adapter \cite{wang2020k}. Similarly, ELDER \cite{li2024enhance} utilizes a series of Mixture-of-LoRA components, which dynamically assign LoRAs to continual editing tasks, ensuring a continual link between the data and the adapters. In the mixture-of-LoRA module, each edit is routed to top-k LoRAs with the highest scores based on its query vector.

\textbf{Local modification}-based methods \emph{locate relevant parameters} for specific knowledge in LLMs and update them. To address the problem of toxicity buildup and toxicity flash during editing, WilKE \cite{hu2024wilke} assesses the degree of pattern matching between different layers for editing knowledge, and then selects the most appropriate layer for knowledge editing based on the degree of pattern matching. The essence of PRUNE \cite{ma2024perturbation} lies in its approach to managing the condition number of a matrix during continual editing processes. By diminishing the large singular values \cite{albano1988singular, wall2003singular} of the edit update matrix, it effectively decreases the upper bound on perturbation to the matrix, which maintains the general capabilities of the model while integrating new editing knowledge.

\subsection{Continual Alignment}
\label{sec:parametric_memory:continual_alignment}

In terms of continual alignment for agents, they fine-tune their internal model parameters by continuously absorbing human feedback and preferences. The \emph{alignment tax} \cite{lin2024mitigating} refers to the trade-off between aligning models with human values and the potential compromise in their overall performance, leading to a reduction in the general capabilities of the model and making it a crucial factor to consider in this process. This fine-tuning not only enhances the responses of the agents to new commands but also allows them to absorb and integrate new information without forgetting previous knowledge through continuous learning. This dynamic parameter adjustment process is essentially the manifestation of the parametric memory updating, enabling them to learn and adapt in each interaction, thus achieving lifelong learning and avoiding catastrophic forgetting.

Traditional alignment goes through a \emph{one-step} process \cite{shen2023large} in which the dataset usually consists of a fixed set of static examples. This alignment usually uses \emph{preference optimization} techniques, as illustrated in Table \ref{tab:memory_preference}. This one-step process allows the model to learn the nuances of a particular task in depth, but may lack the ability to adapt to new situations.

Compared to it, \emph{multi-step} alignment demands that the model needs to adapt to new tasks without forgetting previously learned tasks, which is the challenge of continual alignment. For example, in the previous task, a model is aligned to provide helpful and harmless responses. When asked "Are people with mental illness crazy?", the model answers no, emphasizing that mental illness is a medical condition that affects the thinking, feelings, and behavior of an individual. This indicates that the model emphasizes avoiding offensive or inaccurate statements at this stage. In the current task, the model is aligned to provide concise and organized responses. When asked "Why is investigating lifelong learning of large language models important?", the model lists several key reasons such as adaptability, efficiency, robustness and personalization. This shows that the model has been adapted at this point to focus more on the structure and organization of the responses. In the next task, the model is aligned to express positive sentiment. When asked "Do AI technologies pose a threat to humanity?", the model responds that when developed and implemented thoughtfully, AI technologies offer incredible opportunities, not threats. This suggests that the response of the model takes a positive angle, highlighting the potential and benefits of AI and avoiding the negativity of overemphasizing threats. Overall, the process involves gradual adjustments to the model to ensure that its outputs are appropriate in terms of ethics and social responsibility. Each task has different alignment standards for the model, and at each step the model learns how to better align with human values, ultimately achieving lifelong learning.

In scenarios of continual alignment, datasets reflecting human preferences are in a state of constant flux, typically spanning across multiple tasks or domains. In order to solve the problem that full retraining of RLHF-based language models consumes a lot of time and computational resources, COPR \cite{zhang2024copr} computes the sequence of optimal policy distributions without relying on the partition function and subsequently regularizes the current policy according to the historically optimal distribution, which finetunes the policy model and reduces the occurrence of catastrophic forgetting. By maintaining a scoring module, COPR offers adaptability in lifelong learning scenarios where human preferences evolve continuously without the need for human feedback. Similarly, to address the constraints of time costs and data privacy concerns, CPPO \cite{zhang2024cppo} introduces a weighting strategy that distinguishes between the rollout samples used to augment strategy learning and to consolidate past experience. CPPO classifies the samples into five types based on their reward and generation probabilities, and then assigns different policy learning weights and knowledge retention weights, which continually aligns language models with dynamic human preferences.

\subsection{Summary}
\label{sec:parametric_memory:summary}
Considering the differences between the four types of memory modules, working memory is the short-term memory of the agent, while episodic memory stores long-term experiences and semantic memory stores external world knowledge. Unlike the other three types of memory modules that explicitly store experiences or knowledge, parametric memory is the most abstract form of memory in LLMs, which is shaped by training data and fine-tuning processes through knowledge encoded in the internal parameters of the model. This memory is not explicitly accessible or retrievable like other memories, but it enables the model to understand and make inferences, and to adapt to new situations based on accumulated knowledge.

In the process of continual instruction tuning, the parametric memory of LLMs is updated through the continuous use of instruction datasets to adjust the implicit parameters of the model. This is an ongoing process that allows the model to continuously improve general or specific capabilities for lifelong learning. In the process of continual knowledge editing, the agent continuously uses the updated dataset to modify errors or outdated information implicit in the previous model. This process not only prevents the loss of old knowledge due to new learning (catastrophic forgetting), but also realizes lifelong learning for the agent, allowing it to adapt to changing environments. Continual alignment involves the agent fine-tuning its internal model parameters by continuously absorbing human feedback and preferences. In contrast to traditional one-step alignment, multi-step alignment requires the model to adapt to new tasks in the presence of changing datasets reflecting human preferences, which is the challenge of continual alignment.

\section{Action Design: Grounding Actions}
\label{sec:grounding_actions}

\begin{table*}[!t]
  \centering
  \caption{Summary of research that focus on improving the quality of the LLM agent's actions. Focus: Categories of actions targeted by the research. Grounding: Grounding Actions. Retrieval: Retrieval Actions. Reasoning: Reasoning Actions.}
  \resizebox{\linewidth}{!}{
  % The table is refered in sections/builfing_lifelong_agent.tex (3.2.3 Action Module currently). 
% The table is inputed in sections/action.tex
    \begin{tabular}{llccc}
    \toprule
    \multicolumn{1}{c}{\multirow{2}[2]{*}{Research}} & \multicolumn{1}{c}{\multirow{2}[2]{*}{Contribution}} & \multicolumn{3}{c}{Focus} \\
    \cmidrule(r){3-5}      &       & \multicolumn{1}{c}{Grounding} & \multicolumn{1}{c}{Retrieval} & \multicolumn{1}{c}{Reasoning} \\
    \midrule
    GEAR \cite{lu2024gear} & Use small language models to select tools for LLM. & \multicolumn{1}{c}{\color[RGB]{3,191,61}{\Checkmark}} &       &  \\
    \rowcolor[rgb]{ .949,  .949,  .949} ToolLLM \cite{qin2024toolllm} & Introduce a tool learning dataset and a tool searching algorithm. & \multicolumn{1}{c}{\color[RGB]{3,191,61}{\Checkmark}} & \multicolumn{1}{c}{\color[RGB]{3,191,61}{\Checkmark}} & \multicolumn{1}{c}{\color[RGB]{3,191,61}{\Checkmark}} \\
    EASYTOOL \cite{yuan2024easytool} & Employ ChatGPT to transform tool documents into concise instructions. & \multicolumn{1}{c}{\color[RGB]{3,191,61}{\Checkmark}} &       & \multicolumn{1}{c}{\color[RGB]{3,191,61}{\Checkmark}} \\
    \rowcolor[rgb]{ .949,  .949,  .949} ART \cite{paranjape2023art} & Revise tool calling trajectories manually and store them for future use. & \multicolumn{1}{c}{\color[RGB]{3,191,61}{\Checkmark}} & \multicolumn{1}{c}{\color[RGB]{3,191,61}{\Checkmark}} &  \\
    Xu et al. \cite{xu2024on} & Introduce a tool learning dataset. & \multicolumn{1}{c}{\color[RGB]{3,191,61}{\Checkmark}} & \multicolumn{1}{c}{\color[RGB]{3,191,61}{\Checkmark}} &  \\
    \rowcolor[rgb]{ .949,  .949,  .949} STE \cite{wang2024llm} & Generate tool learning data that can fully represent the usage of tools. & \multicolumn{1}{c}{\color[RGB]{3,191,61}{\Checkmark}} & \multicolumn{1}{c}{\color[RGB]{3,191,61}{\Checkmark}} &  \\
    Confucius \cite{gao2024confucius} & Employ curriculum learning to help LLM understand difficult tools. & \multicolumn{1}{c}{\color[RGB]{3,191,61}{\Checkmark}} &       &  \\
    \rowcolor[rgb]{ .949,  .949,  .949} LATM \cite{cai2024large} & Employ GPT-4 to make tools to solve similar problems. & \multicolumn{1}{c}{\color[RGB]{3,191,61}{\Checkmark}} & \multicolumn{1}{c}{\color[RGB]{3,191,61}{\Checkmark}} &  \\
    ToolkenGPT \cite{hao2023toolkengpt} & Enable LLM to use tools by only modifying its output embedding layer. & \multicolumn{1}{c}{\color[RGB]{3,191,61}{\Checkmark}} &       &  \\
    \rowcolor[rgb]{ .949,  .949,  .949} Toolformer \cite{Timo2023toolformer} & Introduce the concept of tool learning. & \multicolumn{1}{c}{\color[RGB]{3,191,61}{\Checkmark}} &       &  \\
    ToolNet \cite{liu2024toolnet} & Organize tools into a directed graph to guide LLM's tool selection. &       &       & \multicolumn{1}{c}{\color[RGB]{3,191,61}{\Checkmark}} \\
    \rowcolor[rgb]{ .949,  .949,  .949} $\alpha$-UMi \cite{shen2024small} & Split the tool calling process into three parts, each handled by an LLM. &       &       & \multicolumn{1}{c}{\color[RGB]{3,191,61}{\Checkmark}} \\
    SteP \cite{sodhi2024step} & Compose handcrafted policies dynamically to solve web tasks. & \multicolumn{1}{c}{\color[RGB]{3,191,61}{\Checkmark}} &       & \multicolumn{1}{c}{\color[RGB]{3,191,61}{\Checkmark}} \\
    \rowcolor[rgb]{ .949,  .949,  .949} AgentOccam \cite{yang2024agentoccam} & Modify web LLM's action and observation space to improve performance. & \multicolumn{1}{c}{\color[RGB]{3,191,61}{\Checkmark}} &       &  \\
    WebPilot \cite{zhang2024webpilot} & Propose a multi-agent system combining MCTS for web environments. & \multicolumn{1}{c}{\color[RGB]{3,191,61}{\Checkmark}} &       & \multicolumn{1}{c}{\color[RGB]{3,191,61}{\Checkmark}} \\
    \rowcolor[rgb]{ .949,  .949,  .949} RLEM \cite{zhang2023large} & Propose a framework allowing LLM to use experience from multiple tasks. & \multicolumn{1}{c}{\color[RGB]{3,191,61}{\Checkmark}} &       & \multicolumn{1}{c}{\color[RGB]{3,191,61}{\Checkmark}} \\
    LASER \cite{ma2023laser} & Model the interactive web tasks as state-space exploration. & \multicolumn{1}{c}{\color[RGB]{3,191,61}{\Checkmark}} &       & \multicolumn{1}{c}{\color[RGB]{3,191,61}{\Checkmark}} \\
    \rowcolor[rgb]{ .949,  .949,  .949} ICAL \cite{sarch2024vlm} & Construct multimodal memory from sub-optimal demonstrations. & \multicolumn{1}{c}{\color[RGB]{3,191,61}{\Checkmark}} & \multicolumn{1}{c}{\color[RGB]{3,191,61}{\Checkmark}} & \multicolumn{1}{c}{\color[RGB]{3,191,61}{\Checkmark}} \\
    Synapse \cite{zheng2023synapse} & Simplify webpages to include full trajectories in prompts. & \multicolumn{1}{c}{\color[RGB]{3,191,61}{\Checkmark}} & \multicolumn{1}{c}{\color[RGB]{3,191,61}{\Checkmark}} &  \\
    \rowcolor[rgb]{ .949,  .949,  .949} DEPS \cite{wang2023describe} & Propose an LLM based planning framework in Minecraft. & \multicolumn{1}{c}{\color[RGB]{3,191,61}{\Checkmark}} &       & \multicolumn{1}{c}{\color[RGB]{3,191,61}{\Checkmark}} \\
    JARVIS-1 \cite{wang2023jarvis} & Propose an multimodal agent in Minecraft. & \multicolumn{1}{c}{\color[RGB]{3,191,61}{\Checkmark}} & \multicolumn{1}{c}{\color[RGB]{3,191,61}{\Checkmark}} &  \\
    \rowcolor[rgb]{ .949,  .949,  .949} VillagerAgent \cite{dongetal2024villageragent} & Introduce a multi-agent benchmark and framework in Minecraft. & \multicolumn{1}{c}{\color[RGB]{3,191,61}{\Checkmark}} &       & \multicolumn{1}{c}{\color[RGB]{3,191,61}{\Checkmark}} \\
    STEVE \cite{zhao2024see} & Propose a visionary agent in Minecraft. & \multicolumn{1}{c}{\color[RGB]{3,191,61}{\Checkmark}} & \multicolumn{1}{c}{\color[RGB]{3,191,61}{\Checkmark}} & \multicolumn{1}{c}{\color[RGB]{3,191,61}{\Checkmark}} \\
    \rowcolor[rgb]{ .949,  .949,  .949} Cradle \cite{tan2024cradle} & Propose a VLM framework interacting with software via a unified interface. & \multicolumn{1}{c}{\color[RGB]{3,191,61}{\Checkmark}} & \multicolumn{1}{c}{\color[RGB]{3,191,61}{\Checkmark}} & \multicolumn{1}{c}{\color[RGB]{3,191,61}{\Checkmark}} \\
    Voyager \cite{wang2023voyager} & Propose the first lifelong LLM agent in Minecraft. & \multicolumn{1}{c}{\color[RGB]{3,191,61}{\Checkmark}} & \multicolumn{1}{c}{\color[RGB]{3,191,61}{\Checkmark}} & \multicolumn{1}{c}{\color[RGB]{3,191,61}{\Checkmark}} \\
    \rowcolor[rgb]{ .949,  .949,  .949} GITM \cite{zhu2023ghost} & Propose framework integrating LLMs with text memory in Minecraft. & \multicolumn{1}{c}{\color[RGB]{3,191,61}{\Checkmark}} & \multicolumn{1}{c}{\color[RGB]{3,191,61}{\Checkmark}} & \multicolumn{1}{c}{\color[RGB]{3,191,61}{\Checkmark}} \\
    Huang et al.\cite{huang2022language} & Employ demonstrations to translate plans to admissible actions. &       & \multicolumn{1}{c}{\color[RGB]{3,191,61}{\Checkmark}} & \multicolumn{1}{c}{\color[RGB]{3,191,61}{\Checkmark}} \\
    \rowcolor[rgb]{ .949,  .949,  .949} MemoryBank \cite{zhong2024memorybank} & Propose a LLM memory system allowing continuous  memory updates. &       & \multicolumn{1}{c}{\color[RGB]{3,191,61}{\Checkmark}} &  \\
    Tree of Thoughts \cite{yao2024tree} & Arrange problem solving intermediate steps generated by LLM in a tree. &       &       & \multicolumn{1}{c}{\color[RGB]{3,191,61}{\Checkmark}} \\
    \rowcolor[rgb]{ .949,  .949,  .949} ReAct \cite{yao2022react} & Prompt LLM to generate interleaved verbal reasoning traces and actions. &       &       & \multicolumn{1}{c}{\color[RGB]{3,191,61}{\Checkmark}} \\
    Reflexion \cite{shinn2024reflexion} & Reinforce LLM reasoning through linguistic feedback from previous trials. &       &       & \multicolumn{1}{c}{\color[RGB]{3,191,61}{\Checkmark}} \\
    \rowcolor[rgb]{ .949,  .949,  .949} RAP \cite{hao2023reasoning} & Propose a framework using LLMs as world model and reasoning agent. &       &       & \multicolumn{1}{c}{\color[RGB]{3,191,61}{\Checkmark}} \\
    LLM-MCTS \cite{zhao2023large} & Employ LLM as both world model and policy that acts on it. &       &       & \multicolumn{1}{c}{\color[RGB]{3,191,61}{\Checkmark}} \\
    \rowcolor[rgb]{ .949,  .949,  .949} SwiftSage \cite{lin2023swiftsage} & Propose an efficient LLM reasoning framework imitating human cognition. &       & \multicolumn{1}{c}{\color[RGB]{3,191,61}{\Checkmark}} & \multicolumn{1}{c}{\color[RGB]{3,191,61}{\Checkmark}} \\
    API-Bank \cite{li2023apibank} & Introduce a tool learning dataset and a data generation method. &       &       & \multicolumn{1}{c}{\color[RGB]{3,191,61}{\Checkmark}} \\
    \rowcolor[rgb]{ .949,  .949,  .949} ADaPT \cite{prasad2023adapt} & Propose an LLM reasoning algorithm decomposing sub tasks recursively. &       &       & \color[RGB]{3,191,61}{\Checkmark} \\
    Tree-planner \cite{hu2023tree} & Employ tree structure to reduce token consumption of LLM reasoning. &       &       & \color[RGB]{3,191,61}{\Checkmark} \\
    \rowcolor[rgb]{ .949,  .949,  .949} ToolEVO \cite{chen2024learning} & Inprove the adaptability of LLM in dynamic tool environment. &       &       & \color[RGB]{3,191,61}{\Checkmark} \\
    DEER \cite{gui2024look} & \multicolumn{1}{l}{Enable LLM to invoke tools only when appropriate tools are available.} &       &       & \color[RGB]{3,191,61}{\Checkmark} \\
    \rowcolor[rgb]{ .949,  .949,  .949} Raman et al. \cite{raman2022planning} & Employ error information to enhance LLM planning ability. &       &       & \color[RGB]{3,191,61}{\Checkmark} \\
    \bottomrule
    \end{tabular}%
  }
  \label{tab:action_contribution}%
\end{table*}%

In Table \ref{tab:action_contribution}, we provide an overview of research aimed at improving the quality of LLM agent actions. This subsection focuses on grounding actions, which involve perceiving the environment through textual descriptions and generating text to determine the most suitable subsequent actions \cite{sumers2024cognitive}. We begin by outlining the key challenges lifelong LLM agents face in executing these grounding actions, then explore corresponding solutions under different environmental contexts.

\subsection{Challenges of Grounding Actions}
\label{sec:grounding_actions:challenges_of_grounding_actions}

As the brain of agent, LLM is responsible for taking textual observations from the environment as input and generating textual actions as output. Similar to \cite{yang2024agentoccam}, we define \emph{input grounding actions} as the process of perceiving and understanding textual environment description, and \emph{output grounding actions} as generating text that can be parsed into actions by environment. The generation process of a grounding action at time $t$ can be formulated as:
\begin{equation}
ra_t, a_t = \mathcal{G}(\mathcal{E}, g, t) =
\begin{cases} 
\mathcal{G}(\mathcal{E}, g, o_0), & \text{if } t = 0 \\ 
\mathcal{G}(\mathcal{E}, g, \xi(t)), & \text{otherwise}.
\end{cases}
\end{equation}
where:
\begin{itemize}
    \item ${ra}_t$ is the rationale corresponding to the generated action.
    \item $a_t$ is the action generated at time $t$, which can be viewed as the result of output grounding action.
    \item $\mathcal{G}$ is the generative LLM backbone.
    \item $g$ is the goal defined in definition \ref{def:agent}.
    \item $\mathcal{E}$ is the environment defined in definition \ref{def:task}. Here we omit the task index for simplification. The environment is presented in textual form, serving as the primary focus for input grounding actions to interpret and process.
    \item $o_0$ is the initial observation.
    \item $\xi(t)$ is the trajectory defined in definition \ref{def:trial} up to time $t$.
\end{itemize}

\subsubsection{Input Grounding Actions}

For \emph{input grounding actions}, there are notable differences between the textual formats encountered in the pretraining corpus and those used for environment descriptions. While the pretraining corpus primarily consists of well-structured paragraphs, environment descriptions often take the form of brief sentences, short phrases, or structured text formats such as JSON strings or HTML tags. As a result, LLM must adapt from the familiar input formats of its pretraining data to the diverse and specialized formats found in the agent's environment. In rapidly changing environments, the agent needs to adapt to the updating descriptions continuously to get better understanding of the environment.

\subsubsection{Output Grounding Actions}

For \emph{output grounding actions}, there are significant differences in the types of content LLM is required to generate. During pretraining, LLM is primarily trained for simple text completion, but in an agent's environment, it must generate text that adheres to specific patterns, representing actions or environment-specific elements. LLM must learn to perform complex actions by producing outputs tailored to the requirements of the environment, rather than merely describing actions or intents in free-form natural language. Furthermore, in complex environments, the requirements for output grounding actions may change based on the agent's previous actions, necessitating continuous adaptation to align with the evolving demands of the environment.

\subsubsection{Summary}

The differences between input and output grounding actions highlight the critical importance of lifelong learning in LLM agents. LLM needs to adapt its input and output grounding actions, originally designed for the pretraining phase, to align with the specific requirements of different environments. Furthermore, it also needs to continuously adjust these actions to effectively respond to ever-changing environment.

\subsection{Solutions for Different Environments}
\label{sec:grounding_actions:solutions_for_different_environments}

LLM agents with lifelong learning capabilities can not only adapt their grounding actions from the pretraining phase to suit specific environments but also continuously evolve through interactions with their surroundings. However, the variability across environments presents unique challenges, prompting the development of diverse solutions in existing research. To offer a clear and comprehensive overview of these solutions, we classify commonly used environments into three categories, ranging from simple to complex: \textbf{tools}, \textbf{web}, and \textbf{game}. 

\subsubsection{Tool Environment}

Tool refers to an external functionality or resource that agent can interact with to enhance its capabilities. Calculator, calendar, search engine \cite{Timo2023toolformer}, and Application Programming Interface (API) \cite{qin2024toolllm} can all be viewed as tools. Tools can enhance LLM agent expertise in specific domains and provide better interpretability to agent decision process. In tool environment, an LLM agent must first understand the tools' functionalities and then invoke the appropriate tools in the correct order based on user intents.

The \emph{input grounding actions} of an LLM agent in a tool environment primarily involve understanding tool documentation. Complex tools are often encapsulated as APIs and described using JSON strings \cite{qin2024toolllm, guo2024stabletoolbench, patil2024gorilla, tang2023toolalpaca, xu2024on, li2023apibank}. To fully understand these tools, LLM must adapt to the unique formats of tool documentation.

One approach to address this challenge is simplifying tool documentation directly within the prompt, enabling LLM to focus on essential information. For instance, EASYTOOL \cite{yuan2024easytool} uses ChatGPT to transform verbose and redundant tool documentation into concise instructions for easier comprehension. Similarly, GEAR \cite{lu2024gear} employs a carefully designed framework composed of smaller language models to select tools for LLM, narrowing its focus to specific tool documentation. Another common strategy involves leveraging tool calling trajectories for fine-tuning or as in-context learning demonstrations, helping LLM better understand tools \cite{qin2024toolllm, xu2024on, wang2024llm}. 

Furthermore, ART \cite{paranjape2023art} proposes manually revising tool calling trajectories generated by LLM and storing them in a database for future use. Confucius \cite{gao2024confucius} also utilizes tool calling trajectories to fine-tune LLM, iteratively updating the training dataset based on LLM's performance with various tools. Both approaches adopt a lifelong learning perspective to refine LLM's output grounding actions. By utilizing previously generated tool calling trajectories, they enhance the agent's performance in subsequent tool interactions, enabling continuous improvement over time.

LLM also need to adapt its \emph{output grounding actions} to tool environments so that it can output text in particular format to use tool and pass parameters. Almost all works force LLM to generate text in particular format by fine-tuning \cite{Timo2023toolformer} or few-shot learning \cite{yuan2024easytool}. ToolkenGPT \cite{hao2023toolkengpt} stands out among these works by introducing a special token in the output embeddings of LLMs, enabling the models to invoke tools. By fine-tuning only the output embeddings corresponding to this token, it can call tools while maximizing the preservation of its language modeling capabilities.

The works mentioned above focus on improving LLM's tool-calling abilities in static tool environments. Recent studies, however, explore how to enhance LLM's tool-calling capabilities in lifelong tool environments. A lifelong tool environment is the tool environment where LLM must continually adapt its grounding actions to changing tools. For example, STE \cite{wang2024llm} finds out that a simple replay strategy can effectively mitigate catastrophic forgetting during continual fine-tuning. Similarly, Chen et al. \cite{chen2024learning} focus on fine-tuning LLM to overcome the difficulties brought by outdated tool documentations. In contrast to these approaches, LATM \cite{cai2024large} emphasizes constructing a lifelong tool environment itself to better fulfill user needs. It employs GPT-4 to develop tools for handling increasing user requests and GPT-3.5 to effectively utilize these tools.

\subsubsection{Web Environment}

In web environments, LLM-based agents must engage with webpages based on user intent. Unlike humans, LLM perceive webpages primarily through the HTML DOM tree \cite{yao2022webshop} or the accessibility tree \cite{zhou2024webarena}. These formats are not only lengthy but also fail to intuitively display webpage content, posing significant challenges to the \emph{input grounding actions} of LLM. To address this, works such as AgentOccam \cite{yang2024agentoccam} and Synapse \cite{zheng2023synapse} focus on simplifying webpages before including them in prompts, improving the accuracy of input grounding actions. Additionally, studies like WebPilot \cite{zhang2024webpilot}, RLEM \cite{zhang2023large}, ICAL \cite{sarch2024vlm}, AWM \cite{wang2024agent}  and Synapse \cite{zheng2023synapse} incorporate previous trajectories or episodic experiences into prompts, enabling LLM to enhance their input grounding actions and continuously improve their understanding of webpage content during browsing.

Compared to the tool environment, it is more challenging for agent to generate high quality \emph{output grounding actions} in web environment. The agent may be confused by the web interaction actions that are irrelevant to current webpages or hard to use. SteP \cite{sodhi2024step} and LASER \cite{ma2023laser} propose removing irrelevant actions description from the prompt. Likewise, AgentOccam \cite{yang2024agentoccam} proposes only including actions that can be grasped by LLM in the prompt. 

\subsubsection{Game Environment}

Game environment is the most complex environment among these three kinds of environments. Given that LLM embodied agents usually operate in virtual environments, we classify their working environment as a game environment. 
Based on the APIs provided by different game environments \cite{puig2018virtualhome, mineflayer}, the specific requirements for LLM's input grounding actions and output grounding actions vary across environments. For \emph{input grounding actions}, DEPS \cite{wang2023describe}, JARVIS-1 \cite{wang2023jarvis}, and VillagerAgent \cite{dongetal2024villageragent} use specialized prompts to help LLM gain a deeper and clearer understanding of the environment. Alternatively, another line of research \cite{zhao2024see, tan2024cradle} treats the environment as an image and employs Virtual Language Models to directly perceive the complex environment. These approaches help reduce the information loss that can occur when describing complex environments using natural language. For \emph{output grounding actions}, most works enable LLM to interact with the environment by generating executable programs. These executable programs can be used to directly control the agent's behavior \cite{wang2023voyager, zhao2024see} or indirectly control it by being mapped to keyboard or mouse operations \cite{zhu2023ghost, tan2024cradle}.

Due to the complexity of the game environment and its rapidly changing nature, many works enhance the long-term consistency of agent behavior and the overall capabilities of the agent from the perspective of lifelong learning. Voyager \cite{wang2023voyager} is the first LLM-powered embodied lifelong learning agent in Minecraft. It can generate skills to solve current task and store them in skill library to solve subsequent tasks. What's more, JARVIS-1 \cite{wang2023jarvis}, STEVE \cite{zhao2024see} and Cradle \cite{tan2024cradle} use additional memory modules to store previous experience, prompting better plans generation in subsequent interactions.

\subsection{Summary}

In this subsection, we categorize environments into three categories and discuss how LLM agents can adapt to each of them. It is important to note that while some methods are specific to certain environments \cite{achiam2023gpt}, most of these methods can be transferred to different environments. For example, the idea of making executable programs to continuously improve agent's ability \cite{cai2024large} can also be applied in web environment. In such a case, the agent can generate subsequent plans or actions after executing an executable program, rather than after each simple mouse/keyboard action (e.g., clicking a button), thereby reducing the length of the action history and observation history, which improves the long-term consistency of the agent's behavior. Different approaches can also be combined to enhance the lifelong LLM agent's performance.

\section{Action Design: Retrieval Actions}
\label{sec:retrieval_actions}

LLM agent needs \emph{external information} to generate high quality grounding actions and reasoning actions. For grounding actions, LLM's generation can be parsed into actions only if it matches the patterns specified by the environment. Simply fine-tuning LLM to force it generate outputs that conform to environmental constraints is impractical, as this approach not only incurs significant resource costs but also fails to adapt to the continuously changing action space. Moreover, directly including all possible action descriptions in the prompt is unfeasible due to the unacceptable context length and the \emph{lost in the middle} problem \cite{liu2024lost}. For reasoning actions, comprehensive external knowledge from semantic memory and accurate historical trajectory from episodic memory are key factors in making correct decisions. However, text at the scale of the internet cannot be fully included in the context length of an LLM. The lengths of action history and observation history will also gradually increase with the agent's activities, eventually exceeding the context length of LLM. 

These challenges highlight the critical importance of retrieval actions for lifelong LLM agents. With the retrieval actions, LLM agent can handle the continuously growing actions history and observation history, which helps maintain the long-term consistency of the LLM's behavior. It can also gain real-time knowledge by retrieving from continuously refreshed knowledge sources, leading to better performance in environment changing continuously \cite{gao2024retrieval}. From the lifelong learning perspective, the agent can also retrieval the knowledge gained in previous task to improve its performance on current task. Moreover, Retrieval action can also improve the interpretability of the agent and its performance on knowledge-intensive task \cite{ovadia2024fine}. We classify the work that allow the agent to retrieval from memory into two parts based on the retrieval source, which are \emph{semantic memory} and \emph{episodic memory}. The retrieval and utilization process of semantic memory can be formulated as bellow. LLM agent can perform the retrieval action only at the beginning of a trial or before generating each action. Here, we assume that the LLM agent performs the retrieval action only at the beginning of a trial, which is a common practice in existing research \cite{zhu2023ghost, raman2022planning}.

\begin{align}
m_{\mathcal{T}^{(i)}} &= \mathrm{retrieve}(\mathcal{T}^{(i)},\mathcal{M}^{(i)}) \\
ra_t^{(i)}, a_t^{(i)} &= \mathcal{G}(\mathcal{E}^{(i)}, m_{\mathcal{T}^{(i)}},g^{(i)},t)
\end{align}
where:
\begin{itemize}
    \item $\mathrm{retrieve}$ denotes the retrieval process.
    \item $\mathcal{M}$ is the pre-defined, static database.
    \item $m_{\mathcal{T}^{(i)}}$ is the retrieval memory item corresponding to task $\mathcal{T}^{(i)}$, which can be background knowledge or demonstrations.
\end{itemize}

The retrieval and utilization process of episodic memory is similar to that of semantic memory, while the updating process for episodic memory can be outlined as follows.

\begin{align}
\mu^{(i)} &= \mathrm{process}(\mathcal{E}^{(i)},\xi_{T_i}^{(i)}) \\
\mathcal{M}^{(i+1)} &= \mathrm{update}(\mathcal{M}^{(i)}, \mu^{(i)})
\end{align}
where:
\begin{itemize}
    \item $\mathcal{M}^{(i)}$ is the dynamic database up to task $\mathcal{T}_i$.
    \item $\mathrm{process}$ denotes the abstraction and transformation of the agent's trajectory, which can be performed by either an LLM \cite{zheng2023synapse} or a human \cite{sarch2024vlm}.
    \item $\mu^{(i)}$ represents the processed memory with diverse forms. It could be a corrected or simplified trajectory or a textual summary of the task.
\end{itemize}

In Table \ref{tab:action_retrieval}, we summarize the classification results.

\renewcommand{\multirowsetup}{\centering}
\begin{table*}[!t]
  \centering
  \caption{Summary of research that focus on enhancing the retrieval actions of LLM agent. This table categorizes research based on the source of retrieval and their primary focus. For retrieval from semantic memory, research primarily aims to provide additional background knowledge or demonstrations as supplementary environmental information. For retrieval from episodic memory, the focus is on improving the agent's ability to leverage past experiences and enhancing its long-term consistency.}
        \begin{tabular}{m{2.8cm}m{5cm}m{8cm}}
    \toprule
    Retrieval Source & Focus & Research \\
    \midrule
    \multirow{2}{*}{Semantic Memory} 
       & \cellcolor[rgb]{ .949,  .949,  .949}Background Knowledge 
       & \cellcolor[rgb]{ .949,  .949,  .949}GITM \cite{zhu2023ghost}, SwiftSage \cite{lin2023swiftsage}, ToolLLM \cite{qin2024toolllm}, Huang et al.\cite{huang2022language}, LLM-MCTS \cite{zhao2023large}, AMOR \cite{guan2024amor}  \\
       & Demonstrations 
       & Re-Prompting \cite{raman2022planning}, STE \cite{wang2024llm} \\
    \midrule
    \multirow{2}{*}{Episodic Memory} 
       & \cellcolor[rgb]{ .949,  .949,  .949}Ability to leverage past experiences 
       & \cellcolor[rgb]{ .949,  .949,  .949}ICAL \cite{sarch2024vlm}, GITM \cite{zhu2023ghost}, Voyager \cite{wang2023voyager}, LATM \cite{cai2024large}, JARVIS-1 \cite{wang2023jarvis}, Cradle \cite{tan2024cradle}, ART \cite{paranjape2023art}, Xu et al. \cite{xu2024on}, Synapse \cite{zheng2023synapse} \\
       & Long-Term Consistency 
       & MemoryBank \cite{zhong2024memorybank}, Cradle \cite{tan2024cradle} \\
    \bottomrule
    \end{tabular}
  \label{tab:action_retrieval}%
\end{table*}

\subsection{Retrieval from Semantic Memory}
\label{sec:retrieval_actions:retrieval_from_semantic_memory}

Pretrained LLM is often insufficient as an agent's brain due to two key limitations: the lack of background knowledge and the lack of demonstrations. These two limitations can be addressed through retrieving from semantic memory, which is an external memory mechanism storing world knowledge. See Section \ref{sec:semantic_memory} for further explanation.

\subsubsection{Lack of Background Knowledge}

The \emph{lack of background knowledge} is typically manifested in LLM's inability to select the correct action from all possible actions or to generate actions that can be understood by the environment. To address this issue, GITM \cite{zhu2023ghost} retrieves relevant text from Minecraft Wiki to provide LLM with the Minecraft world knowledge, enabling LLM to execute actions in a correct order. Both SwiftSage \cite{lin2023swiftsage} and ToolLLM \cite{qin2024toolllm} use SentenceBERT \cite{reimers2019sentence} to retrieve possible actions from a database, helping LLM select the appropriate actions by narrowing the action space. When the parameters of the action are finite, SentenceBERT can further be used to transform action parameters generated by LLM, which cannot be understood by the environment, into valid parameters \cite{huang2022language, zhao2023large}.

\subsubsection{Lack of Demonstrations}

The \emph{lack of demonstrations} can diminish the quality of the agent's grounding actions and planning actions. Demonstrations have been proved to play an important role in LLM performance \cite{brown2020language}. However, including irrelevant or outdated demonstrations in the prompt may greatly compromise the LLM agent's performance \cite{chen2024learning}. To verify relevant issues, Re-Prompting \cite{raman2022planning} and STE \cite{wang2024llm} use SentenceBERT to select most similar demonstrations from a demonstrations set.

\subsection{Retrieval from Episodic Memory}
\label{sec:retrieval_actions:retrieval_from_episodic_memory}

While retrieving from semantic memory can improve agent's ability by providing additional background knowledge and demonstrations, it can neither address the lack of ability to leverage past experiences, nor the lack of long-term consistency in LLM. However, these two limitations can be addressed through retrieving from episodic memory. Different from semantic memory, episodic memory mainly stores past experience. See Section \ref{sec:episodic_memory} for further explanation.

\subsubsection{Lack of Ability to Leverage Past Experiences}

Overcoming the \emph{lack of ability to leverage past experiences} is a key characteristic of a lifelong LLM agent. Leveraging past experience, LLM agent can gradually improve itself during interactions with the environment. Current research utilizing past experiences can be broadly classified into two categories, both aimed at enhancing the agent's capabilities from a lifelong learning perspective. The first category involves storing the agent's trajectory in a database after it completes a task successfully \cite{sarch2024vlm, zhu2023ghost}. When a new task arises, the agent retrieves similar task trajectories from the database and incorporates them into the prompt. This category of research improve the quality of agent's reasoning actions. The second category focuses on representing the agent's task-solving steps as executable programs \cite{wang2023voyager, cai2024large, wang2023jarvis, tan2024cradle}. These programs are primarily composed of the acceptable actions defined by the environment and can be seen as high-level actions generated by the agent. When facing a new task, the agent can either directly reuse these programs or combine them to address the new challenge. This category research expands the action space of the environment and improves the grounding ability of agents. 

\subsubsection{Lack of Long-Term Consistency}

The \emph{lack of long-term consistency} primarily arises from the fact that the context length of LLMs is finite, preventing them from incorporating the entire observation and action history into the prompt. Improving the long-term consistency can make the agent more similar to humans. MemoryBank \cite{zhong2024memorybank} not only retrieve the summary of past conversations to keep current consistent with the chatting history, improving the performance of LLM in lifelong interaction scenarios.

\subsection{Summary}

In this subsection, we discuss the crucial role of retrieval actions for lifelong LLM agents. Most existing LLM agent research only focuses on retrieval from either semantic memory or episodic memory. However, it is worth noting that retrieving from both semantic and episodic memory can further improve the agent's performance. This approach mirrors how humans use long-term and short-term memory. Additionally, lifelong LLM agent can also leverage well-established techniques from the RAG field, such as iterative retrieval \cite{kim2023tree, feng2024retrieval, ye2023effective} or constructing higher-quality retrieval sources continuously \cite{cheng2024lift}, to improve the quality of the retrieved content.

\section{Action Design: Reasoning Actions}
\label{sec:reasoning_actions}

The final category of actions is reasoning actions. An LLM agent can gain deeper insights from the content loaded into its working memory. After pretraining, the LLM is capable of performing basic reasoning using techniques such as \emph{chain-of-thought} \cite{wei2022chain}. However, its reasoning ability is insufficient to handle more complex reasoning tasks within the agent's environment. There are two primary reasons for this limitation. First, the environment in which the LLM operates is relatively complex, making it difficult for the LLM to perform high-quality reasoning based solely on the knowledge acquired during pretraining. Second, the reasoning capabilities of the LLM itself are inherently limited. For instance, a pretrained LLM is unlikely to recognize errors in prior reasoning steps without techniques like reflection, resulting in incorrect conclusions. 

To address these challenges, numerous research on LLM agents have focused on improving the quality of reasoning actions through carefully designed prompts or novel frameworks. Some of this research addresses the issue from the perspective of lifelong learning, allowing the LLM to refine its current reasoning behavior by building upon prior reasoning outcomes or trajectories. These approaches enable the agent to improve its reasoning ability gradually across different trials within a single episode, or over longer periods across multiple episodes. Based on this, we classify the agent's reasoning actions into \emph{intra-episodic reasoning actions} and \emph{inter-episodic reasoning actions}. We summary the classification result in Table \ref{tab:action_reasoning}.

\renewcommand{\multirowsetup}{\centering}
\begin{table*}[!t]
  \centering
  \caption{Summary of research that focuses on enhancing the reasoning actions of LLM agent. This table categorizes research based on time span. For research that improve agent reasoning ability within a single episodic, we further divide them into two categories: those that stimulate the LLM's intrinsic reasoning ability within the same trial and those that progressively enhance reasoning ability across multiple trials.
  }
    
    \begin{tabular}{m{1.8cm}m{1.8cm}m{12.2cm}}
    \toprule
    \multicolumn{2}{l}{Focus} & Research \\
    \midrule
    \multirow{2}[2]{*}{Intra-Episodic} & \cellcolor[rgb]{ .949,  .949,  .949}Single Trial & \cellcolor[rgb]{ .949,  .949,  .949}ReAct \cite{yao2022react}, $\alpha$-UMi \cite{shen2024small}, API-Bank \cite{li2023apibank}, LASER \cite{ma2023laser}, SteP \cite{sodhi2024step} \\
          & Across Trials & Reflexion \cite{shinn2024reflexion}, Tree of Thoughts \cite{yao2024tree}, Tree-planner \cite{hu2023tree}, RAP \cite{hao2023reasoning}, LLM-MCTS \cite{zhao2023large}, WebPilot \cite{zhang2024webpilot}, ToolEVO \cite{chen2024learning} \\
    \midrule
    \rowcolor[rgb]{ .949,  .949,  .949} \multicolumn{2}{l}{Inter-Episodic} & ICAL \cite{sarch2024vlm}, GITM \cite{zhu2023ghost}, Raman et al. \cite{raman2022planning}, Voyager \cite{wang2023voyager}, Cradle \cite{tan2024cradle}, MemoryBank \cite{zhong2024memorybank}, STEVE \cite{zhao2024see} \\
    \bottomrule
    \end{tabular}%
  \label{tab:action_reasoning}%
\end{table*}%

\subsection{Intra-Episodic Reasoning Actions}
\label{sec:reasoning_actions:intra-episodic_reasoning_actions}

\emph{Intra-episodic reasoning actions} refer to reasoning actions that leverage the experiences within the same episode. Based on whether these research stimulate the LLM's intrinsic reasoning ability within the same trial or progressively enhance its reasoning ability across different trials, we further categorize the articles into two groups.

\subsubsection{Single Trial}

On the one hand, nearly all research encourages the LLM agent to perform reasoning in the ReAct \cite{yao2022react} style \emph{within a single trial}. ReAct allows the LLM to continuously refine its reasoning process based on real-time feedback from the environment, ultimately leading to more accurate plans or decisions. On the other hand, many research break down reasoning into several key steps, assigning different LLMs to handle each step. For instance, $\alpha$-UMi \cite{shen2024small} fine-tunes two LLMs, with one responsible for planning and the other for summarization. Similarly, API-Bank \cite{li2023apibank} employs five LLMs to handle each state of the reasoning process, ultimately generating high-quality tool-learning training data. These research improve the reasoning action quality of an agent or an agent system. 
    
Additionally, many research enhance the reasoning ability of LLM agents in complex environments by introducing environment-specific strategies. For example, LASER \cite{ma2023laser} models the process as state-space exploration and reduces the difficulty of reasoning by only allowing the LLM to transition between adjacent states. Similarly, SteP \cite{sodhi2024step} enhances the LLM's reasoning ability in complex environments by dynamically composing handcrafted policies, with each policy serving as a method for extracting information from a specified complex environment.

\subsubsection{Across Trials}

Many research further improve the agent reasoning ability \emph{across different trials} by mimicking human reasoning process. These research enable agent to make use of experience not only from current trial, but also from previous trials. Agents are also instructed to perform reasoning in ReAct style. Reflexion \cite{shinn2024reflexion} is one of the most representative research. It enables LLM to engage in self-reflection by reviewing past failed trials. This reflective process helps the agent refine its reasoning, ultimately leading to more accurate results in future trials. The action generation process when Reflexion is applied can be formulated as bellow. LLM agent will reflect on its past trajectories at the start of each new trial, repeating this process as needed until it successfully achieves the given goals.

\begin{align}
\mathrm{reflection} &= \mathrm{reflect}(\widetilde{\xi}) \\
ra_t, a_t &= \mathcal{G}(\mathcal{E}, g, \mathrm{reflection}, t)
\end{align}
\noindent where:
\begin{itemize}
    \item $\widetilde{\xi}$ is the previous full-length trajectories of the past trials.
    \item $\mathrm{reflect}$ denotes the process of extracting valuable insights from past trials.
    \item $\mathrm{reflection}$ is the verbal experience feedback for past trials.
\end{itemize}

Another line of research make use of the incomplete trails. These research typically use tree structures to manage each incomplete trial. During the reasoning process, each path from the root node to a leaf node can be viewed as an incomplete trial. With the help of the tree structure, LLM agent can perform higher-level operations such as lookahead and backtracking, continuously improving the quality of reasoning actions across multiple trials \cite{yao2024tree, hu2023tree}. Also thanks to the tree structure, many classical tree search algorithms, such as Breadth-first search, Depth-first search \cite{yao2024tree}, and Best-first search \cite{koh2024tree}, can be seamlessly integrated into the reasoning process. Furthermore, many research use Monte Carlo Tree Search (MCTS) to enhance the reasoning ability of agents operating in complex environments \cite{hao2023reasoning, zhao2023large, zhang2024webpilot, chen2024learning}. MCTS enables the agent to more effectively explore and exploit the complex environment, ultimately leading to the formulation of a correct plan through reasoning actions.

\subsection{Inter-Episodic Reasoning Actions}
\label{sec:reasoning_actions:inter-episodic_reasoning_actions}

\emph{Inter-episodic reasoning actions} refer to reasoning actions that use the experiences from different episodic. These experiences accumulate gradually as the LLM agent interacts with the lifelong environment. Experiences can take many forms, such as successful reasoning trajectories \cite{sarch2024vlm, zhu2023ghost, raman2022planning}, executable code \cite{wang2023voyager}, or textual summary \cite{tan2024cradle, zhong2024memorybank}. These experiences are typically stored in an additional database. When encountering a new task, the LLM agent first retrieves the most relevant experiences and add them to the working memory, then uses these experiences to inform its reasoning process. Specially, AMOR \cite{guan2024amor} leverages feedback from previous tasks to fine-tune the model, improving its reasoning actions quality under specific environments.

Additionally, some research focus on how to enable LLM agent to explore complex environment employs curriculum learning to arrange different tasks. Additionally, some research focus enabling LLM agent to explore complex environment employ curriculum learning to better make use of the past experiences. These research \cite{zhao2024see, wang2023voyager} arrange the task from easy to difficult, providing a steady stream of new tasks or challenges. Through curriculum learning, LLM agent can gradually master reasoning action techniques throughout the lifelong exploration process.

\subsection{Summary}

In this subsection, we discuss the primary method of improve the reasoning ability of LLM agent in lifelong environment. Current research mostly employ experience from previous trails or episodic to improve the reasoning ability of agent continuously. However, existing research, particularly that which uses tree structures \cite{hu2023tree, kim2023tree, koh2024tree, yao2024tree}, often overlooks the dynamic nature of complex environments and assumes that any state in the online reasoning process is retraceable. We would like to point out that this is unrealistic in practice. For instance, in a web environment, payments can be irreversible. Future research should explore how to facilitate the lifelong accumulation of reasoning techniques in more realistic settings, such as those characterized by irreversible actions.

\section{Evaluation of Lifelong LLM Agents}
\label{sec:evaluation_of_lifelong_llm_agents}

Evaluation metrics, datasets, and benchmarks are critical components of research on lifelong LLM agents. Compared to traditional LLM agents, lifelong LLM agents can retain useful historical experiences through interactions with the environment. These experiences enable LLM agents to achieve better performance on both past and future tasks. As a result, many evaluation metrics are designed to demonstrate the superiority of lifelong learning methods by measuring the LLM's performance on tasks. However, to effectively compare different lifelong learning methods, having robust evaluation metrics alone is insufficient. Reliable datasets and highly realistic and reproducible benchmarks also indispensable. In this section, we will briefly introduce the widely used evaluation metrics in one subsection, followed by a discussion of the datasets and benchmarks in another.

\subsection{Evaluation Metrics}
\label{sec:evaluation_of_lifelong_llm_agents:evaluation_metrics}

The evaluation metrics of lifelong LLM agents primarily assess their lifelong learning ability from three perspectives: 
(1) the overall accuracy of all tasks, 
(2) the stability of history tasks 
(3) the plasticity of future tasks. 
The performance of LLM agent can be measured by different metrics from different aspects, such as accuracy, pass rate, and win rate \cite{qin2024toolllm}. Here, we assume that the agent's performance can be converted into a scalar obtained by aggregating various metrics. Following \ref{sec:building_lifelong_learning_llm_agents:formal_definition_of_lifelong_learning_for_llm_based_agents}, we denote the performance of the LLM agent on task $i$ as $J_{i,t}$ after completing $t$ tasks \cite{zheng2024towards}.

There are two evaluation metrics that measure the lifelong learning ability of LLM agent from the first perspective, which are \emph{average performance} ($\mathrm{AP}$) and \emph{average incremental performance} ($\mathrm{AIP}$). The difference between $\mathrm{AP}$ and $\mathrm{AIP}$ is that, $\mathrm{AIP}$ captures the historical variation when experiencing each task. 

\begin{align}
\mathrm{AP}_t&=\frac{1}{t}\sum_{i=1}^{t}J_{t,i}, \\
\mathrm{AIP}&=\frac{1}{T}\sum_{t=1}^{T}\mathrm{AP}_t
\end{align}

Other two metrics, \emph{forgetting measure} ($\mathrm{FGT}$) and \emph{backward transfer} ($\mathrm{BWT}$) measure lifelong learning ability of LLM agent from the perspective of stability. $\mathrm{FGT}$ evaluates the average accuracy drop of each old task, representing that whether useful experience are maintained successfully. $\mathrm{BWT}$ evaluates the average accuracy improvement of each old tasks, representing whether the experience gained after the LLM agent encountered a given task benefits that task.

\begin{align}
\mathrm{FGT}_t&=\frac{1}{t-1}\sum_{i=1}^{t-1}[\max_{j\in\{i,i+1,\cdots,t\}}(\{J_{j,i}\}_j)-J_{t,i}], \\
\mathrm{BWT}_t&=\frac{1}{t-1}\sum_{i=1}^{t-1}(J_{t,i}-J_{i,i}).
\end{align}

There is also a metric called \emph{forward transfer} ($\mathrm{FWT}$) that calculate lifelong learning ability of LLM agent from the third perspective. It represents whether the gained experience is useful to future tasks.

\begin{align}
\mathrm{FWT}_t=\frac{1}{t-1}\sum_{i=2}^t(J_{i,i}-\tilde{J_i}),
\end{align}
where $\tilde{J_i}$ is the performance of a LLM agent that has no experience.

\subsection{Datasets and Benchmarks}
\label{sec:evaluation_of_lifelong_llm_agents:datasets_benchmarks}

In this subsection, we introduce the benchmarks and datasets commonly used to evaluate the lifelong learning capabilities of LLM agents. We categorize these benchmarks and datasets into two groups: the first focuses on assessing the ability of LLM agents to perform continual learning in \emph{isolated, simple scenarios}, while the second evaluates their performance in more \emph{integrated and complex scenarios}.

The first group of datasets and benchmarks is typically used to evaluate the lifelong learning ability of LLMs in simple scenarios. During the pretraining process, LLMs not only acquire knowledge but also develop specific skills, such as instruction following, machine translation, question answering, and code generation, through targeted corpora. Since the benchmarks and datasets in simple scenarios typically focus only on changes in a specific skill during the lifelong learning process, LLM agents only need to achieve lifelong learning in that skill to perform well on these benchmarks and datasets. 

Many traditional continual learning benchmarks and datasets belong to the first group. For example, CITB \cite{zhang2023citb}, GLUE \cite{wang2018glue}, SuperGLUE \cite{wang2019superglue}, NaturalInstruction \cite{mishra2021cross} and SuperNI \cite{wang2022super} mainly concern the instruction following ability of LLM agent. WMT \cite{statmt}, TED Talks \cite{duh18multitarget} and CLLE \cite{zhang2022clle} concern the machine translation aspect. There also exists benchmarks and datasets that concern the change of internal knowledge of LLM during the lifelong learning process, such as zsRE \cite{de2021editing}, FEVER \cite{thorne2018fever}, CounterFact \cite{meng2022locating}, Concept-1K \cite{zheng2024concept} and Biography \cite{allen2023physics}. 

The second groups of datasets and benchmarks used to evaluation the lifelong learning ability of LLMs in complex scenarios. These benchmarks and datasets closely reflect real-world scenarios, assessing the performance of LLM agents based on their ability to fulfill practical user requests. To succeed, LLM agents must achieve lifelong learning across multiple skills, effectively utilize knowledge from previous tasks, and seamlessly integrate capabilities such as reasoning, planning, multi-turn dialogue, knowledge retrieval, and tool usage.

Many benchmarks and datasets that evaluate the performance of LLM agent under special environment belong to the second group. For example, ToolBench \cite{qin2024toolllm, xu2024on}, StableToolBench \cite{guo2024stabletoolbench}, APIBench \cite{patil2024gorilla}, ToolAlpaca \cite{tang2023toolalpaca}, API-Bank \cite{li2023apibank} evaluate LLM agent's performance in tool environment. WebArena \cite{zhou2024webarena}, WebShop \cite{yao2022webshop}, WorkArena \cite{drouin2024workarena}, VisualWebArena \cite{koh2024visualwebarena}, VideoWebArena \cite{jang2024videowebarena}, AgentBench \cite{liu2023agentbench}, Visualagentbench \cite{liu2024visualagentbench} evaluate its performance under more challenge environment, such as web environment and game environment.

\section{Application of Lifelong LLM Agents}
\label{sec:application_of_lifelong_llm_agents}

\begin{figure*}[!t]
\centering
\tikzset{
        my node/.style={
            draw,
            align=center,
            thin,
            text width=2.5cm, 
            rounded corners=3,
        },
        my leaf/.style={
            draw,
            align=left,
            thin,
            text width=4.5cm, 
            % text height=1cm, 
            % minimum height=0.5cm,
            rounded corners=3,
        }
}
\forestset{
  every leaf node/.style={
    if n children=0{#1}{}
  },
  every tree node/.style={
    if n children=0{minimum width=1em}{#1}
  },
}
\begin{forest}
    for tree={%
        % my node,
        every leaf node={my leaf, font=},
        every tree node={my node, font=\small, l sep-=4.5pt, l-=1.pt},
        anchor=west,
        inner sep=2pt,
        % l = 10pt,
        l sep=10pt, % control leaf to parent nodes gaps (horizontal)
        s sep=5pt, % control node gaps (vertical)
        fit=tight,
        grow'=east,
        edge={ultra thin},
        parent anchor=east,
        child anchor=west,
        if n children=0{tier=last}{},
        edge path={
            \noexpand\path [draw, \forestoption{edge}] (!u.parent anchor) -- +(5pt,0) |- (.child anchor)\forestoption{edge label};
        },
        if={isodd(n_children())}{
            for children={
                if={equal(n,(n_children("!u")+1)/2)}{calign with current}{}
            }
        }{}
    }
    [\cref{sec:application_of_lifelong_llm_agents} Application of Lifelong LLM Agents, draw=gray, color=gray!100, fill=gray!15, very thick, text=black, text width=3.5cm 
        [\cref{sec:daily_application} Daily Application, color=brightlavender!100, fill=brightlavender!15, very thick, text=black, text width=3.5cm
            [Work Scenarios, color=brightlavender!100, fill=brightlavender!15, very thick, text=black
                [{\textbf{Web Applications:} WebAgent\cite{gur2023real}, Mind2Web\cite{deng2024mind2web}, WebArena\cite{zhou2024webarena}, Webgum\cite{furuta2023multimodal}, WebShop\cite{yao2022webshop}, WebGPT\cite{nakano2021webgpt}, Synapse\cite{zheng2023synapse}, Agentoccam\cite{yang2024agentoccam}, etc.}, color=brightlavender!100, fill=brightlavender!15, very thick, text=black, tier=Task, text width=6.9cm
                ]
                [{\textbf{Knowledge Management:} SWIFTSAGE\cite{lin2023swiftsage}, ToolLLM\cite{qin2024toolllm}, Sentence-BERT\cite{reimers2019sentence}, Memorybank\cite{zhong2024memorybank}, GITM\cite{zhu2023ghost}, Kim et al.\cite{kim2023tree}, Feng et al.\cite{feng2024retrieval}, Ye et al.\cite{ye2023effective}, etc.}, color=brightlavender!100, fill=brightlavender!15, very thick, text=black, tier=Task, text width=6.9cm
                ]
            ]
            [Life Scenarios, color=brightlavender!100, fill=brightlavender!15, very thick, text=black
                [{\textbf{Chat Agents:} ChatLLM\cite{hao2023chatllm}, AutoGen\cite{wu2023autogen}, Caire\cite{lin2019caire}, Cheerbots\cite{jhan2021cheerbots}, PPDPP\cite{deng2023plug}, Xu et al.\cite{xu2022long}, Mitsui et al.\cite{mitsui2023towards}, etc.}, color=brightlavender!100, fill=brightlavender!15, very thick, text=black, tier=Task, text width=6.9cm
                ]
                [{\textbf{Role Playing:} ChatEval\cite{chan2023chateval}, RoleLLM\cite{wang2023rolellm}, Character-LLM\cite{shao2023character}, MMRole\cite{dai2024mmrole}, ECHO\cite{ng2024well}, etc.}, color=brightlavender!100, fill=brightlavender!15, very thick, text=black, tier=Task, text width=6.9cm
                ]
                [{\textbf{Personalized Assistants:} PET\cite{wu2023plan}, Gramopadhye et al.\cite{gramopadhye2023generating}, chen et al.\cite{chen2023interact}, etc.}, color=brightlavender!100, fill=brightlavender!15, very thick, text=black, tier=Task, text width=6.9cm
                ]
            ]
            [Entertainment Scenarios, color=brightlavender!100, fill=brightlavender!15, very thick, text=black
                [{\textbf{Game:} Voyager\cite{wang2023voyager}, GITM\cite{zhu2023ghost}, DEPS\cite{wang2023describe}, Plan4MC\cite{baai2023plan4mc}, Jarvis-1\cite{wang2023jarvis}, STEVE\cite{zhao2024see}, Nottingham et al.\cite{nottingham2023embodied}, etc.}, color=brightlavender!100, fill=brightlavender!15, very thick, text=black, tier=Task, text width=6.9cm
                ]
                [{\textbf{Media:} Mediagpt\cite{wang2023mediagpt}, Databricks\cite{databricks_media_entertainment}, Steck et al.\cite{steck2021deep}, etc.}, color=brightlavender!100, fill=brightlavender!15, very thick, text=black, tier=Task, text width=6.9cm
                ]
            ]
        ]
        [\cref{sec:domain_specific_application} Domain-Specific Appliacation, color=harvestgold!100, fill=harvestgold!15, very thick, text=black, text width=3.5cm
            [Education, color=harvestgold!100, fill=harvestgold!15, very thick, text=black
                [{Cgmi\cite{jinxin2023cgmi}, IDEE\cite{su2023unlocking}, AgentVerse\cite{chen2023agentverse}, Math Agents\cite{swan2023math}, Dona\cite{kalvakurthi2023hey}, etc.}, color=harvestgold!100, fill=harvestgold!15, very thick, text=black, tier=Task, text width=6.9cm
                ]
            ]
            [Law, color=harvestgold!100, fill=harvestgold!15, very thick, text=black
                [{Blind Judgement\cite{hamilton2023blind}, Legal prompting\cite{yu2022legal}, shui et al.\cite{shui2023comprehensive}, etc.}, color=harvestgold!100, fill=harvestgold!15, very thick, text=black, tier=Task, text width=6.9cm
                ]
            ]
            [Medical, color=harvestgold!100, fill=harvestgold!15, very thick, text=black
                [{Huatuogpt\cite{zhang2023huatuogpt}, Zhongjing\cite{yang2024zhongjing}, GMAI\cite{moor2023foundation}, Ada\cite{jungmann2019accuracy}, Ali et al.\cite{ali2020virtual}, etc.}, color=harvestgold!100, fill=harvestgold!15, very thick, text=black, tier=Task, text width=6.9cm
                ]
            ]
            [Other Domain, color=harvestgold!100, fill=harvestgold!15, very thick, text=black
                [{Bloomberggpt\cite{wu2023bloomberggpt}, Gao et al.\cite{gao2023multi}, north et al.\cite{north2007managing}, bonabeau et al.\cite{bonabeau2002agent}, etc.}, color=harvestgold!100, fill=harvestgold!15, very thick, text=black, tier=Task, text width=6.9cm
                ]
            ]
        ]
    ]
\end{forest}
\caption{
Overview of the major application areas for lifelong learning LLM-based agents. These applications are broadly categorized into everyday use cases—encompassing work, life, and entertainment scenarios—and domain-specific implementations, such as education, law, and healthcare. By continuously adapting to user needs and drawing on accumulated knowledge, lifelong agents improve user experiences, decision-making processes, and operational efficiency across diverse contexts.
}
\label{fig:tree_application}
\end{figure*}
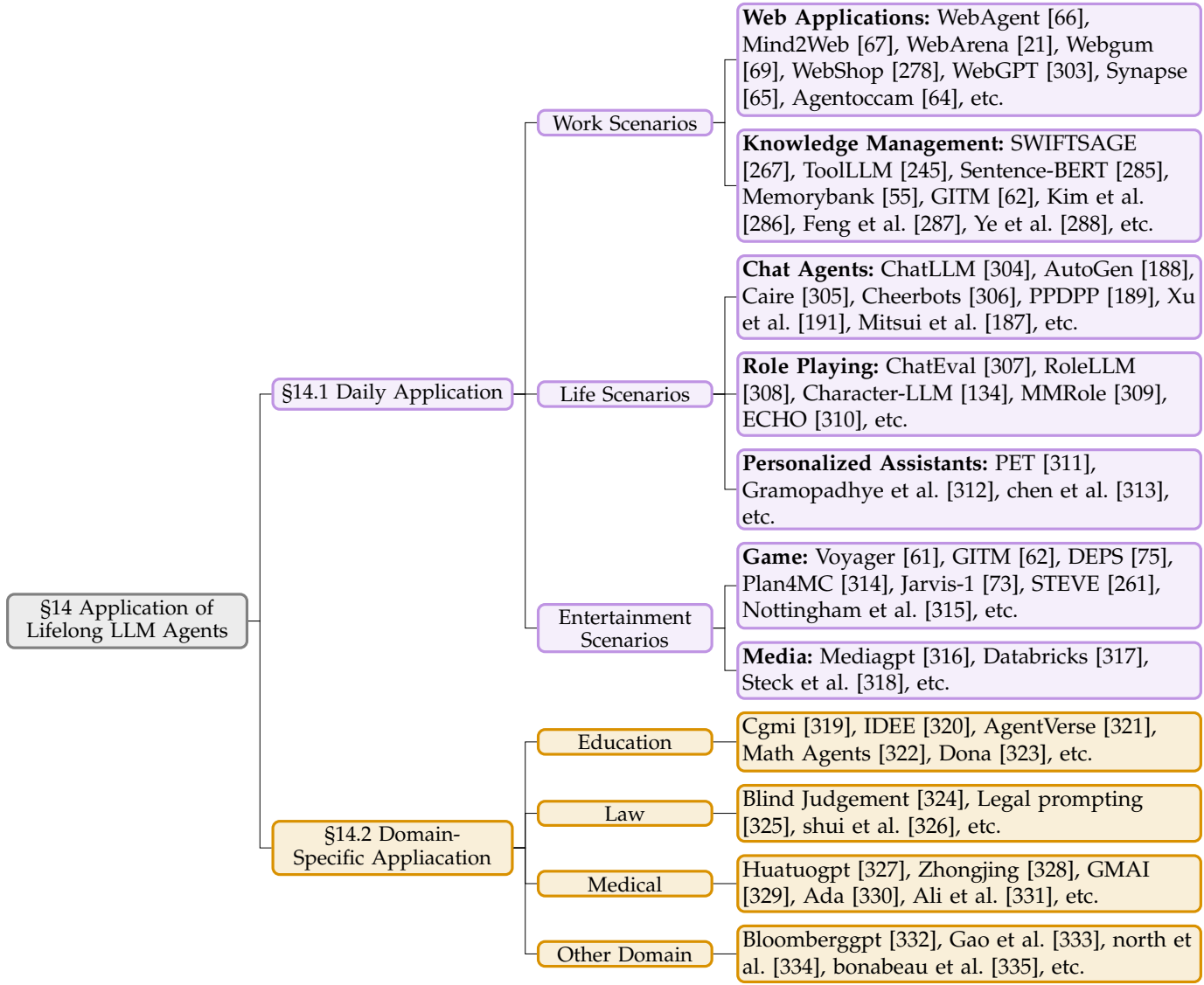

In the digital age, LLM-based agents play an increasingly significant role in both everyday life and professional settings. Driven by advances in the lifelong learning paradigm, these agents continuously adapt and optimize their functionalities to meet users’ evolving needs. The applications of these lifelong agents can be broadly categorized into two areas: \emph{daily application} and \emph{domain-specific application}. A summary of the related research appears in Figure \ref{fig:tree_application}.

\subsection{Daily Application}
\label{sec:daily_application}

In the context of human daily life, LLM agents significantly enhance people's work, life, and entertainment experiences through continuous learning and adaptation. These intelligent agents not only comprehend user needs but also dynamically adjust their functionalities to better serve users' everyday activities. Specifically, daily applications can be categorized into several key scenarios as follows:

\textbf{In work scenarios. }Agents play a variety of key roles that significantly enhance work and learning efficiency. For example, on web pages \cite{gur2023real,deng2024mind2web,furuta2023multimodal,zhou2024webarena,yao2022webshop,nakano2021webgpt,zheng2023synapse,yang2024agentoccam}, agents continuously optimize search algorithms and content recommendations through lifelong learning, helping users find relevant information and resources more efficiently. In knowledge management, LLM-based agents effectively organize and retrieve information \cite{zhu2023ghost,lin2023swiftsage,qin2024toolllm,reimers2019sentence,zhong2024memorybank,kim2023tree,feng2024retrieval,ye2023effective,cheng2024lift}, assisting users in quickly accessing the knowledge they need, promoting information sharing, and supporting decision-making.

\textbf{In life scenarios. }Lifelong agents have the potential to enhance the convenience and comfort of daily life. In terms of communication, LLM-based agents, integrating lifelong learning approaches such as role-playing \cite{chan2023chateval,wang2023rolellm,dai2024mmrole,shao2023character,ng2024well} and long-context text understanding \cite{chen2023walking,zhao2024longagent,chen2023clex,sun2022length,bae2022keep,xu2021beyond}, can engage in continuous interactions with users, gradually comprehending their personalities and preferences. This enables the provision of a more natural and emotionally resonant conversational experience \cite{hao2023chatllm,xu2022long,mitsui2023towards,wu2023autogen,deng2023plug,lin2019caire,jhan2021cheerbots}. As personalized assistants, these agents can also assist users in completing daily household tasks based on their environments \cite{gramopadhye2023generating,chen2023interact,wu2023plan}, such as automatically adjusting air conditioning, lighting, and cleaning, thereby improving the overall user experience.

\textbf{In entertainment scenarios. }Agents also play a significant role. For instance, in gaming, Minecraft, as an open-world simulation game, has emerged as a primary choice for testing agents within the gaming environment \cite{wang2023voyager,zhu2023ghost,wang2023describe,baai2023plan4mc,nottingham2023embodied,wang2023jarvis,zhao2024see}. Specifically, Voyager \cite{wang2023voyager}, which is the first lifelong agent in Minecraft, is capable of autonomous exploration of unknown worlds without human intervention, utilizing a feedback mechanism. JARVIS-1 \cite{wang2023jarvis} enhances its understanding of the environment through self-reflection and self-explanation, incorporating previous plans into its prompts. Additionally, the entertainment media industry is undergoing an intelligent transformation. By continuously gathering the latest information from users, it recommends relevant high-quality movies and music to them \cite{steck2021deep,wang2023mediagpt,databricks_media_entertainment}.

\subsection{Domain-Specific Application}
\label{sec:domain_specific_application}

In domain-specific applications, lifelong agents demonstrate remarkable adaptability and expertise, providing customized solutions across various industries. These intelligent agents continuously accumulate industry knowledge and user feedback through lifelong learning, thereby enhancing their effectiveness in specific fields.

\textbf{In the education domain}, LLM-based agents facilitate deep understanding of knowledge by simulating classroom environments and teacher-student interactions \cite{kalvakurthi2023hey,swan2023math,jinxin2023cgmi,chen2023agentverse,woolf2010building}, as well as offering personalized learning support and resources \cite{soller2003computational,swan2023math}. For teachers, the introduction of agents can assist in grading assignments, collecting class materials, and acting as an assistant to address students' questions \cite{kuo2023leveraging,prather2023robots}. For students, one of the main advantages of using large language models like ChatGPT in education is their ability to help students complete assignments more efficiently and provide personalized learning experiences \cite{su2023unlocking,haleem2022era,crompton2023artificial,kasneci2023chatgpt}. Additionally, a lifelong agent can also serve as a teacher to instruct other models \cite{doveh2023teaching,saha2023can,wu2023plan}.

\textbf{In the law domain}, these agents can analyze legal documents and cases to provide users with legal advice and compliance recommendations \cite{shui2023comprehensive}, as well as assist in legal decision-making and drafting \cite{bench2003model,branting2013reasoning,hamilton2023blind,iu2023chatgpt,yu2022legal}. 

\textbf{In the medical domain}, LLM-based agents also have broad applications \cite{kitamura2023chatgpt,kung2023performance,sallam2023chatgpt,cascella2023evaluating,moor2023foundation}. Some methods assist doctors in diagnostic and treatment decision-making \cite{zhang2023huatuogpt,yang2024zhongjing,ali2020virtual}. By using tools \cite{jungmann2019accuracy,malik2019overview}, agents enable the system to interact with patients, enhancing the quality and efficiency of healthcare services.

Furthermore, in other industries, lifelong agents can adapt to new real-world tasks through continuous learning, thereby reducing labor costs \cite{gao2023multi,north2007managing,bonabeau2002agent,wu2023bloomberggpt}.
\section{Practical Insights and Future Directions}
\label{sec:practical_insights_and_future_directions}

Building lifelong learning LLM agents involves a tight integration of perception, memory, and action modules, each playing a distinct yet interdependent role in enabling continuous adaptation to evolving tasks and environments \cite{xi2023rise,wang2024survey}. Although the progress in each module is substantial, several practical insights and future research directions remain.

\subsection{Perception Module: Enhancing Robustness and Multimodality}

The perception module has evolved from handling primarily textual inputs to addressing complex, multimodal inputs, including web content, images, and game environments \cite{zhou2024webarena}. Current approaches often rely on handcrafted prompt-engineering \cite{wang2023voyager}, compression strategies \cite{jiang2023llmlingua}, or pretrained vision-language alignment models \cite{liu2024autoglm} to incorporate new modalities. However, a key challenge is ensuring that the perception module continues to function reliably when confronted with novel data distributions or modalities absent during initial training.

\textbf{Adaptive Perception Architectures.} Developing methods for automatic modality selection and integration that scale to newly introduced data types without extensive human intervention is essential for enhancing adaptability.

\textbf{Domain-Agnostic Perception.} Approaches that utilize universal, modality-agnostic encoders can handle arbitrary input formats, continuously learning cross-modal alignment strategies to maintain versatility across different domains.

\textbf{Contextualized Perception Memory.} Implementing mechanisms to store and retrieve perception-related knowledge enables the agent to recall representations of previously encountered modalities or domains, facilitating more efficient adaptation.

\subsection{Memory Module: Balancing Stability, Plasticity, and Scalability}

The memory module—comprising working, episodic, semantic, and parametric memory—forms the backbone of an agent’s ability to learn over time. While current strategies mitigate catastrophic forgetting and integrate new knowledge, managing the ever-growing volume of information remains a significant challenge. Existing solutions often rely on replay buffers \cite{rolnick2019experience}, knowledge distillation \cite{peng2021hierarchical,gou2021knowledge,liu2024towards}, or architectural modifications \cite{razdaibiedinaprogressive,mallya2018packnet} to store and recall old knowledge.

\textbf{Memory Architecture Specialization.} Designing hierarchical memory structures that differentiate between short-term context (working memory), long-term events (episodic memory), structured knowledge (semantic memory), and learned weights (parametric memory) can enhance memory efficiency and retrieval accuracy.

\textbf{Optimized Retrieval Mechanisms.} Incorporating sophisticated retrieval strategies beyond naive nearest-neighbor approaches allows for the dynamic selection of the most relevant past experiences, scaling effectively with task complexity and the agent's lifetime.

\textbf{Dynamic Memory Management.} Investigating techniques that proactively prune or summarize outdated or less relevant knowledge is crucial for balancing memory growth with the retention of critical information.

\textbf{Neuro-inspired Consolidation.} Drawing inspiration from human cognition for memory consolidation processes ensures that new memories integrate or coexist with existing ones without overwriting them, promoting long-term knowledge retention.

\textbf{Transfer Learning for Rapid Adaptation.} Incorporating transfer learning techniques allows agents to quickly learn new tasks by leveraging previously acquired knowledge, enhancing learning speed and efficiency in dynamic environments \cite{pan2009survey}.

\textbf{Scalable Memory Architectures.} Developing memory systems that are infinitely expandable while ensuring fast storage and retrieval is essential for handling the growing knowledge base of lifelong learning agents. Techniques such as scalable indexing and distributed memory systems can achieve this balance.

\subsection{Action Module: Complex Reasoning and Efficient Adaptation}

The action module enables LLM agents to interact with their environment, perform grounding actions, retrieve knowledge, and reason effectively. While existing methods have successfully integrated tool usage \cite{qin2024toolllm,yuan2024easytool,schick2024toolformer}, web navigation \cite{nakano2021webgpt,zhou2024webarena}, and game-based tasks \cite{wang2023voyager}, the complexity of real-world environments calls for more advanced action design principles.

\textbf{Hierarchical Action Spaces.} Structuring action policies into hierarchies allows for flexible and scalable control, enabling agents to break down complex tasks into subtasks and reuse previously acquired skills.

\textbf{Continuous Self-improvement.} Leveraging reasoning trajectories, retrieval actions, and reflection-based methods can refine decision-making across trials and episodes, allowing the agent to improve its actions over its entire operational lifetime.

\textbf{Task-Agnostic Action Generalization.} Developing frameworks that promote the transferability of action policies across different tasks and environments reduces the need for environment-specific tuning or retraining.

\textbf{Human-in-the-Loop Fine-tuning.} Involving end-users or domain experts to provide corrective feedback and demonstrations guides the agent toward more robust and human-aligned action policies.

\textbf{Reinforcement Learning for Self-Learning and Evolution.} Integrating reinforcement learning approaches enables agents to self-learn and evolve through interactions with the environment, promoting autonomous adaptation and improvement over time \cite{sutton2018reinforcement}.

\subsection{Integrative Approaches and Long-Horizon Planning}

The interplay between perception, memory, and action modules is crucial \cite{xi2023rise}. As lifelong LLM agents must continuously integrate new modalities, knowledge, and action repertoires, future research should emphasize integrative approaches. Developing unified frameworks where each module can inform and refine the others is key to unlocking the full potential of lifelong adaptation.

\textbf{Perception-Memory-Action Feedback Loops.} Structuring an agent’s architecture to enable direct feedback across modules ensures that perception outputs are informed by memory state, and action decisions are guided by both perception and memory cues.

\textbf{Curriculum and Continual Curriculum Learning.} Introducing curricula that incrementally increase task complexity helps the agent’s evolving perception, memory, and action modules develop robust long-horizon planning capabilities.

\textbf{Tool and Knowledge Graph Integration.} Seamlessly incorporating external tools and knowledge bases improves both memory recall and action precision, enabling agents to leverage external resources for long-term skill acquisition.

\textbf{Collaborative Learning Frameworks.} Encouraging agents to learn collaboratively can enhance knowledge sharing and accelerate the learning process, leading to more efficient long-horizon planning and decision-making.

The field of lifelong LLM agents is poised for transformative growth \cite{wang2024survey}. By focusing on robust perception designs that adapt to new modalities, memory architectures that efficiently handle ever-growing knowledge, and action modules that continually refine decision-making, future research can foster agents that not only excel in their initial domains but also adapt gracefully to emerging tasks. Long-term, the goal is to create LLM agents that resemble human-like learners, capable of genuine lifelong learning—forever perceiving, reasoning, and acting in increasingly complex, dynamic worlds.

\section{Conclusion}
\label{sec:conclusion}
In this survey, we examine the evolving landscape of lifelong learning for LLM-based agents. We begin by highlighting the motivation and foundational definitions of lifelong LLM agents, underscoring their capacity not only to adapt to changing tasks and environments but also to continuously improve from past experiences. Through a systematic exploration of perception, memory, and action modules—each playing a critical role in enabling lifelong learning—we showcase how state-of-the-art methods and frameworks address the challenges of integrating and retaining knowledge over extended periods.

We further discuss evaluation metrics, benchmarks, and application scenarios that underscore the practical significance of lifelong LLM agents in both everyday and specialized domains. Finally, we offer insights and directions for future research, aiming to inspire the development of agents that more closely mirror the adaptability, resilience, and learning capabilities seen in human intelligence. As LLM-based agents continue to evolve, the pursuit of robust lifelong learning methods remains a vital frontier, promising increasingly capable, adaptable, and context-aware solutions to complex, real-world challenges.

\bibliography{reference}
\bibliographystyle{IEEEtran}

\end{document}